\documentclass{article}

\usepackage{microtype}
\usepackage{graphicx}
\usepackage{booktabs} 
\usepackage{mathtools}
\usepackage{enumitem}

 \usepackage{multirow} 
\usepackage{amsmath}
\usepackage{amsthm}
\usepackage{color}
\usepackage{bbm}
\usepackage{bm}
\usepackage{caption}
\usepackage{subcaption}
\usepackage{amssymb}
\usepackage{url}

\DeclareMathSymbol{\shortminus}{\mathbin}{AMSa}{"39}
\usepackage{soul}
\usepackage{tikz}
\usetikzlibrary{decorations.pathreplacing,positioning, arrows.meta}

\usepackage[accepted]{icml2021}
\icmltitlerunning{RNN with Particle Flow for Probabilistic Spatio-temporal Forecasting}

\begin{document}

\twocolumn[
\icmltitle{RNN with Particle Flow for Probabilistic Spatio-temporal Forecasting}

\begin{icmlauthorlist}
\icmlauthor{Soumyasundar Pal$^{\dagger}$}{mc}
\icmlauthor{Liheng Ma$^{\dagger}$}{mc}
\icmlauthor{Yingxue Zhang}{hw}
\icmlauthor{Mark Coates}{mc}
\end{icmlauthorlist}
\icmlaffiliation{mc}{Department of Electrical and Computer Eningeering, McGill University, Montr{\'e}al, QC,  Canada.}
\icmlaffiliation{hw}{ Huawei Noah's Ark Lab, Montr\'{e}al Research Center, Montr\'{e}al, QC, Canada}
\icmlcorrespondingauthor{Soumyasundar Pal}{soumyasundar.pal@mail.mcgill.ca}
\icmlkeywords{Machine Learning, ICML}
\vskip 0.3in
]
\printAffiliationsAndNotice{} 
\begin{abstract}
\noindent
Spatio-temporal forecasting has numerous applications in analyzing wireless, traffic, and financial networks. Many classical statistical models often fall short in handling the complexity and high non-linearity present in time-series data. Recent advances in deep learning allow for better modelling of spatial and temporal dependencies. While most of these models focus on obtaining accurate point forecasts, they do not characterize the prediction uncertainty. In this work, we consider the time-series data as a random realization from a nonlinear state-space model and target Bayesian inference of the hidden states for probabilistic forecasting. We use particle flow as the tool for approximating the posterior distribution of the states, as it is shown to be highly effective in complex, high-dimensional settings. Thorough experimentation on several real world time-series datasets demonstrates that our approach provides better characterization of uncertainty while maintaining comparable accuracy to the state-of-the-art point forecasting methods.    
\vspace{-0.5cm}
\end{abstract}
\let\thefootnote\relax\footnotetext{$\dagger$ Part of the research performed as an intern at Huawei Noah's Ark Lab, Montr\'{e}al Research Center.}
\let\thefootnote\relax\footnotetext{Code to reproduce our experiments is available at \url{https://github.com/networkslab/rnn_flow}}

\vspace{-0.1cm}
\section{Introduction}
\vspace{-0.1cm}
Spatio-temporal forecasting has many applications in intelligent traffic management, computational biology and finance, wireless networks and demand forecasting. Inspired by the surge of novel learning methods for graph structured data, many deep learning based spatio-temporal forecasting techniques have been proposed recently~\cite{li2018diffusion, bai2020}. In addition to the temporal patterns present in the data, these approaches can effectively learn and exploit spatial relationships among the time-series using various Graph Neural Networks (GNNs)~\cite{defferrard2016, kipf2017}. Recent works establish that graph-based spatio-temporal models outperform the graph-agnostic baselines~\cite{li2018diffusion, wu2020}. In spite of their accuracy in providing point forecasts, these models have a serious drawback as they cannot gauge the uncertainty in their predictions. When decisions are made based on forecasts, the availability of a confidence or prediction interval can be vital. 

There are numerous probabilistic forecasting techniques for multivariate time-series, for example, DeepAR~\cite{salinas2020}, DeepState~\cite{rangapuram2018}, DeepFactors~\cite{wang2019}, and the normalizing flow-based algorithms in~\cite{kurle2020,rasul2021}. Although these algorithms can characterize uncertainty via confidence intervals, they are not designed to incorporate side-knowledge provided in the form of a graph.  

In this work, we model multivariate time-series as random realizations from a nonlinear state-space model, and target Bayesian inference of the hidden states for probabilistic forecasting. The general framework we propose can be applied to univariate or multivariate forecasting problems, can incorporate additional covariates, can process an observed graph, and can be combined with data-adaptive graph learning procedures. For the concrete example algorithm deployed in experiments, we build the dynamics of the state-space model using graph convolutional recurrent architectures. We develop an inference procedure that employs particle flow, an alternative to particle filters, that can conduct more effective inference for high-dimensional states.

The novel contributions in this paper are as follows:\\
1) we propose a graph-aware stochastic recurrent network architecture and inference procedure that combine graph convolutional learning, a probabilistic state-space model, and particle flow;\\
2) we demonstrate via experiments on graph-based traffic datasets that a specific instantiation of the proposed framework can provide point forecasts that are as accurate as the state-of-the-art deep learning based spatio-temporal models. The prediction error is also comparable to the existing deep learning based techniques for benchmark non-graph multivariate time-series datasets;\\
3) we show that the proposed method provides a superior characterization of the prediction uncertainty compared to existing probabilistic multivariate time-series forecasting methods, both for datasets where a graph is available and for settings where no graph is available.
\vspace{-0.1cm}
\section{Related Work}
\vspace{-0.1cm}
Our work is related to (i) multivariate and spatio-temporal forecasting using deep
learning and graph neural networks; (ii)
stochastic/probabilistic modelling, prediction and forecasting for
multivariate time-series; and (iii) neural
(ordinary) differential equations. Recently, neural network-based techniques have started to
offer the best predictive performance for multivariate time-series prediction~\cite{bao2017, qin2017, lai2018,
  guo2018,  chang2018, li2019, sen2019, oreshkin2020,smyl2020}. In
some settings, a graph
is available that specifies spatial or causal relationships between
the time-series. Numerous algorithms have been proposed that combine
GNNs with temporal neural network
architectures~\cite{li2018diffusion,yu2018,huang2019,
  bai2019,chen2019,guo2019, wu2019,yu2019b,zhao2019,bai2020,
  huang2020,park2020,shi2020,song2020,xu2020,wu2020,zheng2020,oreshkin2021}. Algorithms that take into account the graph provide
superior forecasts, if the graph is accurate and the
indicated relationships have predictive power. However, none of these
algorithms is capable of characterizing the uncertainty of the
provided predictions; all are constructed as deterministic
algorithms.

Recently, powerful multivariate forecasting algorithms that are capable of providing uncertainty
characterization have been proposed. These include DeepAR~\cite{salinas2020},
DeepState~\cite{rangapuram2018}, the Multi-horizon Quantile RNN
(MQRNN)~\cite{wen2017}, the Gaussian copula process
approach of~\cite{salinas2019}, and
DeepFactors~\cite{wang2019}. Normalizing flow
has also been combined with temporal NN architectures~\cite{kumar2019,debezenac2020,gammelli2020,rasul2021}. Various flavours of stochastic recurrent
networks have also been introduced~\cite{boulanger2012, bayer2014, chung2015,
  fraccaro2016,fraccaro2017, karl2017,mattos2016, doerr2018}. In most cases, variational inference is
applied to learn model parameters, although sequential Monte Carlo has also been employed~\cite{gu2015,le2017,maddison2017,zheng2017,karkus2018, naesseth2018, ma2020}.
These are related to methods that determine the parameters of sequential Monte Carlo models via
optimizing Monte Carlo objectives~\cite{maddison2017, naesseth2018,
  le2017}.

Our proposed method is different from this body of work in two
important ways. First, we design a probabilistic state-space modeling
framework that can incorporate information about predictive
relationships that is provided in the form of a graph. Second, our
inference procedure employs particle flow, which avoids the need for
some of the approximations required by a variational inference
framework and is much better suited to high-dimensional states than
particle filtering. Our particle flow method has connections to normalizing flows~\cite{kobyzev2020} and to neural ordinary differential equations~\cite{chen2018,chen2019b}. In particular, \citet{chen2019b} address a Bayesian inference task by solving a differential equation to transport particles from the prior to the posterior distribution. However, such flow-based methods were first introduced by~\citet{daum2007} in the sequential inference research literature.

\vspace{-0.1cm}
\section{Problem Statement}
\vspace{-0.1cm}
We address the task of discrete-time multivariate time-series prediction, with the
goal of forecasting multiple time-steps ahead. We assume that there is
access to a historical dataset for training, but after training the
model must perform prediction based on a
limited window of historical data. Let
$\mathbf{y}_t \in \mathbb{R}^{N \times 1}$ be an observed multivariate
signal at time $t$ and $\mathbf{Z}_t \in \mathbb{R}^{N \times d_z}$ be an associated set of
covariates. The $i$-th element of $\mathbf{y}_t$ is the observation associated with
time-series $i$ at time-step $t$. 

We also allow for the possibility that there is access to
a graph $\mathcal{G}=(\mathcal{V}, \mathcal{E})$, where $\mathcal{V}$
is the set of $N$ nodes and $\mathcal{E} \subset \mathcal{V} \times
\mathcal{V}$ denotes the set of edges. In this case, each node
corresponds to one time-series. The edges indicate probable predictive
relationships between the variables, i.e., the presence of an edge
$(i,j)$ suggests that the historical data for time-series $i$ is
likely to be useful in predicting time-series $j$. The graph may be directed or undirected.

The goal is to construct a model
that is capable of processing, for some time offset $t_0$, the data $\mathbf{Y}_{t_0+1:t_0+P}$,
$\mathbf{Z}_{t_0+1:t_0+P+Q}$ and (possibly) the graph $\mathcal{G}$, to
estimate $\mathbf{Y}_{t_0+P+1:t_0+P+Q}$. The prediction algorithm
should produce both point estimates and prediction intervals. The
performance metrics for the point estimates include mean absolute
error (MAE), mean absolute percentage error (MAPE), and root mean
squared error (RMSE). For the prediction intervals, the performance
metrics include the Continuous Ranked Probability Score
(CRPS)~\cite{gneiting2007}, and the P10, P50, and P90 Quantile Losses
(QL)~\cite{salinas2020, wang2019}. Expressions for these performance
metrics are provided in the supplementary material. 

\begin{figure*}[h!]
\centering
\includegraphics[height=6cm]{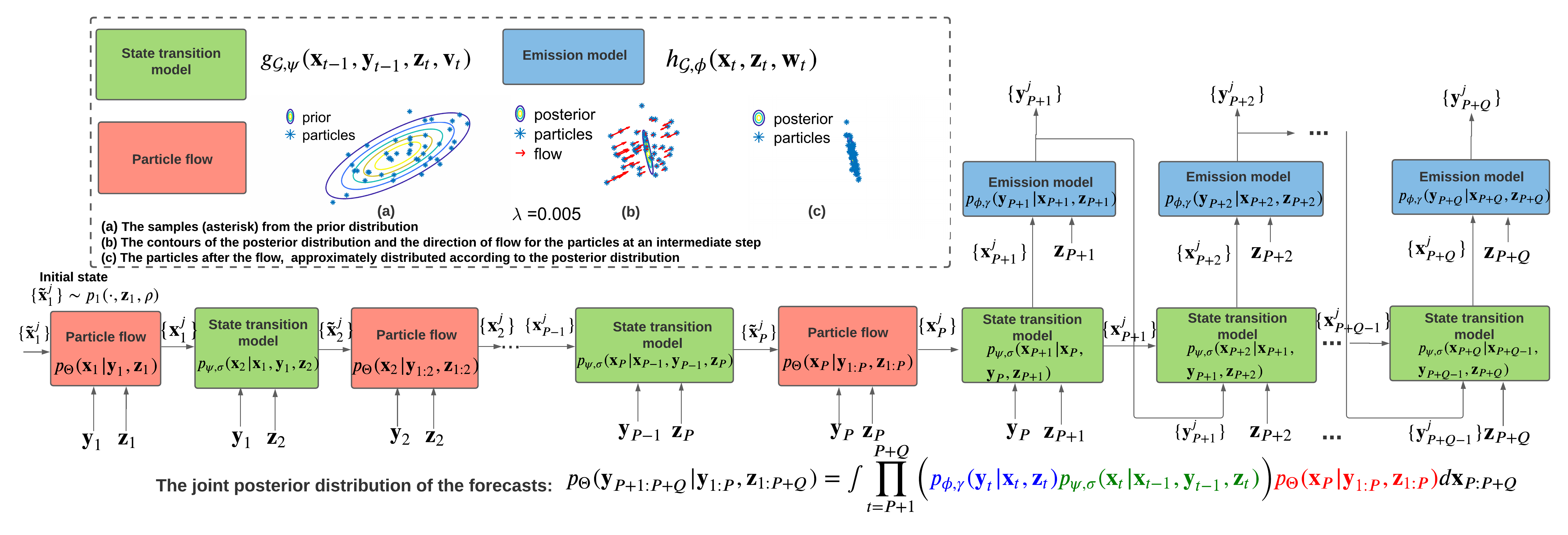}
\vspace{-0.5cm}
\caption{Probabilistic forecasting from the state-space model using particle flow. Migration of particles from a 2-d Gaussian prior to a 2-d Gaussian posterior distribution is illustrated as an example.}
\label{fig:model}
\vspace{-0.4cm}
\end{figure*}

\vspace{-0.1cm}
\section{Background: Particle Flow}
\vspace{-0.1cm}
\label{sec:flow}
Particle flow is an alternative to particle filtering for Bayesian
filtering (and prediction) in a state-space model. The filtering task is to approximate the posterior distribution of the state trajectory
$p_{\Theta}(\mathbf{x}_{t}|\mathbf{y}_{1:t})$ recursively, where
$\mathbf{x}_{t}$ denotes the state at time $t$ and $\mathbf{y}_{1:t}$
are observations from times $1$ to $t$. A particle filter~\cite{gordon1993,doucet2009}
maintains a population of $N_p$ samples (particles) and associated weights $\{\mathbf{x}_{t}^j,w_{t}^j\}_{j=1}^{N_p}$ that it uses to 
approximate the marginal posterior distribution of $\mathbf{x}_{t}$:
\vspace{-0.25cm}
\begin{align}
p_{\Theta}(\mathbf{x}_{t}|\mathbf{y}_{1:t}) \approx \frac{1}{N_p} \sum_{j=1}^{N_p} w^{(j)}_{t}\delta(\mathbf{x}_{t} - \mathbf{x}_{t}^j)\,.\label{eq:approx_posterior_prev}
\end{align}
Here, $\delta(\cdot)$ denotes the Dirac-delta function. Particles are propagated by the application of importance sampling
using a proposal distribution; the weights are updated
accordingly. When the disparity in the weights becomes too great,
resampling is applied, with particles being sampled proportionally to
their weights and the weights being reset to 1. Constructing
well-matched proposal distributions to the posterior distribution in
high-dimensional state-spaces is extremely challenging. A mismatch
between the proposal and the posterior leads to weight degeneracy after resampling, which results in poor performance of particle filters in high-dimensional problems~\cite{bengtsson2008,snyder2008,beskos2014}.
Instead of sampling, particle flow filters offer a significantly better solution by transporting particles continuously from the prior to the posterior~\cite{daum2007,ding2012}.

For a given time step $t$, particle flow algorithms solve differential equations to gradually
migrate particles from the predictive distribution so that they
represent the posterior distribution after the flow. A particle flow
can be modelled by a background stochastic process $\eta_\lambda$ in a pseudo-time interval $\lambda \in [0, 1]$,
such that the distribution of $\eta_0$ is the predictive distribution
$p_{\Theta}(\mathbf{x}_{t}|\mathbf{y}_{1:t-1})$ and the distribution
of $\eta_1$ is the posterior distribution
$p_{\Theta}(\mathbf{x}_{t}|\mathbf{y}_{1:t})$.

One approach~\cite{daum2010a}, is to use an ordinary differential
equation (ODE) with zero diffusion to govern the flow of $\eta_\lambda$:
\begin{equation}
\vspace{-0.1cm}
\dfrac{d\eta_\lambda}{d\lambda} = \varphi(\eta_\lambda,\lambda)\,\,.
\label{eq:dynamic_no_fusion}
\end{equation}
For linear Gaussian state-space models, the flow can be expressed in the form:
\begin{equation}
\vspace{-0.15cm}
\varphi(\eta_\lambda,\lambda)=A(\lambda)\eta_\lambda+b(\lambda)\,\,,
\label{eq:exact_flow}
\vspace{-0.05cm}
\end{equation}
and we can derive analytical expressions for $A(\lambda)$ and
$b(\lambda)$ (see supplementary material for details). For non-linear and
non-Gaussian models, we employ Gaussian approximations and
repeated local linearizations.

\vspace{-0.1cm}
\section{Methodology}
\vspace{-0.1cm}

\subsection{State-space model}
\begin{figure}[ht]
\vspace{-0.1cm}
\centering
\includegraphics[trim=0.5cm 0.6cm 0.5cm 0.5cm, clip, width=0.25\textwidth]{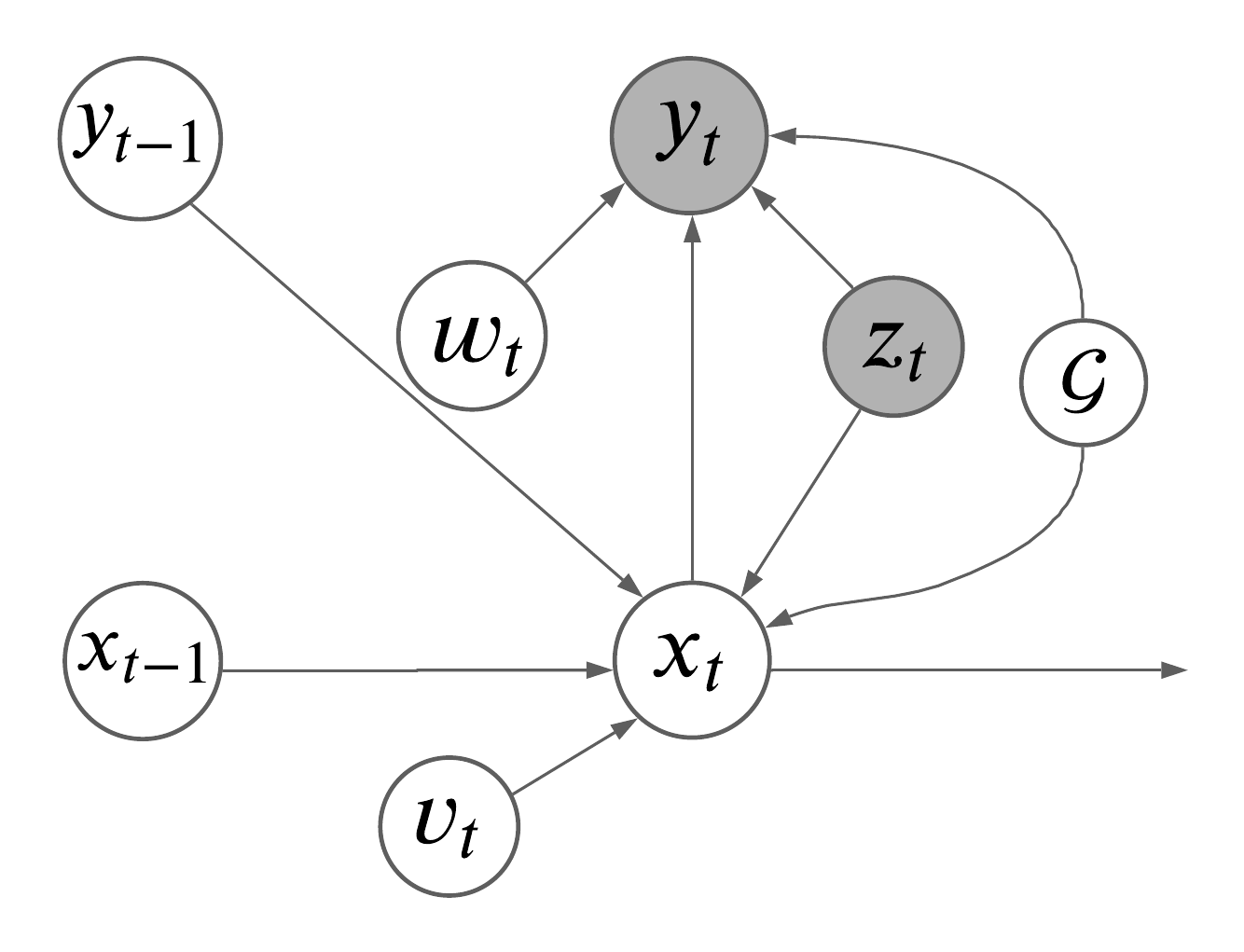}

\vspace{-0.14cm}
\caption{The graphical model representation of the state-space model in Section~\ref{sec:ssm}}
\vspace{-0.1cm}
\label{fig:graphical}
\end{figure}

\vspace{-0.15cm}
\label{sec:ssm}
We postulate that $\mathbf{y}_{t} \in \mathbb{R}^{N \times 1}$ is
the observation from a Markovian state space model with hidden state
$\mathbf{X}_t\in\mathbb{R}^{N \times d_x}$. We denote by $\mathbf{x}_t$ and $\mathbf{z}_t$ the
vectorizations of $\mathbf{X}_t$ and $\mathbf{Z}_t$, respectively.
The state space model is:
\vspace{-0.25cm}
\begin{align}
    \mathbf{x}_1 &\sim p_1(\cdot, \mathbf{z}_1, \rho)\,,\label{eq:initial}\\
    \mathbf{x}_t &= g_{\mathcal{G},\psi}(\mathbf{x}_{t-1},
                   \mathbf{y}_{t-1}, \mathbf{z}_{t}, \mathbf{v}_{t}), \text{ for } t > 1\,,\label{eq:dynamic_v1} \\
    \mathbf{y}_{t} &= h_{\mathcal{G},\phi}(\mathbf{x}_{t},
                     \mathbf{z}_{t}, \mathbf{w}_{t}), \text{ for } t \geqslant 1\,.\label{eq:measurement_v1}
\end{align}
Here $\mathbf{v}_t \sim p_v(\cdot| \mathbf{x}_{t-1}, \sigma)$ and $\mathbf{w}_t \sim p_w(\cdot| \mathbf{x}_{t}, \gamma)$ are the noises in the dynamic and measurement models respectively. $\rho$, $\sigma$ and $\gamma$ are the parameters of the distribution of the initial state $\mathbf{x}_1$, process noise $\mathbf{v}_t$ and measurement noise  
$\mathbf{w}_t$ respectively. $g$ and $h$ denote the state transition
and measurement functions, possibly linear or nonlinear, with
parameters $\psi$ and $\phi$ respectively. The subscript $\mathcal{G}$ in $g$ and $h$ indicates that the functions are potentially dependent on the graph topology. We assume that $h_{\mathcal{G},\phi}(\mathbf{x}_t, \mathbf{z}_t, 0)$ is a $\mathcal{C}^1$ function in $\mathbf{x}_t$, i.e., $h_{\mathcal{G},\phi}(\mathbf{x}_t, \mathbf{z}_t, \mathbf{0})$ is a differentiable function whose
first derivative w.r.t. $\mathbf{x}_t$ is continuous. The complete set
of the unknown parameters is formed as: $\Theta = \{\rho, \psi,
\sigma, \phi, \gamma\}$. Figure~\ref{fig:graphical} depicts the
graphical model relating the observed variables ($\mathbf{y}_t$ and
$\mathbf{z}_t$) to the latent variables ($v_t$, $w_t$) and the graph ($\mathcal{G}$).

With the proposed formulation, we can modify recurrent graph
convolutional architectures when designing the function $g$. When a
meaningful graph is available, such architectures significantly
outperform models that ignore the graph.  For
example, we conduct experiments by incorporating into our general model the Adaptive Graph Convolutional
Gated Recurrent Units (AGCGRU) presented in~\cite{bai2020}. The AGCGRU
combines (i) a module that adapts the provided graph based on observed
data, (ii) graph convolution to capture spatial relations, and (iii) a
GRU to capture evolution in time. The example model used for
experiments thus employs an $L$-layer AGCRU with additive Gaussian
noise to model the system dynamics $g$:
\vspace{-0.1cm}
\begin{align}
    \mathbf{x}_t &= AGCGRU_{\mathcal{G},\psi}^{(L)}(\mathbf{x}_{t-1},
  \mathbf{y}_{t-1}, \mathbf{z}_{t}) +
  \mathbf{v}_t\,,\label{eq:dynamic_agcgru}\\
  \mathbf{y}_t &=  \mathbf{W}_\phi\mathbf{x}_t +  \mathbf{w}_t\,.\label{eq:meas_linear}
\end{align}
In this model, we have 
$p_v(\mathbf{v}_t) = \mathcal{N}(\mathbf{0}, \sigma^2\mathbf{I})$,
i.e., the latent variables for the dynamics are independent. The
initial state distribution is also chosen to be isotropic Gaussian,
i.e., $p_1(\mathbf{x}_1, \mathbf{z}_1, \rho) = \mathcal{N}(\mathbf{0},
\rho^2\mathbf{I})$. The parameters $\rho$ and $\sigma$ are learnable
variance parameters. The observation model $g$ incorporates a linear
projection matrix $\mathbf{W}_{\phi}$. The latent variable
$\boldsymbol{w}_t$ for the emission model is modelled as Gaussian with
variance dependent on $\mathbf{x}_t$ via a learnable softplus function:
\vspace{-0.2cm}
\begin{equation}
p_w(\mathbf{w}_t| \mathbf{x}_{t}) = \mathcal{N}\Big(\mathbf{0}, \text{diag}\big(\text{softplus}(\boldsymbol{C}_\gamma\boldsymbol{x}_t)\big)^2\Big)\,.\label{eq:softplus}
\end{equation}

\vspace{-0.1cm}
\subsection{Inference}
\vspace{-0.15cm}
We assume that a dataset 
$\mathcal{D}_{trn}$ is available for training. Although this data may
be derived from a single time-series, because our task is to predict
$\mathbf{y}_{t_0+P+1:t_0+P+Q}$ using a limited historical window
$\mathbf{y}_{t_0+1:t_0+P}$, we splice the time-series and thus
construct multiple training examples, denoted by
$(\mathbf{y}_{1:P}^{(m)}, \mathbf{y}_{P+1:P+Q}^{(m)})$. In the
training set, all of these observations are available; in the test set
$\mathbf{y}_{P+1:P+Q}$ are not. In addition, the associated covariates $\mathbf{z}_{1:P+Q}$ are known for both training and test sets.

Inference involves an iterative process. We randomly initialize the
parameters of the model to obtain $\Theta_0$. Subsequently, at the
$k$-th iteration of the algorithm (processing the $k$-th training
batch), we first draw samples $\{\mathbf{y}^i_{P+1:P+Q}\}_{i=1}^{N_p}$ from the
distribution $p_{\Theta_{k-1}}(\mathbf{y}_{P+1:P+Q}|\mathbf{y}_{1:P}, \mathbf{z}_{1:P+Q})$. With
this set of samples, we can subsequently apply a gradient descent
procedure to obtain the updated model parameters $\Theta_k$. We
discuss each of these steps in turn as follows. 

\vspace{-0.1cm}
\subsubsection{Sampling}
\vspace{-0.15cm}
In a Bayesian setting with {\em known} model parameters $\Theta = \Theta_{k-1}$, we
would aim to form a prediction by approximating the posterior
distribution of the forecasts,
i.e., $p_{\Theta}(\mathbf{y}_{P+1:P+Q}|\mathbf{y}_{1:P})$. (For
conciseness we drop the time-offset $t_0$). 
\vspace{-0.5cm}
\begin{align}
&p_{\Theta}(\mathbf{y}_{P+1:P+Q}|\mathbf{y}_{1:P}, \mathbf{z}_{1:P+Q}) = \int \prod_{t=P+1}^{P+Q} \Big (p_{\phi, \gamma}(\mathbf{y}_{t} | \mathbf{x}_{t}, \mathbf{z}_{t})\,\nonumber \\
&p_{\psi, \sigma}(\mathbf{x}_{t} | \mathbf{x}_{t-1}, \mathbf{y}_{t-1}, \mathbf{z}_{t}) \Big) p_{\Theta}(\mathbf{x}_{P}| \mathbf{y}_{1:P}, \mathbf{z}_{1:P})d\mathbf{x}_{P:P+Q}\,.\label{eq:int_pred}
\end{align}
Since the integral in eq.~\eqref{eq:int_pred} is analytically intractable for a general nonlinear state-space model, we take a Monte Carlo approach as follows:

\textbf{Step 1: }
 For $1 \leqslant t \leqslant P$, we apply a particle flow algorithm (details in Sec.~\ref{sec:flow}) with $N_p$ particles for the state-space model specified by eqs.~\eqref{eq:initial},~\eqref{eq:dynamic_v1} and~\eqref{eq:measurement_v1} to recursively approximate the posterior distribution of the states:
 \vspace{-0.25cm}
\begin{equation}
p_{\Theta}(\mathbf{x}_{t}| \mathbf{y}_{1:t},  \mathbf{z}_{1:t}) \approx \frac{1}{N_p}\sum_{j=1}^{N_p} \delta(\mathbf{x}_{t} - \mathbf{x}_{t}^i)\,.\label{eq:approx_posterior}
 \vspace{-0.2cm}
\end{equation}
 Here $\{\mathbf{x}_{t}^j\}_{j=1}^{N_p}$ are approximately distributed
according to the posterior distribution of $\mathbf{x}_t$. The
generation of each sample $\mathbf{x}_{t}^j$ involves an associated
sampling of the latent variables $\mathbf{v}_{1:t}$ and implies a
sampling of $\mathbf{w}_{1:t}$, but these samples are not required
since the proposed model only needs
$\mathbf{x}_P$ to construct the forecast, 

\vspace{-0.1cm}
\textbf{Step 2: }
For $P+1 \leqslant t \leqslant P+Q$, we iterate between the following two steps:
\begin{enumerate}[leftmargin=*]
\vspace{-0.3cm}
\setlength\itemsep{0.1em}
\item[a.] We sample $\mathbf{x}_{t}^j$ from  $p_{\psi, \sigma}(\mathbf{x}_t | \mathbf{x}_{t-1}^j, \mathbf{y}_{t-1}^j, \mathbf{z}_{t})$ (for $t>P+1$) or from $p_{\psi, \sigma}(\mathbf{x}_t | \mathbf{x}_{t-1}^j, \mathbf{y}_{t-1}, \mathbf{z}_{t})$ (for $t=P+1$) for $1 \leqslant j \leqslant N_p$. This amounts to a state transition at time $t$ to obtain the current state $\mathbf{x}_t^j$ from the previous state $\mathbf{x}_{t-1}^j$, using eq.~\eqref{eq:dynamic_v1}.
\vspace{-0.1cm}
\item[b.] We sample $\mathbf{y}_{t}^j$ from  $p_{\phi, \gamma}(\mathbf{y}_{t} | \mathbf{x}_{t}^j, \mathbf{z}_{t})$ for $1 \leqslant j \leqslant N_p$, i.e., we use $\mathbf{x}_{t}^j$ in the measurement model, specified by eq.~\eqref{eq:measurement_v1}, to sample $\mathbf{y}_{t}^j$. 
\end{enumerate}
A Monte Carlo (MC) approximation of the integral in eq.~\eqref{eq:int_pred} is then formed as:
\vspace{-0.25cm}
\begin{equation}
p_{\Theta}(\mathbf{y}_{P+1:P+Q}|\mathbf{y}_{1:P}, \mathbf{z}_{1:P+Q}) \approx  \prod_{t=P+1}^{P+Q} \frac{1}{N_p}\sum_{j=1}^{N_p} \delta(\mathbf{y}_{t} - \mathbf{y}_{t}^j)\,.\label{eq:mc_pred} 
\end{equation}
Each $\mathbf{y}_{P+1:P+Q}^j$ is approximately distributed according to the joint posterior distribution of $\mathbf{y}_{P+1:P+Q}$. The resulting algorithm is summarized in Algorithm~\ref{alg:seq_seq}. A block diagram of the probabilistic forecasting procedure is shown in Figure~\ref{fig:model}.
\vspace{-0.1cm}

\begin{algorithm}[ht]
\caption{Sequence to sequence prediction}
\small
\label{alg:seq_seq}
\begin{algorithmic}[1]
\STATE {\bfseries Input:}  $\mathbf{y}_{1:P}, \mathbf{z}_{1:P+Q}$, and $\Theta$
\vspace{0.1cm}
\STATE {\bfseries Output:}  $\{\mathbf{y}_{P+1:P+Q}^j\}_{j=1}^{N_p}$
\vspace{0.1cm}
\STATE {\bfseries Initialization:} Sample $\eta_0^j \sim p_1(\mathbf{x}_{1}, \mathbf{z}_{1}, \rho)$, $j=1:N_p$.

\FOR{$t =1,2,...,P$}
\IF {$t > 1$}
\STATE Sample $\eta_0^j \sim p_{\psi, \sigma}(\mathbf{x}_{t}|\mathbf{x}_{t-1}^j, \mathbf{y}_{t-1}, \mathbf{z}_{t})$, $j=1:N_p$ as: $\eta_0^j = g_{\mathcal{G},\psi}(\mathbf{x}_{t-1}^j, \mathbf{y}_{t-1}, \mathbf{z}_{t}, \mathbf{v}_{t})$. 
\ENDIF
\STATE Use particle flow (details in supplementary material) to obtain $\{\eta_1^j\}_{j=1}^{N_p}$ from $\{\eta_0^j\}_{j=1}^{N_p}$, $\mathbf{z}_{t}$, and $\mathbf{y}_{t}$.
\STATE Set $\mathbf{x}_{t}^j = \eta_1^j$.
\ENDFOR

\FOR{$t =P+1,P+2,...,P+Q$}
\IF {$t = P+1$}
  \STATE Sample $\mathbf{x}_{P+1}^j \sim  p_{\psi, \sigma}(\mathbf{x}_{P+1} | \mathbf{x}_{P}^j, \mathbf{y}_{P}, \mathbf{z}_{P+1})$, $j=1:N_p$ as: $\mathbf{x}_{P+1}^j = g_{\mathcal{G},\psi}(\mathbf{x}_{P}^j, \mathbf{y}_{P}, \mathbf{z}_{P+1}, \mathbf{v}_{P+1})$.
  \ELSE 
  \STATE Sample $\mathbf{x}_{t}^j \sim  p_{\psi, \sigma}(\mathbf{x}_{t} | \mathbf{x}_{t-1}^j, \mathbf{y}_{t-1}^j, \mathbf{z}_{t})$, $j=1:N_p$ as: $\mathbf{x}_{t}^j = g_{\mathcal{G},\psi}(\mathbf{x}_{t-1}^j, \mathbf{y}_{t-1}^j,\mathbf{z}_{t}, \mathbf{v}_{t})$.
  \ENDIF

\STATE Sample $\mathbf{y}_{t}^j \sim  p_{\phi, \gamma}(\mathbf{y}_{t} | \mathbf{x}_{t}^j, \mathbf{z}_{t})$, $j=1:N_p$ as: $\mathbf{y}_{t}^j = h_{\mathcal{G},\phi}(\mathbf{x}_{t}^j, \mathbf{z}_{t}, \mathbf{w}_{t})$.
\ENDFOR
\STATE Form the Monte Carlo estimate using eq.~\eqref{eq:mc_pred}.
\end{algorithmic}
\vspace{-0.1cm}
\end{algorithm}

\vspace{-0.35cm}
\subsubsection{Parameter Update}
\label{sec:params_update}
\vspace{-0.15cm}
With the predictive samples $\{\mathbf{y}^j_{P+1:P+Q}\}_{j=1}^{N_p}$,
we can update the model parameters via Stochastic Gradient Descent (SGD) to
obtain $\Theta = \Theta_k$.

If our focus is on obtaining a point estimate, then we can perform
optimization on the training set with respect to a loss function
derived from Mean Absolute Error (MAE) or
Mean Square Error (MSE). The point forecast $\mathbf{\hat{y}}_{P+1:P+Q}^{(m)}$ is obtained based
on a statistic such as the mean or median of
the samples $\{\mathbf{y}_{P+1:P+Q}^{j,(m)}\}_{j=1}^{N_p}$.
The MAE loss function on a
dataset indexed by $\mathcal{D}$ can then be expressed as:
\vspace{-0.25cm}
\begin{equation}
    \mathcal{L}_{\text{MAE}}(\Theta, \mathcal{D}) = \frac{1}{N Q \lvert \mathcal{D} \rvert} \sum_{m \in \mathcal{D}} \sum_{t=P+1}^{P+Q} \lvert \lvert \mathbf{y}_{t}^{(m)} - \mathbf{\hat{y}}_{t}^{(m)}\rvert \rvert_1\,.\label{eq:mae_loss}
\end{equation}
In an alternate approach, we could consider the maximization of the
marginal log-likelihood over the training set. In that case, a suitable loss function is:
\vspace{-0.25cm}
\begin{equation}
\vspace{-0.4cm}
\mathcal{L}_{\text{prob}}(\Theta, \mathcal{D}) = -\frac{1}{\lvert \mathcal{D} \rvert} \sum_{m \in \mathcal{D}} \log p_{\Theta}(\mathbf{y}_{P+1:P+Q}^{(m)}|\mathbf{y}_{1:P}^{(m)}, \mathbf{z}_{1:P+Q}^{(m)})\,,\label{eq:prob_loss}   
\end{equation}
where we approximate the posterior probability as:
\vspace{-0.2cm}
\begin{equation}
\begin{aligned}
&\widehat{p}_{\Theta}(\mathbf{y}_{P+1:P+Q}|\mathbf{y}_{1:P}, \mathbf{z}_{1:P+Q}) =\nonumber\\
&\quad \prod_{t=P+1}^{P+Q}\Bigg[\frac{1}{N_p}\sum_{j=1}^{N_p}p_{\phi, \gamma}(\mathbf{y}_t|\mathbf{x}_t^j, \mathbf{z}_{t}) \Bigg]\,,\nonumber
\end{aligned}
\end{equation}
using eq.~\eqref{eq:int_pred}.
This loss formulation is similar to the MC variational
objectives in~\cite{maddison2017, naesseth2018,le2017}. If we use the particle flow particle
filter~\cite{li2017}, then the sampled particles and the propagated
forecasts form an unbiased approximation of the distribution
$p_{\Theta}(\mathbf{y}_{P+1:P+Q}|\mathbf{y}_{1:P}, \mathbf{z}_{1:P+Q})$. By Jensen's
inequality, the summation over the $\log$ terms
in~\eqref{eq:prob_loss} is thus a lower bound for the desired
$\mathbb{E}[\log  p_{\Theta}(\mathbf{y}_{P+1:P+Q}|\mathbf{y}_{1:P}, \mathbf{z}_{1:P+Q})]$
that converges as $N_p \rightarrow \infty$.

In each training mini-batch, for each
training example, we perform a forward pass through the model using
Algorithm~\ref{alg:seq_seq} to obtain approximate forecast
posteriors and then update all the model parameters using SGD via backpropagation. 

\begin{figure}[ht]
\centering
\hspace{-0.5em}
\includegraphics[trim=0cm 0.2cm 0cm 1.25cm, clip, width=0.49\textwidth]{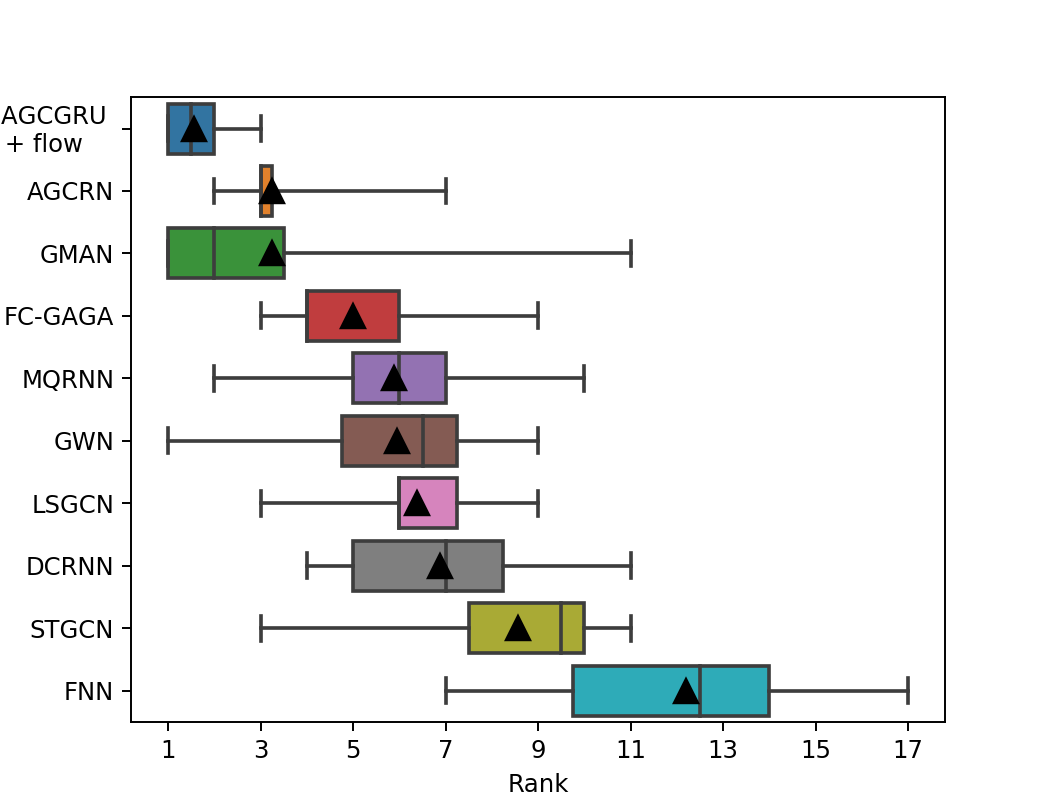}
\vspace{-0.5cm}
\caption{Boxplot of ranks of the top 10 algorithms across the four traffic datasets. The means of the ranks are shown by the black triangles; whiskers extend to the minimum and maximum ranks.}
\label{fig:rank}
\vspace{-0.25cm}
\end{figure}

\begin{table*}[htbp]
\footnotesize
\caption{Average MAE for PeMSD3, PeMSD4, PeMSD7, and PeMSD8 for 15/30/45/60 minutes horizons. The best and the second best results in each column are shown in bold and marked with underline respectively. Lower numbers are better.}
\centering
\begin{tabular}{c|c|c|c|c}
\hline
\multirow{2}{*}{Algorithm} & \multicolumn{4}{c}{MAE (15/ 30/ 45/ 60 min)} \\ \cline{2-5} 
                  &PeMSD3       &PeMSD4       &PeMSD7  &PeMSD8       \\ \hline \hline

DCRNN                 &14.42/15.87/17.10/18.29   
&1.38/1.78/2.06/2.29  
&2.23/3.06/3.67/4.18  
&1.16/1.49/1.70/1.87\\ 
STGCN                 &15.22/17.54/19.74/21.59    
&1.42/1.85/2.14/2.39
&2.21/2.96/3.47/3.90
&1.22/1.56/1.79/1.98\\ 
GWN                   &14.63/16.56/18.34/20.08        
&\underline{1.37}/1.76/2.03/2.24     
&2.23/3.03/3.56/3.98 
&\textbf{1.11}/1.40/1.59/1.73 \\ 
GMAN                  &14.73/15.44/\underline{16.15}/\underline{16.96} 
&1.38/\textbf{1.61}/\textbf{1.76}/\textbf{1.88}
&2.40/\underline{2.76}/\textbf{2.98}/\textbf{3.16}   
&1.23/\textbf{1.36}/\textbf{1.46}/\textbf{1.55}\\ 
AGCRN                 &\underline{14.20}/\underline{15.34}/16.28/17.38   
&1.41/1.67/1.84/\underline{2.01}   
&\underline{2.19}/2.81/3.15/3.42
&1.16/1.39/1.53/1.67     \\ 
LSGCN                 &14.28/16.08/17.77/19.23 
&1.40/1.78/2.03/2.20   
&2.23/2.99/3.50/3.95 
&1.21/1.54/1.75/1.89\\ 
FC-GAGA               &14.68/15.85/16.40/17.04     
&1.43/1.78/1.95/2.06  
&2.22/2.85/3.18/3.36   
&1.18/1.47/1.62/1.72  \\ \hline
DeepAR                &15.84/18.15/20.30/22.64           
&1.51/2.01/2.38/2.68 
&2.53/3.61/4.48/5.20
&1.25/1.61/1.87/2.10 \\ 
DeepFactors           &17.53/20.17/22.78/24.87            
&1.54/2.01/2.34/2.61      
&2.51/3.47/4.17/4.71 
&1.26/1.63/1.88/2.07 \\ 
MQRNN                 &14.60/16.55/18.34/20.12  
&\underline{1.37}/1.76/2.03/2.25           
&2.22/3.03/3.58/4.00
&\underline{1.13}/1.43/1.62/1.77  \\ 
AGCGRU+flow           &\textbf{13.79}/\textbf{14.84}/\textbf{15.58}/\textbf{16.06} 
&\textbf{1.35}/\underline{1.63}/\underline{1.78}/\textbf{1.88}   
&\textbf{2.15}/\textbf{2.70}/\underline{2.99}/\underline{3.19}
&\underline{1.13}/\underline{1.37}/\underline{1.49}/\underline{1.57} \\\hline
\end{tabular} 
\label{tab:mae}
\end{table*}

\begin{table*}[h!]
\footnotesize
\caption{Average MAE for PeMSD3, PeMSD4, PeMSD7, and PeMSD8 for 15/30/45/60 minutes horizons for the proposed flow based approach and deterministic encoder-decoder models. Lower numbers are better.}
    \centering
    \begin{tabular}{c|c|c|c|c}
        \hline
        \multirow{2}{*}{Algorithm} & \multicolumn{4}{c}{MAE (15/ 30/ 45/ 60 min)} \\ \cline{2-5} 
        &PeMSD3       &PeMSD4       &PeMSD7  &PeMSD8       \\ \hline \hline 
        AGCGRU+flow  &\textbf{13.79}/\textbf{14.84}/\textbf{15.58}/\textbf{16.06} &\textbf{1.35}/\textbf{1.63}/\textbf{1.78}/\textbf{1.88} &\textbf{2.15}/\textbf{2.70}/\textbf{2.99}/\textbf{3.19} &\textbf{1.13}/\textbf{1.37}/\textbf{1.49}/\textbf{1.57}\\
        FC-AGCGRU &13.96/15.37/16.52/17.45 &1.37/1.74/2.00/2.20 &2.21/2.99/3.56/4.05&1.16/1.48/1.70/1.87
       \\ \hline
        DCGRU+flow &14.48/\textbf{15.67}/\textbf{16.52}/\textbf{17.36} &\textbf{1.38}/\textbf{1.71}/\textbf{1.92}/\textbf{2.08} &\textbf{2.19}/\textbf{2.87}/\textbf{3.29}/\textbf{3.61} &1.17/\textbf{1.44}/\textbf{1.58}/\textbf{1.70} \\
        FC-DCGRU  &\textbf{14.42}/15.87/17.10/18.29 &\textbf{1.38}/1.78/2.06/2.29 &2.23/3.06/3.67/4.18 &\textbf{1.16}/1.49/1.70/1.87 
        \\ \hline
        GRU+flow  &\textbf{14.40}/\textbf{16.10}/\textbf{17.63}/\textbf{19.18} &\textbf{1.37}/\textbf{1.76}/\textbf{2.02}/\textbf{2.23} &\textbf{2.24}/\textbf{3.02}/\textbf{3.55}/\textbf{3.96} &\textbf{1.12}/\textbf{1.41}/\textbf{1.59}/\textbf{1.74}
        \\
        FC-GRU &15.82/18.37/20.61/22.93 &1.46/1.91/2.25/2.54 &2.41/3.40/4.17/4.84 &1.20/1.56/1.81/2.02
        \\ \hline
    \end{tabular}
    \label{tab:flow_no_flow}
\end{table*}

\begin{figure*}[h!]
\centering
\includegraphics[trim=5cm 0.1cm 3cm 2.0cm, clip, height=5.75cm]{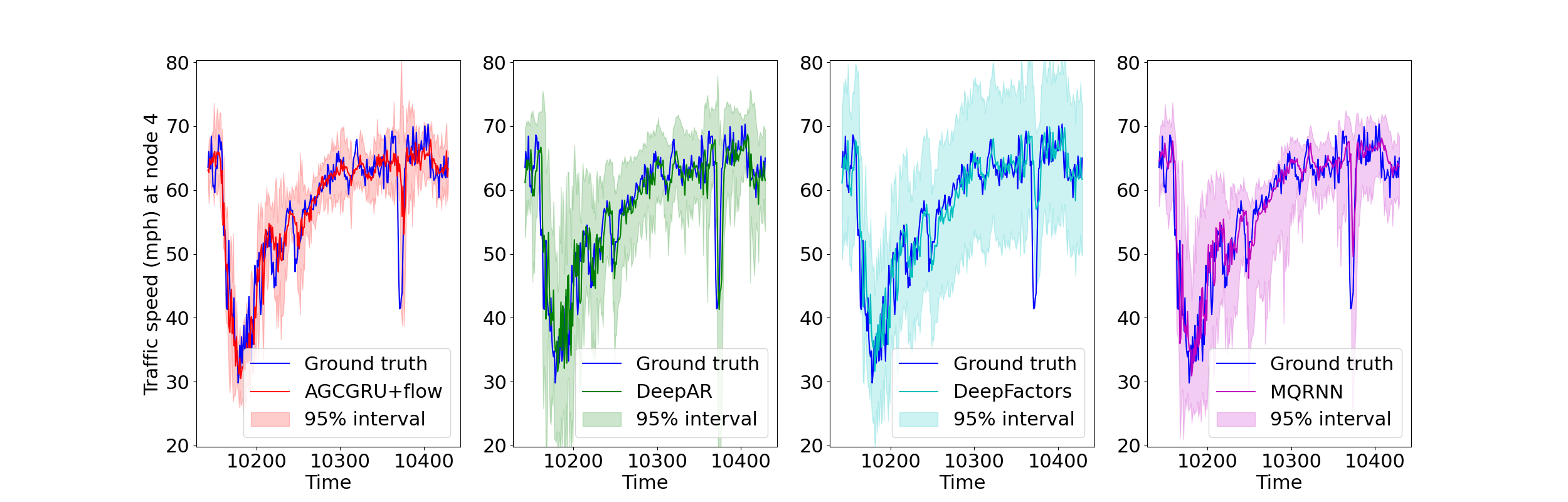}

\vspace{-0.25cm}
\caption{15 minutes ahead predictions from the probabilistic forecasting algorithms with confidence intervals at node 4 of PeMSD7 dataset for the first day in the test set. The proposed AGCGRU+flow algorithm provides tighter confidence interval than its competitors, which leads to lower quantile error.}
\label{fig:interval}
\vspace{-0.45cm}
\end{figure*}

\vspace{-0.1cm}
\section{Experiments}
\vspace{-0.1cm}
\label{sec:experiment}
We perform experiments on four graph-based and four non-graph based public datasets to evaluate  proposed methods. 


\vspace{-0.1cm}
\subsection{Datasets}
\label{sec:dataset}
\vspace{-0.15cm}
We evaluate our proposed algorithm on four publicly available traffic datasets, namely PeMSD3, PeMSD4, PeMSD7 and PeMSD8. These are obtained from the Caltrans Performance Measurement System (PeMS)~\cite{chen2000} and have been used in multiple previous works~\cite{yu2018,guo2019,song2020,bai2020,huang2020}. Each of these datasets consists of the traffic speed records, collected from loop detectors, and aggregated over 5 minute intervals, resulting in 288 data points per detector per day. In non-graph setting, we use Electricity~\cite{dua2017} (hourly time-series of the electricity consumption), Traffic~\cite{dua2017} (hourly occupancy rate, of different car lanes in San Francisco), Taxi~\cite{salinas2019}, and Wikipedia~\cite{salinas2019} (count of clicks to different web links) datasets. The detailed statistics of these datasets are summarized in the supplementary material.


\vspace{-0.1cm}
\subsection{Preprocessing}
\vspace{-0.15cm}
For the PeMS datasets, missing values are filled by the last known value in the same series. The training, validation and test split is set at 70/10/20\% chronologically and standard normalization of the data is used as in~\cite{li2018diffusion}. We use one hour of historical data ($P=12$) to predict the traffic  for the next hour ($Q=12$). Graphs associated with the datasets are constructed using the procedure in~\cite{huang2020}. 

\vspace{-0.3cm}
\subsection{Baselines}
\vspace{-0.15cm}
To demonstrate the effectiveness of our model, we compare to the
following forecasting methods. A detailed description of each baseline is provided in the supplementary material. 


\vspace{-0.1cm}
\underline{\textit{Spatio-temporal point forecast models}}: DCRNN~\cite{li2018diffusion}, STGCN~\cite{yu2018}, ASTGCN~\cite{guo2019}, GWN~\cite{yu2019}, GMAN~\cite{zheng2020}, AGCRN~\cite{bai2020}, LSGCN~\cite{huang2020}. \\
\underline{\textit{Deep-learning based point forecasting methods}}:\\ DeepGLO~\cite{sen2019}, N-BEATS~\cite{oreshkin2020}, and FC-GAGA~\cite{oreshkin2021}.\\
\underline{\textit{Deep-learning based probabilistic forecasting methods}}: DeepAR~\cite{salinas2020},  DeepFactors~\cite{wang2019}, and MQRNN~\cite{wen2017}. 

The detailed comparison of our approach with all of these models is provided in the supplementary material for space constraints. Here, we show results of a subset, focusing on those with the most competitive performance (Figure~\ref{fig:rank}). \footnote{} \footnotetext{Some of the recent spatio-temporal models such as~\cite{chen2020, zhang2020, park2020} do not have publicly available code. Although the codes for~\cite{wu2020, song2020, pan2019} are available, these works use different datasets for evaluation. We could not obtain sensible results from these models for our datasets, even with considerable hyperparameter tuning. The code for~\cite{kurle2020, debezenac2020} is not publicly available.}

\begin{table*}[h!]
\footnotesize
    \caption{Average CRPS, P10QL, P50QL, and P90QL for PeMSD3, PeMSD4, PeMSD7, and PeMSD8 for 15/30/45/60 minutes horizons. The best and the second best results in each column are shown in bold and marked with underline respectively. Lower numbers are better.}
    \centering
    \begin{tabular}{c|c|c|c|c}
    \hline 
    Dataset  &PeMSD3       &PeMSD4       &PeMSD7    &PeMSD8  \\   \hline \hline
    Algorithm &
    \multicolumn{4}{c}{CRPS (15/ 30/ 45/ 60 min)}            \\ \hline 
    DeepAR 
    &11.41/13.11/14.62/16.27  
    &\underline{1.13}/1.52/1.82/2.07
    &1.92/2.78/3.44/3.99 
    &\underline{0.94}/1.24/1.46/1.64\\
    DeepFactors  
    &14.16/15.87/17.59/18.99 
    &1.52/1.84/2.07/2.26
    &2.35/3.00/3.48/3.87
    &1.26/1.51/1.69/1.83\\
    GRU+flow 
    &11.23/12.70/13.98/15.25   
    &1.14/1.50/1.75/1.95
    &1.88/2.61/3.09/3.46 
    &0.95/1.23/1.42/1.57\\ 
    DCGRU+flow    
    &\underline{11.21}/\underline{12.14}/\underline{12.87}/\underline{13.64}     
    &\underline{1.13}/\underline{1.43}/\underline{1.63}/\underline{1.79}
    &\underline{1.85}/\underline{2.51}/\underline{2.95}/\underline{3.27}
    &\underline{0.94}/\underline{1.18}/\underline{1.35}/\underline{1.47}\\
    AGCGRU+flow &\textbf{10.53}/\textbf{11.39}/\textbf{12.03}/\textbf{12.47}     
    &\textbf{1.08}/\textbf{1.32}/\textbf{1.46}/\textbf{1.56}
    &\textbf{1.73}/\textbf{2.18}/\textbf{2.43}/\textbf{2.58}
    &\textbf{0.90}/\textbf{1.10}/\textbf{1.20}/\textbf{1.28}\\
    \hline
    \hline
     
    Algorithm & \multicolumn{4}{c}{P10QL(\%) (15/ 30/ 45/ 60 min)} \\ \hline 
    DeepAR 
    &4.11/4.69/5.21/5.69
    &1.37/1.96/2.45/2.86
    &2.56/3.90/4.92/5.78
    &1.14/1.59/1.93/2.24 \\
    DeepFactors  
    &5.85/6.33/6.91/7.51
    &2.13/2.61/3.01/3.34
    &3.49/4.53/5.46/6.26
    &1.77/2.17/2.49/2.76\\
    MQRNN 
    &\underline{4.03}/\underline{4.60}/5.13/5.68 
    &\textbf{0.95}/\textbf{1.18}/\textbf{1.31}/\textbf{1.40} 
    &\textbf{1.70}/\textbf{2.20}/\textbf{2.47}/\textbf{2.66}
    &\textbf{0.77}/\textbf{0.94}/\textbf{1.04}/\textbf{1.10} \\
    GRU+flow 
    &4.19/4.71/5.14/5.55
    &1.36/1.87/2.25/2.56
    &2.50/3.57/4.29/4.85
    &1.12/1.52/1.80/2.04 \\
    DCGRU+flow 
    &4.28/4.69/\underline{4.99}/\underline{5.28}
    &1.33/1.75/2.06/2.30
    &2.41/3.35/3.97/4.43
    &\underline{1.10}/\underline{1.43}/1.67/1.87\\
    AGCGRU+flow 
    &\textbf{4.01}/\textbf{4.44}/\textbf{4.76}/\textbf{4.97}
    &\underline{1.28}/\underline{1.62}/\underline{1.82}/\underline{1.97}
    &\underline{2.27}/\underline{2.97}/\underline{3.36}/\underline{3.60}
    &\underline{1.10}/\underline{1.43}/\underline{1.61}/\underline{1.73} \\
     \hline  \hline
     
    Algorithm & \multicolumn{4}{c}{P50QL(\%) (15/ 30/ 45/ 60 min)} \\ \hline 
    DeepAR 
    &9.11/10.44/11.68/13.03
    &2.37/3.15/3.73/4.20
    &4.35/6.21/7.70/8.95
    &1.97/2.52/2.94/3.30 \\
    DeepFactors  
    &10.08/11.60/13.11/14.31
    &2.42/3.15/3.68/4.10
    &4.31/5.97/7.16/8.10
    &1.97/2.55/2.95/3.25 \\
    MQRNN 
    &8.40/9.52/10.55/11.58
    &\underline{2.15}/2.77/3.19/3.53
    &3.82/5.21/6.16/6.88
    &\underline{1.77}/2.24/2.54/2.77 \\
    GRU+flow 
    &\underline{8.28}/9.26/10.15/11.04
    &2.16/2.76/3.17/3.50
    &3.84/5.19/6.10/6.81
    &\textbf{1.76}/\underline{2.21}/\underline{2.49}/2.72 \\
    DCGRU+flow 
    &8.33/\underline{9.01}/\underline{9.50}/\underline{9.99}
    &2.16/\underline{2.69}/\underline{3.01}/\underline{3.26}
    &\underline{3.77}/\underline{4.94}/\underline{5.66}/\underline{6.20} 
    &1.83/2.25/\underline{2.49}/\underline{2.66} \\
    AGCGRU+flow 
    &\textbf{7.93}/\textbf{8.54}/\textbf{8.96}/\textbf{9.24}
    &\textbf{2.11}/\textbf{2.55}/\textbf{2.79}/\textbf{2.94}
    &\textbf{3.70}/\textbf{4.65}/\textbf{5.14}/\textbf{5.49}
    &1.78/\textbf{2.15}/\textbf{2.34}/\textbf{2.46} \\
     \hline  \hline 
    
    Algorithm & \multicolumn{4}{c}{P90QL(\%) (15/ 30/ 45/ 60 min)} \\  \hline 
    DeepAR 
    &4.40/5.13/5.70/6.40
    &\underline{1.10}/1.45/1.67/1.84
    &2.13/3.03/3.65/4.08
    &\underline{0.93}/1.22/1.40/1.53  \\
    DeepFactors  
    &6.19/6.95/7.61/8.04
    &1.98/2.24/2.39/2.50
    &3.22/3.70/3.97/4.14
    &1.62/1.82/1.93/1.99  \\ 
    MQRNN 
    &\textbf{3.75}/\textbf{4.27}/\underline{4.70}/\underline{5.09}
    &1.22/1.68/2.03/2.32
    &2.19/3.12/3.78/4.30
    &0.99/1.34/1.59/1.80     \\
    GRU+flow 
    &4.33/4.94/5.48/6.04
    &1.11/1.43/1.63/1.77
    &2.02/2.74/3.16/3.44
    &\underline{0.93}/1.18/1.33/1.44   \\
    DCGRU+flow 
    &4.30/4.67/4.97/5.31
    &\underline{1.10}/\underline{1.34}/\underline{1.50}/\underline{1.61}
    &\underline{2.00}/\underline{2.62}/\underline{3.01}/\underline{3.28}
    &\underline{0.93}/\underline{1.13}/\underline{1.25}/\underline{1.34}  \\
    AGCGRU+flow 
    &\underline{4.06}/\underline{4.38}/\textbf{4.63}/\textbf{4.82} 
    &\textbf{1.05}/\textbf{1.26}/\textbf{1.37}/\textbf{1.45}
    &\textbf{1.83}/\textbf{2.25}/\textbf{2.48}/\textbf{2.62}
    &\textbf{0.87}/\textbf{1.01}/\textbf{1.09}/\textbf{1.14}  \\ \hline 
    \end{tabular}
    \label{tab:crps}
    \vspace{-0.25cm}
\end{table*}

\vspace{-0.1cm}
\subsection{Hyperparameters and training setup}
\vspace{-0.15cm}
For our model, we use an $L=2$ layer AGCGRU~\cite{bai2020} as the state-transition function. The dimension of the learnable node embedding is $d_e=10$, and the number of RNN units is $d_x=64$. We treat $\rho$ and $\sigma$ as fixed hyperparameters and set $\rho=1$ and $\sigma=0$ (no process noise). We train for 100 epochs using the Adam optimizer, with a batch size of 64. The initial learning rate is set to 0.01 and we follow a decaying schedule as in~\cite{li2018diffusion}. Hyperparameters associated with scheduled sampling~\cite{bengio2015}, gradient clipping, and early stoppng are borrowed from~\cite{li2018diffusion}. We set the number of particles $N_p=1$ during training and $N_p=10$ for validation and testing. The number of exponentially spaced discrete steps~\cite{li2017} for integrating the flow is $N_{\lambda}=29$. For each dataset, we conduct two separate experiments minimizing the training MAE (results are used to report MAE, MAPE, RMSE, and P50QL) and the training negative log posterior probability (results are used to report CRPS, P10QL, and P90QL). We also experiment with alternative state transition functions, including the DCGRU~\cite{li2018diffusion} and GRU~\cite{chung2014}. For these, the hyperparameters are fixed to the same values as presented above.

\vspace{-0.1cm}
\subsection{Results and Discussion}
\vspace{-0.15cm}
\paragraph{Comparison with baselines :}
Results for the point forecasting task are summarized in Table~\ref{tab:mae}. We observe that most of the spatio-temporal models perform better than graph agnostic baselines in most cases. Moreover, the proposed AGCGRU+flow algorithm achieves on par or better performance with the best-performing spatio-temporal models, such as GWN, GMAN and AGCRN. We present a comparison of the average rankings across datasets in Figure~\ref{fig:rank}. Our proposed method achieves the best average ranking and significantly outperforms the baseline methods. 
Table~\ref{tab:crps} summarizes the results for probabilistic forecasting. We observe that in most cases, the proposed flow based algorithms outperform the competitors. MQRNN also shows impressive performance in predicting the forecast quantiles, as it is explicitly trained to minimise the quantile losses. In particular, comparison of GRU+flow with the DeepAR model reveals that even without a sophisticated RNN architecture, the particle flow based approach shows better characterization of prediction uncertainty in most cases. Figure~\ref{fig:interval} provides a qualitative comparison of the uncertainty characterization, showing example confidence intervals for 15-minute ahead prediction for the PeMSD7 dataset. We see that the proposed algorithm provides considerably tighter intervals, while still achieving coverage of the observed values. 

\vspace{-0.3cm}
\paragraph{Generalization of particle flow inference across architectures :}
Table~\ref{tab:flow_no_flow} shows that in comparison to deterministic encoder-decoder based sequence to sequence prediction models, the proposed flow based approaches perform better in almost all cases for three different architectures of the RNN. In each case, both of the encoder-decoder model and our approach use a 2-layer architecture with 64 RNN units.

\vspace{-0.3cm}
\paragraph{Comparison to the particle filter :}
Table~\ref{tab:flow_bpf} demonstrates the effectiveness of particle flow~\cite{daum2007}, comparing it to a Bootstrap Particle Filter (BPF)~\cite{gordon1993} with the same number of particles. The use of the bootstrap particle filter leads to a computationally faster algorithm (requiring approximately 60\% of the training time of the particle flow-based method). 

\begin{table*}[h!]
\footnotesize
\caption{Average MAE and average CRPS for PeMSD3, PeMSD4, PeMSD7, and PeMSD8 for 15/30/45/60 minutes horizons for AGCGRU+flow and AGCGRU+BPF. Lower numbers are better.}
    \centering
    \begin{tabular}{c|c|c|c|c}
        \hline 
        Dataset        &PeMSD3       &PeMSD4       &PeMSD7  &PeMSD8  \\  \hline  \hline
         Algorithm  & \multicolumn{4}{c}{MAE (15/ 30/ 45/ 60 min)}   \\ \hline 
        AGCGRU+flow     &\textbf{13.79}/\textbf{14.84}/\textbf{15.58}/\textbf{16.06} &\textbf{1.35}/\textbf{1.63}/\textbf{1.78}/\textbf{1.88} &\textbf{2.15}/\textbf{2.70}/\textbf{2.99}/3.19   &\textbf{1.13}/\textbf{1.37}/\textbf{1.49}/\textbf{1.57}                      \\
        AGCGRU+BPF  &14.19/15.13/15.85/16.35    &1.36/1.65/1.80/1.90    &2.19/2.73/\textbf{2.99}/\textbf{3.17}    &1.18/1.41/1.52/1.59 \\ 
        \hline
        \hline 
        Algorithm & \multicolumn{4}{c}{CRPS (15/ 30/ 45/ 60 min)}   \\ \hline 
        AGCGRU+flow   &\textbf{10.53}/\textbf{11.39}/\textbf{12.03}/\textbf{12.47} &\textbf{1.08}/\textbf{1.32}/1.46/1.56  &\textbf{1.73}/\textbf{2.18}/\textbf{2.43}/\textbf{2.58} &\textbf{0.90}/\textbf{1.10}/\textbf{1.20}/\textbf{1.28}  \\
        AGCGRU+BPF &11.32/11.94/12.55/12.92 &1.10/\textbf{1.32}/\textbf{1.45}/\textbf{1.54} &1.79/2.24/2.49/2.66 &0.96/1.13/1.22/\textbf{1.28}\\
        \hline
    \end{tabular}
    \label{tab:flow_bpf}
    \vspace{-0.48cm}
\end{table*}

\begin{table*}[ht]
\footnotesize
\caption{Average CRPS for PeMSD3, PeMSD4, PeMSD7, and PeMSD8 for 15/30/45/60 minutes horizons for AGCRN-ensemble, GMAN-ensemble, and AGCGRU+flow. The best result in each column is shown in bold. Lower numbers are better.}
    \centering
    \begin{tabular}{c|c|c|c|c}
        \hline
        \multirow{2}{*}{Algorithm} & \multicolumn{4}{c}{CRPS (15/ 30/ 45/ 60 min)} \\ \cline{2-5} 
        &PeMSD3       &PeMSD4       &PeMSD7  &PeMSD8       \\ \hline 
        AGCRN-ensemble  &12.64/13.44/13.96/14.27 &1.20/1.44/1.56/1.68 &1.90/2.39/2.60/2.81 &1.03/1.20/1.28/1.38\\
        GMAN-ensemble
        &12.79/13.49/14.13/14.77 &1.16/1.38/1.51/1.62 &1.96/2.31/2.53/2.73 &0.95/1.10/1.19/1.28\\
        AGCGRU+flow       &\textbf{10.53}/\textbf{11.39}/\textbf{12.03}/\textbf{12.47}  &\textbf{1.08}/\textbf{1.32}/\textbf{1.46}/\textbf{1.56}  &\textbf{1.73}/\textbf{2.18}/\textbf{2.43}/\textbf{2.58} &\textbf{0.90}/\textbf{1.10}/\textbf{1.20}/\textbf{1.28}\\
        \hline
    \end{tabular} 
    \label{tab:ensemble}
    \vspace{-0.25cm}
\end{table*}

\begin{table*}[h!]
\footnotesize
\caption{Normalized Deviation on Electricity and Traffic datasets. The best and the second best results in each column are shown in bold and marked with underline respectively. Lower numbers are better.}
\centering
\begin{tabular}{c|cccc|ccc} 
\hline
\multirow{2}{*}{Algorithm} & \multicolumn{4}{c}{Electricity} & \multicolumn{3}{|c}{Traffic} \\ \cline{2-8}
 & 2014-09-01 & 2014-03-31 & 2014-12-18 & last 7 days & 2008-06-15 & 2008-01-14 & last 7 days\\ \hline
 TRMF & 0.160 & n/a & 0.104 & 0.255 & 0.200 & n/a & 0.187 \\ 
 DeepAR & 0.070 & 0.272 & 0.086 & n/a & 0.170 & 0.296 & 0.140\\
 DeepState & 0.083 & n/a & n/a & n/a & 0.167 & n/a & n/a\\
 DeepFactors & n/a & 0.112 & n/a  & n/a & n/a &\textbf{0.225} & n/a\\
 DeepGLO & n/a & n/a & \underline{0.082} & n/a & n/a & n/a & 0.148\\
 N-BEATS &\textbf{0.064} &\textbf{0.065} & n/a & \underline{0.171} &\textbf{0.114} &\underline{0.230} &\textbf{0.110} \\ \hline
 FC-GRU &0.102 &0.118 &0.098 &0.203 &0.259 &0.528 &0.233 \\
 GRU+flow &\underline{0.070} &\underline{0.071} &\textbf{0.069} &\textbf{0.140} &\underline{0.133} &0.322 &\underline{0.125} \\ \hline 
\end{tabular}
\label{table:electricity}
\label{tab:nbeats_exp}
\vspace{-0.25cm}
\end{table*}

\begin{table*}[h!]
\centering
\footnotesize
\caption{Average $\text{CRPS}_{sum}$ for Electricity, Traffic, Taxi, and Wikipedia datasets. The best and the second best results in each column are shown in bold
and marked with underline respectively. Lower numbers are better}
\begin{tabular}{c|cccccccc} 
\hline
Dataset     & \begin{tabular}[c]{@{}c@{}}Vec-LSTM\\ ind-scaling\end{tabular} & \begin{tabular}[c]{@{}c@{}}Vec-LSTM\\ lowrank-Copula\end{tabular} & \begin{tabular}[c]{@{}c@{}}GP\\ scaling\end{tabular} & \begin{tabular}[c]{@{}c@{}}GP\\ Copula\end{tabular} & \begin{tabular}[c]{@{}c@{}}LSTM\\ Real-NVP\end{tabular} & \begin{tabular}[c]{@{}c@{}}LSTM\\ MAF\end{tabular} & \begin{tabular}[c]{@{}c@{}}Transformer\\ MAE\end{tabular} & \begin{tabular}[c]{@{}c@{}}GRU+\\ flow\end{tabular} \\ \hline
Electricity & 0.025                                                          & 0.064                                                             & 0.022                                                & 0.024                                               & 0.024                                                   & \underline{0.021}                                              & \underline{0.021}                                                     &\textbf{0.013}                                                     \\
Traffic     & 0.087                                                          & 0.103                                                             & 0.079                                                & 0.078                                               & 0.078                                                   & 0.069                                              & \underline{0.056}                                                     & \textbf{0.028}                                                    \\
Taxi        & 0.506                                                          & 0.326                                                             & 0.183                                                & 0.208                                               & 0.175                                                   & \underline{0.161}                                              & 0.179                                                     &\textbf{0.140}                                                     \\
Wikipedia   & 0.133                                                          & 0.241                                                             & 1.483                                                & 0.086                                               & 0.078                                                   & 0.067                                              & \underline{0.063}                                                     & \textbf{0.054}           \\ \hline                               
\end{tabular}
\label{tab:compare_rasul}
\vspace{-0.48cm}
\end{table*}

\vspace{-0.3cm}
\paragraph{Comparison to ensembles :}
We compare the proposed approach with an ensemble of competitive deterministic forecasting techniques. We choose the size of the ensemble so that the algorithms have an approximately equal execution time. We use AGCRN and GMAN to form the ensembles, as they are the best point-forecast baseline algorithms. From Table~\ref{tab:ensemble}, we observe that the proposed AGCGRU+flow achieves lower average CRPS compared to the ensembles in all cases. 

\vspace{-0.3cm}
\paragraph{Point forecasting results on non-graph datasets :}
We evaluate our proposed flow-based RNN on the Electricity and Traffic datasets, following the setting described in Appendix C.4 in~\cite{oreshkin2020}. We augment the results table in~\cite{oreshkin2020} with the results from an FC-GRU (a fully connected GRU encoder-decoder) and GRU+flow. We use a 2 layer GRU with 64 RNN units in both cases. We follow the preprocessing steps in~\cite{oreshkin2020}. In the literature, four different data splits have been used for the Electricity dataset, and three different splits have been used for the Traffic dataset. The evaluation metric is P50QL (Normalized Deviation). 

In Table~\ref{tab:nbeats_exp}, we observe that the flow based approach performs comparably or better than the state-of-the-art N-BEATS algorithm for the Electricity dataset, even with a simple GRU as the state transition function. The better performance of the univariate N-BEATS compared to the multivariate models suggests that most time-series in these datasets do not provide valuable additional information for predicting other datasets. This is in contrast to the graph-based datasets, where the performance of N-BEATS was considerably worse than the multivariate algorithms. The proposed flow-based algorithm achieves prediction performance on the Traffic dataset that is comparable to N-BEATS except for one split with limited training data. Across all datasets and split settings, our flow-based approach significantly outperforms the FC-GRU. The proposed algorithm outperforms TRMF, DeepAR, DeepState and DeepGLO. It outperforms DeepFactors for the Electricity dataset, but is worse for the Traffic dataset (for the same split with limited available training data). 
\vspace{-0.3cm}
\paragraph{Probabilistic forecasting results on non-graph datasets :}
For comparison with state-of-the-art deep learning based probabilistic forecasting methods on standard non-graph time-series datasets, we evaluate the proposed GRU+flow algorithm following the setting in~\cite{rasul2021}. The results reported in Table 1 of~\cite{rasul2021} are augmented with the results of the GRU+flow algorithm. We use a 2 layer GRU with 64 RNN units in each case. We
follow the preprocessing steps as in~\cite{salinas2019, rasul2021}. The evaluation metric is (normalized) $\text{CRPS}_{sum}$ (defined in the supplementary material),  which is obtained by first summing across the different time-series, both for the ground-truth test data, and samples of forecasts, and then computing the (normalized) CRPS on the summed data. The results are summarized in Table~\ref{tab:compare_rasul}. We observe that the proposed GRU+flow achieves the lowest $\text{CRPS}_{sum}$ for all datasets.
\vspace{-0.35cm}
\paragraph{Computational complexity :} 
For simplicity, we consider a GRU instead of a graph convolution based RNN and we only focus on one sequence instead of a batch. Our model has to perform both GRU computation and particle flow for the first $P$ time steps and then apply the GRU and the linear projection for the next $Q$ steps to generate the predictions. For an $L$-layer GRU with $d_x$ RNN units and $N$-dimensional input, the complexity of the GRU operation for $N_p$ particles is $\mathcal{O}((P+Q)N_pLNd_x^2)$~\cite{chung2014}. The total complexity of the EDH particle flow~\cite{choi2011} is $\mathcal{O}(PN_{\lambda}N^3)$ for computing the flow parameters and $\mathcal{O}(PN_pN_{\lambda}Nd_x^2)$ for applying the particle flow (more details in the supplementary material). The total complexity of the measurement model for $N_p$ particles is $\mathcal{O}(QN_pNd_x^2)$. Since in most cases $N \gg d_x$ and $N \gg N_p$, the complexity of our algorithm for forecasting of one sequence is $\mathcal{O}(PN_{\lambda}N^3)$. Many of the other algorithms exhibit a similar $\mathcal{O}(N^3)$ complexity, e.g. TRMF, GMAN. We specify the execution time and memory usage in the supplementary material. Scaling the proposed methodology to extremely high dimensional settings is of significant importance and can be addressed in several ways. For spatio-temporal predictions using the graph-based recurrent architectures, this can be done if the graph can be partitioned meaningfully. For non-graph datasets, we can use the cross-correlation among different time-series to group them into several lower-dimensional problems. Alternatively, we can train a univariate model based on all the time-series as in~\cite{rangapuram2018}.

\vspace{-0.2cm}
\section{Conclusion}
\vspace{-0.1cm}
In this paper, we propose a state-space probabilistic modeling framework for multivariate time-series prediction that can process information provided in the form of a graph that specifies (probable) predictive or causal relationships. We develop a probabilistic forecasting algorithm based on the Bayesian inference of hidden states via particle flow. For spatio-temporal forecasting, we use GNN based architectures to instantiate the framework. Our method demonstrates comparable or better performance in point forecasting and considerably better performance in uncertainty characterization compared to existing techniques.
\vspace{-0.2cm}
\section*{Acknowledgments}
\vspace{-0.1cm}
This work was partially supported by the Department of National Defence’s Innovation for Defence Excellence and Security (IDEaS) program, Canada. We also acknowledge the support of the Natural Sciences and Engineering Research Council of Canada (NSERC), [260250].

\section*{\centering \Large Supplementary Material}

\section{Particle flow for Bayesian inference in a state-space model}
\subsection{Background}
Particle flow is an alternative to particle filters for Bayesian filtering in a state-space model. Recall the first order Markov model specified in eqs.~\eqref{eq:initial},~\eqref{eq:dynamic_v1}, and~\eqref{eq:measurement_v1} in Section~\ref{sec:ssm} of the main paper.
\begin{align}
    \mathbf{x}_1 &\sim p_1(\cdot, \mathbf{z}_1, \rho)\,,\label{eq:initial_v2}\\
    \mathbf{x}_t &= g_{\mathcal{G},\psi}(\mathbf{x}_{t-1},
                   \mathbf{y}_{t-1}, \mathbf{z}_{t}, \mathbf{v}_{t}), \text{ for } t > 1\,,\label{eq:dynamic_v2} \\
    \mathbf{y}_{t} &= h_{\mathcal{G},\phi}(\mathbf{x}_{t},
                     \mathbf{z}_{t}, \mathbf{w}_{t}), \text{ for } t \geqslant 1\,.\label{eq:measurement_v2}
\end{align}
Here $\mathbf{y}_t$ is the observation from the state-space model at time $t$. $\mathbf{x}_t$ and $\mathbf{z}_t$ denote the unobserved state variable and observed covariates at time $t$ respectively.
The filtering task is to compute the posterior distribution of the state trajectory
$p_{\Theta}(\mathbf{x}_{t}|\mathbf{y}_{1:t}, \mathbf{z}_{1:t})$ recursively. Suppose we have a set of $N_p$ samples (particles) $\{\mathbf{x}_{t-1}\}_{j=1}^{N_p}$ which approximates the posterior distribution of $\mathbf{x}_{t-1}$. 
\begin{align}
p_{\Theta}(\mathbf{x}_{t-1}|\mathbf{y}_{1:t-1}, \mathbf{z}_{1:t-1}) \approx \frac{1}{N_p} \sum_{j=1}^{N_p} \delta(\mathbf{x}_{t-1} - \mathbf{x}_{t-1}^j)\,.\label{eq:approx_posterior_prev_v1} 
\end{align}
In the `predict' step, we approximate the  predictive posterior distribution at time $t$ as follows:
\begin{align}
p_{\Theta}(\mathbf{x}_{t}|\mathbf{y}_{1:t-1}, \mathbf{z}_{1:t}) &= \int p_{\psi,\sigma}(\mathbf{x}_{t}|\mathbf{x}_{t-1},\mathbf{y}_{t-1}, \mathbf{z}_{t})\,\nonumber\\
&\qquad  p_{\Theta}(\mathbf{x}_{t-1}|\mathbf{y}_{1:t-1}, \mathbf{z}_{1:t-1}) d\mathbf{x}_{t-1}\,,\nonumber \\
&\approx \frac{1}{N_p} \sum_{j=1}^{N_p} \delta(\mathbf{x}_{t} - \mathbf{\tilde{x}}_{t}^j)  \,,\label{eq:predictive_approx}
\end{align}
where, the particles $\{\mathbf{\tilde{x}}_{t}^j\}_{j=1}^{N_p}$ from the predictive posterior distribution $p_{\Theta}(\mathbf{x}_{t}|\mathbf{y}_{1:t-1}, \mathbf{z}_{1:t})$ are obtained by propagating $\{\mathbf{x}_{t-1}^j\}_{j=1}^{N_p}$ through the state-transition model specified by eq.~\eqref{eq:dynamic_v2}. Subsequently, the `update' step applies Bayes' theorem to compute the posterior distribution at time $t$ as follows:
\begin{align}
p_{\Theta}(\mathbf{x}_{t}|\mathbf{y}_{1:t}, \mathbf{z}_{1:t}) \propto p_{\Theta}(\mathbf{x}_{t}|\mathbf{y}_{1:t-1}, \mathbf{z}_{1:t}) p_{\phi, \gamma}(\mathbf{y}_{t}|\mathbf{x}_{t}, \mathbf{z}_{t})\,.\label{eq:update_step}    
\end{align}
For non-linear state space models, particle filters~\cite{gordon1993,doucet2009} employ importance sampling to approximate the `update' step in eq.~\eqref{eq:update_step}. However, constructing well-matched proposal distributions to the posterior distribution in high-dimensional state-spaces is extremely challenging. A mismatch between
the proposal and the posterior leads to weight degeneracy after resampling, which results in poor performance of particle filters in high-dimensional problems~\cite{bengtsson2008,snyder2008,beskos2014}.
Instead of sampling, particle flow filters offer a significantly better solution in complex problems by transporting particles continuously from the prior to the posterior~\cite{daum2007,ding2012,daum2014b,daum2017}.

\subsection{Particle flow}
\begin{figure*}[htbp]
\centering
\includegraphics[trim=3cm 1cm 3cm 2cm, clip, height=7cm]{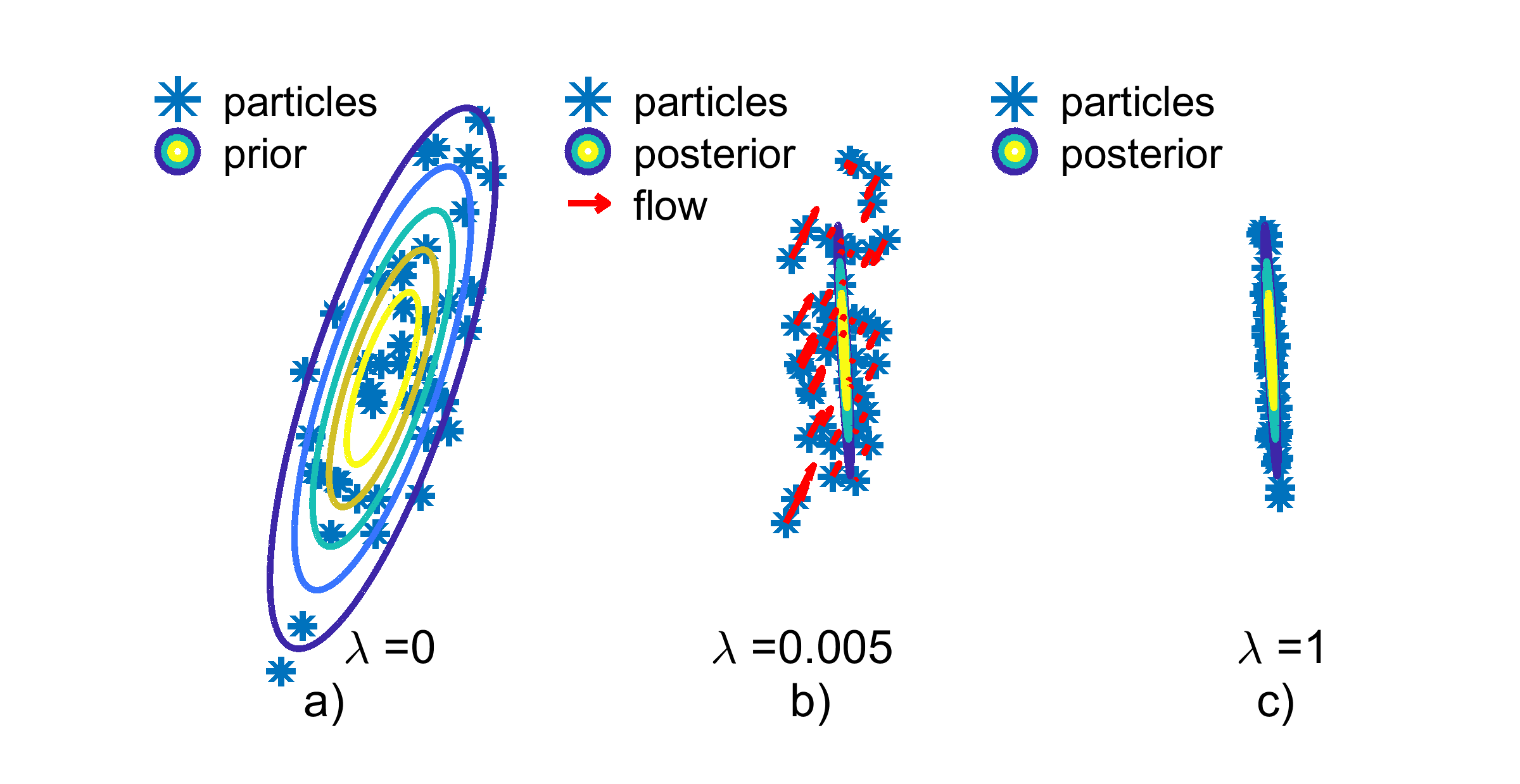}
\caption{Migration of particles from a 2-d Gaussian prior to a 2-d Gaussian posterior distribution. a) The samples (asterisk) from the prior distribution, b) The contours of the posterior distribution and the direction of flow for the particles at an intermediate step, c) The particles after the flow, approximately distributed according to the posterior distribution.}
\label{fig:flow}
\vspace{-0.4cm}
\end{figure*}
In a given time step $t$, particle flow algorithms~\cite{daum2007,daum2010a} solve differential equations to gradually
migrate particles from the predictive distribution  such that they represent the posterior distribution for the same time step after the flow. A particle flow can be modelled by a background stochastic process
$\eta_\lambda$ in a pseudo-time interval $\lambda \in [0, 1]$,
such that the distribution of $\eta_0$ is the predictive distribution
$p_{\Theta}(\mathbf{x}_{t}|\mathbf{y}_{1:t-1}, \mathbf{z}_{1:t})$ and the distribution of $\eta_1$ is the posterior distribution $p_{\Theta}(\mathbf{x}_{t}|\mathbf{y}_{1:t}, \mathbf{z}_{1:t})$. Since particle flow only considers migration of particles within a single time step, we omit the time index $t$ in $\eta_{\lambda}$, $\mathbf{y}$, and $\mathbf{z}$ to simplify notation.

In~\cite{daum2010a}, an ordinary differential equation (ODE) with zero diffusion governs the flow of $\eta_\lambda$:
\begin{align}
\dfrac{d\eta_\lambda}{d\lambda} = \varphi(\eta_\lambda,\lambda)\,\,.
\label{eq:dynamic_no_fusion_v1}
\end{align}
If the predictive distribution and the additive measurement noise is Gaussian and the measurement function $h$ is linear, the Exact Daum-Huang (EDH) flow is given as:
\begin{align}
\varphi(\eta_\lambda,\lambda)=A(\lambda)\eta_\lambda+b(\lambda)\,\,,
\label{eq:exact_flow_v1}
\end{align} 
where,
\begin{align}
A(\lambda)&=-\dfrac{1}{2}\bar{P}H^T(\lambda H\bar{P}H^T+R)^{-1}H\,\,,\label{eq:EDH_A_linear}\\
b(\lambda)&=(I+2\lambda A(\lambda))[(I+\lambda A(\lambda))\bar{P}H^TR^{-1}\mathbf{y}+A(\lambda)\bar{\eta}_0]\,\,,\label{eq:EDH_b_linear}
\end{align}
Here $\bar{\eta}_0$ and $\bar{P}$ are the mean vector and the covariance matrix of the predictive distribution respectively. For a general nonlinear state-space model, we usually a Gaussian approximation of the predictive distribution based on sample estimates of $\bar{\eta}_0$ and $\bar{P}$ or via the Extended Kalman Filter (EKF). $\mathbf{y}$ denotes the new observation at time $t$. The linear measurement model in $\mathbf{x}$ is specified by the measurement matrix $H = \frac{h_{\mathcal{G}, \phi}(\mathbf{x}, \mathbf{z}, \mathbf{0})}{\partial \mathbf{x}}$, and $R$ denotes the covariance matrix of the zero mean additive Gaussian measurement noise. For a nonlinear measurement model, we use a first order Taylor series approximation at the mean of the particles $\bar{\eta}_{\lambda}$ and replace $H$ by $H(\lambda) = \frac{\partial h_{\mathcal{G}, \phi}(\eta, \mathbf{z}, \mathbf{0})}{\partial \eta}\Bigg|_{\eta= \bar{\eta}_{\lambda}}$ and $\mathbf{y}$ by $\big(\mathbf{y}-e(\lambda)\big)$, where the linearization error $e(\lambda) = h_{\mathcal{G}, \phi}(\bar{\eta}_{\lambda}, \mathbf{z}, \mathbf{0}) - H(\lambda)\bar{\eta}_{\lambda}$ in eq.~\eqref{eq:EDH_A_linear} and~\eqref{eq:EDH_b_linear}. Similarly, for a zero mean non-Gaussian measurement noise, we use a Gaussian approximation to replace $R$ in eq.~\eqref{eq:EDH_A_linear} and~\eqref{eq:EDH_b_linear} by $R(\lambda) = \mathrm{Cov}\left[\mathbf{y}|\bar{\eta}_{\lambda}, \mathbf{z}\right]$. A detailed description of the implementation of the exact Daum-Huang (EDH) filter is provided in~\cite{choi2011}.

\begin{algorithm}[ht]
\caption{Particle flow}
\label{alg:flow}
\begin{algorithmic}[1]
\STATE {\bfseries Input:}  $\{\eta_0^j = \mathbf{\tilde{x}}_t^j\}_{j=1}^{N_p}$, $\mathbf{y}_t$, $\mathbf{z}_t$, $\{\epsilon_m\}_{m=1}^{N_\lambda}$, and $\Theta$
\vspace{0.1cm}
\STATE {\bfseries Output:}  $\{\mathbf{x}_t^j= \eta_1^j\}_{j=1}^{N_p}$
\vspace{0.1cm}
\STATE Compute $\bar{\eta}_0 = \frac{1}{N_p}\sum_{j=1}^{N_p} \eta_0^{j}$
\STATE Compute $\bar{P} = \frac{1}{N_p}\sum_{j=1}^{N_p}\left[(\eta_0^j-\bar{\eta}_0)(\eta_0^j-\bar{\eta}_0)^T\right]$
\STATE Set $\lambda_0 = 0$
\FOR{$m =1,2,...,N_{\lambda}$}
\STATE $\lambda_m = \lambda_{m-1} + \epsilon_m$
\STATE Linearize the measurement model at $\bar{\eta}_{\lambda_{m-1}}=\frac{1}{N_p}\sum_{j=1}^{N_p} \eta_{\lambda_{m-1}}^{j}$ to compute $H(\lambda_{m-1})$ and $e(\lambda_{m-1})$. 
\STATE Compute $R(\lambda_{m-1}) = \textbf{Cov}\left[\mathbf{y}_t|\bar{\eta}_{\lambda_{m-1}}, \mathbf{z}_t\right]$.
\STATE Compute $A(\lambda_{m-1})$ and $b(\lambda_{m-1})$ using eq.~\eqref{eq:EDH_A_linear} and~\eqref{eq:EDH_b_linear}.
\STATE Apply particle flow to all particles: $\eta_m^{j} = \eta_{m-1}^{j} + \epsilon_m\big(A(\lambda_{m-1})\eta_{m-1}^{j} +b(\lambda_{m-1})\big)$ 
\ENDFOR
\STATE Set $\mathbf{x}_t^j = \eta_1^{j}$ for $1 \leqslant j \leqslant N_p$
\end{algorithmic}
\end{algorithm}

Numerical integration is usually used to solve the ODE in Equation~\eqref{eq:exact_flow_v1}. The integral between $\lambda_{m-1}$ and $\lambda_{m}$ for $1 \leqslant m \leqslant N_\lambda$, where $\lambda_{0} = 0$
and $\lambda_{N_\lambda}=1$,
is approximated via the Euler update rule.
For the $j$-th particle, the EDH flow in the $m$-th pseudo-time interval becomes: 
\begin{align}
\eta_{\lambda_{m}}^{j} &= \eta_{\lambda_{m-1}}^{j}
+ \epsilon_m\big(A(\lambda_{m-1})\eta_{\lambda_{m-1}}^j
+ b(\lambda_{m-1})\big)\,,
\label{eq:discrete_update}
\end{align}
where the step size $\epsilon_m = \lambda_m - \lambda_{m-1}$ and $\displaystyle\sum_{m=1}^{N_\lambda} \epsilon_m =1$.
We start the particle flow from $\eta_0^j = \mathbf{\tilde{x}}_t^j$ and after the flow is complete, we set $\mathbf{x}_t^j = \eta_1^j$ to approximate the posterior distribution of $\mathbf{x}_t$ as:
\begin{align}
p_{\Theta}(\mathbf{x}_{t}|\mathbf{y}_{1:t}, \mathbf{z}_{1:t}) \approx \frac{1}{N_p} \sum_{j=1}^{N_p} \delta(\mathbf{x}_{t} - \mathbf{x}_{t}^j)\,.\label{eq:approx_posterior_current} 
\end{align}
The overall EDH particle flow algorithm is summarized in Algorithm~\ref{alg:flow}. Figure~\ref{fig:flow} demonstrates the migration of the particles from the prior to the posterior distribution for a Gaussian predictive distribution and a linear-Gaussian measurement model. 

\section{Model training}
Algorithm~\ref{alg:train_test} summarizes the learning of the model parameters $\Theta$, described in Section~\ref{sec:params_update} of the main paper.

\begin{algorithm}[ht]
\caption{Model training and testing}
\label{alg:train_test}
\begin{algorithmic}[1]
\STATE {\bfseries Input:} Training and test data: $\{\mathbf{y}_{1:P+Q}^{(m)},$ $\mathbf{z}_{1:P+Q}^{(m)}\}_{m \in \mathcal{D}_{trn}}$, $\{\mathbf{y}_{1:P}^{(n)}, \mathbf{z}_{1:P+Q}^{(n)}\}_{n \in \mathcal{D}_{test}}$
\vspace{0.1cm}
\STATE {\bfseries Output:}  $\{\hat{p}_{\hat{\Theta}}(\mathbf{y}_{P+1:P+Q}^{(n)}|\mathbf{y}_{1:P}^{(n)}, \mathbf{z}_{1:P+Q}^{(n)})\}_{n \in \mathcal{D}_{test}}$
\STATE {\bfseries Hyperparameters:} Number of iterations $N_{iter}$, step-size $\{\zeta_k\}_{k=1}^{N_{iter}}$
\vspace{0.1cm}
\STATE {\bfseries Initialization:} random initialization for the system parameters $\Theta_0$
\STATE {\bfseries Model training:}
\STATE Set $k = 1$
\WHILE{$k \leqslant N_{iter}$}
\STATE Sample a minibatch $\mathcal{D} \subset \mathcal{D}_{trn}$.

\STATE Compute the approximate posterior distribution of the forecasts $\{\hat{p}_{\Theta_{k-1}}(\mathbf{y}_{P+1:P+Q}^{(m)}|\mathbf{y}_{1:P}^{(m)}, \mathbf{z}_{1:P+Q}^{(m)})\}_{m \in \mathcal{D}}$ using Algorithm~\ref{alg:seq_seq} in the main paper with the current parameters $\Theta_{k-1}$.

\STATE Compute the gradient of the chosen loss function $\mathcal{L}(\Theta, \mathcal{D})$ w.r.t. model parameters $\Theta$ at $\Theta_{k-1}$ 

\STATE Update the system parameters using SGD algorithm: $\Theta_{k} = \Theta_{k-1} - \zeta_k \nabla_{\Theta} \mathcal{L}(\Theta, \mathcal{D})\big|_{\Theta=\Theta_{k-1}}$
\STATE $k = k+1$
\ENDWHILE

\STATE Save the estimated model $\widehat{\Theta} = \Theta_{N_{iter}}$

\STATE {\bfseries Testing:}

\STATE Compute the test set forecast posterior distributions $\{\hat{p}_{\hat{\Theta}}(\mathbf{y}_{P+1:P+Q}^{(n)}|\mathbf{y}_{1:P}^{(n)}, \mathbf{z}_{1:P+Q}^{(n)})\}_{n \in \mathcal{D}_{test}}$ using Algorithm~\ref{alg:seq_seq} in the main paper with the estimated model parameters $\widehat{\Theta}$.

\end{algorithmic}
\end{algorithm}

\section{Description and statistics of datasets}
\begin{table}[htbp]
\centering
\caption{Summary statistics of the PeMS road traffic datasets}
\small
\setlength{\tabcolsep}{2pt}
\begin{tabular}{c|c|c|c|c}
\hline
Dataset        & PeMSD3    & PeMSD4    & PeMSD7    & PeMSD8    \\ \hline
No. nodes      & 358       & 307       & 228       & 170       \\
No. time steps &26208  &16992 &12672 &17856 \\
Interval       & 5 min & 5 min & 5 min & 5 min \\\hline
\end{tabular}
\label{tab:pems_data}
\end{table}

The statistics of the PeMS datasets and the non-graph datasets used in our experiments are summarized in Tables~\ref{tab:pems_data} and~\ref{tab:non_graph_data} respectively. The description of the PeMS datasets are provided in Section~\ref{sec:dataset} of the main paper. The Electricity\footnotemark \footnotetext{\url{https://archive.ics.uci.edu/ml/datasets/ElectricityLoadDiagrams20112014}} dataset contains electricity consumption for
370 clients. The Traffic\footnotemark \footnotetext{\url{https://archive.ics.uci.edu/ml/datasets/PEMS-SF}} dataset is composed of 963 time-series of lane occupancy rates. The Taxi\footnotemark \footnotetext{\url{https://www1.nyc.gov/site/tlc/about/tlc-trip-record-data.page}} dataset contains counts of taxis on different roads and the Wikipedia\footnotemark \footnotetext{\url{https://github.com/mbohlkeschneider/gluon-ts/tree/mv_release/datasets}} dataset specifies clicks to web links.


\begin{table}[htbp]
\centering
\caption{Summary statistics of the multivariate non-graph datasets}
\small
\setlength{\tabcolsep}{2pt}
\begin{tabular}{c|c|c|c|c|c}
\hline
Dataset     & \begin{tabular}[c]{@{}c@{}}No.\\ time\\  series\\ (N)\end{tabular} & Domain           & Freq.      & \begin{tabular}[c]{@{}c@{}}No.\\  time\\  steps\end{tabular} & \begin{tabular}[c]{@{}c@{}}Prediction \\ length\\ (Q)\end{tabular} \\ \hline
Electricity & 370                                                            &$\mathbb{R}^{+}$                  & Hourly     & 5833                                                      & 24                                                               \\
Traffic     & 963                                                            & $(0, 1)$         & Hourly     & 4001                                                      & 24                                                               \\
Taxi        & 1214                                                           & $\mathbb{N}$     & 30 Minutes & 1488                                                      & 24                                                               \\
Wikipedia   & 2000                                                           & $\mathbb{N}$                 & Daily      & 792                                                       & 30  \\ \hline                
\end{tabular}
\label{tab:non_graph_data}
\end{table}

\section{Definitions of evaluation metrics}
The point forecasts are evaluated by computing mean absolute error (MAE), mean absolute percentage error (MAPE), and root mean squared error (RMSE). For the test-set indexed by $\mathcal{D}_{test}$, let $\mathbf{y}_t^{(m)} \in \mathbb{R}^{N}$ and $\mathbf{\hat{y}}_{t}^{(m)} \in \mathbb{R}^{N}$ denote the ground truth and the prediction at horizon $t$ for $m$-th test example respectively. The average MAE, MAPE, and RMSE at horizon $t$ are defined as follows: 
\begin{align}
\text{MAE}(\mathcal{D}_{test}, t) &= \frac{1}{N \lvert \mathcal{D}_{test} \rvert} \sum_{m \in \mathcal{D}_{test}} \lvert \lvert \mathbf{y}_{t}^{(m)} - \mathbf{\hat{y}}_{t}^{(m)}\rvert \rvert_1\,,\label{eq:mae} \\
\text{MAPE}(\mathcal{D}_{test}, t) &= \frac{1}{N \lvert \mathcal{D}_{test} \rvert} \sum_{m \in \mathcal{D}_{test}} \sum_{i=1}^{N} \frac{\lvert \mathbf{y}_{t,i}^{(m)} - \mathbf{\hat{y}}_{t,i}^{(m)} \rvert}{\lvert \mathbf{y}_{t,i}^{(m)}\rvert}\,,\label{eq:mape} \\
\text{RMSE}(\mathcal{D}_{test}, t) &= \sqrt{\frac{1}{N \lvert \mathcal{D}_{test} \rvert} \sum_{m \in \mathcal{D}_{test}} \lvert \lvert \mathbf{y}_{t}^{(m)} - \mathbf{\hat{y}}_{t}^{(m)}\rvert \rvert_2^2}\,,\label{eq:rmse} 
\end{align}

For comparison among the probabilistic forecasting models, we compute the Continuous Ranked Probability Score (CRPS)~\cite{gneiting2007}, and the P10 and P90 Quantile Losses (QL)~\cite{salinas2020, wang2019}. Let $F(\cdot)$ be the Cumulative Distribution Function (CDF) of the forecast of the true value $x \in \mathbb{R}$. We denote by $\mathbf{1}\{x \leqslant z\}$ the indicator function that attains the value 1 if $x \leqslant z$ and the value 0 otherwise. The continuous ranked probability score (CRPS) is defined as:
\begin{align}
\text{CRPS} (F, x) = \int_{-\infty}^{\infty} \Big(F(z) - \mathbf{1}\{x \leqslant z\} \Big)^2 dz\,.\label{eq:crps}
\end{align}
CRPS is a proper scoring function, i.e., it attains its minimum value of zero when the forecast CDF $F$ is a step function at the ground truth $x$. The average CRPS at horizon $t$ is defined as the average marginal CRPS across different time-series.
\begin{align}
&\text{CRPS}_{avg} (\mathcal{D}_{test}, t) =\,\nonumber\\
&\qquad \frac{1}{N \lvert \mathcal{D}_{test} \rvert} \sum_{m \in \mathcal{D}_{test}} \sum_{i=1}^{N} \text{CRPS}(F_{t,i}^{(m)}, \mathbf{y}_{t,i}^{(m)}) \,,\label{eq:crps_avg}
\end{align}
where $F_{t,i}^{(m)}(\cdot)$ is the marginal CDF of the forecast at horizon $t$ for $i$-th time-series in $m$-th test example.

Let $F_{t, sum}^{(m)}(\cdot)$ is the CDF of the sum of the forecasts of all time-series at horizon $t$ in $m$-th test example. The (normalized) $\text{CRPS}_{sum}$ is defined as:
\begin{align}
&\text{CRPS}_{sum} (\mathcal{D}_{test}) = \,\nonumber\\
&\qquad \frac{\sum_t \sum_{m \in \mathcal{D}_{test}} \text{CRPS}(F_{t, sum}^{(m)}, \sum_{i=1}^N \mathbf{y}_{t,i}^{(m)})}{\sum_t \sum_{m \in \mathcal{D}_{test}} \lvert \sum_{i=1}^N  \mathbf{y}_{t,i}^{(m)}\rvert} \,.\label{eq:crps_sum}
\end{align}

For a given quantile $\alpha \in (0,1)$, a true value $x$, and an $\alpha$-quantile prediction $\hat{x}(\alpha) = F^{-1}(\alpha)$, the $\alpha$-quantile loss is defined as:
\begin{align}
&\text{QL}\big(x, \hat{x}(\alpha)\big) = 2\Bigg(\alpha\big(x -\hat{x}(\alpha)\big)\mathbf{1}\{x  >\hat{x}(\alpha)\} +\,\nonumber\\ 
&\qquad \qquad(1-\alpha)\big(\hat{x}(\alpha)-x\big)\mathbf{1}\{x  \leqslant \hat{x}(\alpha)\}\Bigg)\,.\label{eq:ql}
\end{align}

The average (normalized) quantile loss (QL) is defined as follows:
\begin{align}
\text{QL}_{avg} (\mathcal{D}_{test}, t, \alpha) = \frac{\sum_{m \in \mathcal{D}_{test}} \sum_{i=1}^{N} \text{QL}\big( \mathbf{y}_{t,i}^{(m)}, \mathbf{\hat{y}}_{t,i}^{(m)}(\alpha)\big)} {\sum_{m \in \mathcal{D}_{test}} \sum_{i=1}^{N} \lvert  \mathbf{y}_{t,i}^{(m)} \rvert} \,.\label{eq:ql_avg}
\end{align}
The P10QL metric is obtained by setting $\alpha = 0.1$ in eq.~\eqref{eq:ql_avg}; the P90QL metric corresponds to $\alpha=0.9$ and the ND (P50QL) metric is obtained using $\alpha=0.5$.
\begin{table*}[htbp]
\caption{Average MAE, MAPE and RMSE for PeMSD3 dataset for 15/30/45/60 minutes horizons. The best and the second best results in each column are shown in bold and marked with underline respectively. Lower numbers are better.}
\centering
\begin{tabular}{c|ccc}
\hline
\multirow{2}{*}{Algorithm} & \multicolumn{3}{c}{PeMSD3 (15/ 30/ 45/ 60 min)} \\ \cline{2-4} 
                  &MAE       &MAPE(\%)       &RMSE       \\ \hline \hline
HA                  &31.58       &33.78       &52.39       \\ 
ARIMA                  &17.31/22.12/27.35/32.47
&16.53/20.78/25.66/30.84
&26.80/34.60/42.37/49.98       \\ 
VAR                  &18.59/20.80/23.06/24.86       &19.59/21.81/24.24/26.44       &31.05/33.92/36.93/39.32       \\ 
SVR                  &16.66/20.33/24.33/28.34       &16.07/19.45/23.31/27.57       &25.97/32.19/38.30/44.57       \\
FNN                  &16.87/20.30/23.91/27.74 &19.59/23.67/30.09/35.44
&25.46/30.97/36.27/41.86       \\ 
FC-LSTM                  &19.01/19.46/19.92/20.29
&19.77/20.23/20.82/21.30
&32.96/33.59/34.24/34.83       \\ \hline 

DCRNN                 &14.42/15.87/17.10/18.29 &14.57/15.78/16.87/17.95 &24.33/27.05/28.99/30.76  
\\ 
STGCN               &15.22/17.54/19.74/21.59 &16.22/18.44/20.13/21.88 &26.20/29.10/32.19/34.83    \\ 
ASTGCN                 &17.03/18.50/19.58/20.95      &18.02/19.28/20.18/21.61      &29.04/31.81/33.98/36.37      \\ 
GWN                  &14.63/16.56/18.34/20.08       &\textbf{13.74}/15.24/16.82/18.75       &25.06/28.48/31.11/33.58       \\ 
GMAN                 &14.73/15.44/\underline{16.15}/\underline{16.96} &15.63/16.25/16.99/17.91 &24.48/\underline{25.68}/\underline{26.80}/\underline{27.99}    \\ 
AGCRN                 &\underline{14.20}/\underline{15.34}/16.28/17.38       &13.79/\textbf{14.47}/\textbf{15.14}/\underline{16.25}       &24.75/26.61/28.06/29.61       \\ 
LSGCN                  &14.28/16.08/17.77/19.23       &14.80/16.01/17.15/18.21       &25.88/28.11/30.31/32.37       \\ 
\hline
DeepGLO     &14.79/18.89/19.11/23.53
&14.12/16.92/17.75/21.68
&\underline{22.97}/29.17/30.48/35.64       \\ 
N-BEATS                  &15.57/18.12/20.50/23.03   &15.56/18.05/20.50/23.19       
&24.44/28.69/32.62/36.72  \\ 
FC-GAGA                  &14.68/15.85/16.40/17.04   &15.57/15.88/16.32/17.16      
&24.65/26.85/27.90/28.97      \\ \hline
DeepAR                 &15.84/18.15/20.30/22.64       &16.26/18.42/20.19/22.56       &26.33/29.96/33.12/36.65       \\ 
DeepFactors                  &17.53/20.17/22.78/24.87       &19.22/24.42/29.58/34.43       &27.62/31.83/35.36/37.91       \\ 
MQRNN                  &14.60/16.55/18.34/20.12       &15.17/17.34/18.94/20.66       &25.35/28.77/31.50/34.40       \\ \hline 
AGCGRU+flow &\textbf{13.79}/\textbf{14.84}/\textbf{15.58}/\textbf{16.06} &\underline{14.01}/\underline{14.75}/\underline{15.34}/\textbf{15.80} &\textbf{22.08}/\textbf{24.26}/\textbf{25.55}/\textbf{26.43}   \\\hline
\end{tabular} 
\label{tab:pemsd3}
\end{table*}

\begin{table*}[htbp]
\caption{Average MAE, MAPE and RMSE for PeMSD4 dataset for 15/30/45/60 minutes horizons. The best and the second best results in each column are shown in bold and marked with underline respectively. Lower numbers are better.}
\centering
\begin{tabular}{c|ccc}
\hline
\multirow{2}{*}{Algorithm} & \multicolumn{3}{c}{PeMSD4 (15/ 30/ 45/ 60 min)} \\ \cline{2-4} 
                  &MAE       &MAPE(\%)       &RMSE       \\ \hline \hline
HA                  &3.16       &7.00       &6.13       \\ 
ARIMA                  &1.53/2.01/2.37/2.68 &2.92/4.06/4.96/5.73
&3.11/4.36/5.25/5.95       \\
VAR                  &1.66/2.12/2.39/2.57       &3.27/4.33/4.95/5.36       &3.09/4.02/4.51/4.83       \\ 
SVR                  &1.48/1.91/2.23/2.49       &2.88/3.97/4.86/5.61       &3.11/4.29/5.08/5.66       \\
FNN                  &1.48/1.90/2.23/2.51   &3.04/4.09/4.98/5.80
&3.08/4.27/5.08/5.68       \\
FC-LSTM                  &2.20/2.22/2.23/2.26
&4.95/4.97/4.99/5.05
&4.89/4.92/4.95/5.01 \\ \hline 
DCRNN                 &1.38/1.78/2.06/2.29       &2.69/3.72/4.51/5.16 &2.95/4.09/4.81/5.34
\\  
STGCN       &1.42/1.85/2.14/2.39 &2.82/3.92/4.71/5.34 &2.94/4.03/4.70/5.21    \\
ASTGCN             &1.69/2.15/2.40/2.55 &3.70/4.85/5.46/5.79 
&3.54/4.71/5.35/5.62      \\ 
GWN                  &\underline{1.37}/1.76/2.03/2.24       &\textbf{2.67}/3.73/4.52/5.15       &2.94/4.07/4.77/5.28       \\ 
GMAN               &1.38/\textbf{1.61}/\textbf{1.76}/\textbf{1.88}       &2.80/\textbf{3.42}/\textbf{3.84}/\underline{4.18}       &2.98/\textbf{3.70}/\textbf{4.11}/\textbf{4.41}   \\
AGCRN                  &1.41/1.67/1.84/\underline{2.01}       &2.88/3.55/3.99/4.40       &3.04/3.83/4.33/4.73       \\ 
LSGCN                  &1.40/1.78/2.03/2.20       &2.80/3.71/4.27/4.68       &\textbf{2.87}/3.90/4.50/4.89       \\ 
\hline 
DeepGLO      &1.61/1.89/2.25/2.51 
&3.13/4.06/5.03/5.77
&3.06/4.14/4.92/5.55       \\ 
N-BEATS                  &1.49/1.90/2.20/2.44   &2.93/4.00/4.84/5.48       &3.13/4.29/5.05/5.58      \\ 
FC-GAGA                  &1.43/1.78/1.95/2.06   
&2.87/3.80/4.32/4.67     &3.06/4.09/4.55/4.82    \\ \hline
DeepAR                 &1.51/2.01/2.38/2.68       &3.06/4.41/5.45/6.25       &3.11/4.27/5.04/5.60       \\ 
DeepFactors                  &1.54/2.01/2.34/2.61       &3.07/4.26/5.17/5.90       &3.11/4.21/4.90/5.40       \\ 
MQRNN                  &\underline{1.37}/1.76/2.03/2.25       &\underline{2.68}/3.72/4.51/5.17       &2.94/4.05/4.73/5.20       \\ \hline 
AGCGRU+flow &\textbf{1.35}/\underline{1.63}/\underline{1.78}/\textbf{1.88} &\textbf{2.67}/\underline{3.44}/\underline{3.87}/\textbf{4.16} &\underline{2.88}/\underline{3.77}/\underline{4.20}/\underline{4.46}   \\ \hline
\end{tabular}
\label{tab:pemsd4}
\end{table*}

\begin{table*}[htbp]
\caption{Average MAE, MAPE and RMSE for PeMSD7 dataset for 15/30/45/60 minutes horizons. The best and the second best results in each column are shown in bold and marked with underline respectively. Lower numbers are better.}
\centering
\begin{tabular}{c|ccc}
\hline
\multirow{2}{*}{Algorithm} & \multicolumn{3}{c}{PeMSD7 (15/ 30/ 45/ 60 min)} \\ \cline{2-4} 
                  &MAE       &MAPE(\%)       &RMSE       \\ \hline \hline
HA                  &3.98       &10.92       &7.20       \\ 
ARIMA                  &2.49/3.52/4.32/5.03 &5.66/8.30/10.46/12.35
&4.53/6.64/8.17/9.42       \\ 
VAR                  &2.70/3.71/4.37/4.87       &6.23/8.75/10.37/11.56       &4.38/5.95/6.89/7.56       \\ 
SVR                  &2.43/3.40/4.15/4.78       &5.62/8.23/10.38/12.31       &4.52/6.53/7.93/9.02       \\ 
FNN                  &2.36/3.32/4.06/4.71   &5.56/8.20/10.41/12.44
&4.45/6.46/7.84/8.90       \\ 
FC-LSTM                  &3.55/3.59/3.64/3.70
&9.12/9.17/9.25/9.37
&6.83/6.91/6.99/7.11       \\ \hline
DCRNN           &2.23/3.06/3.67/4.18 &\underline{5.19}/7.50/9.31/10.90 &4.26/6.05/7.28/8.24      \\ 
STGCN           &2.21/2.96/3.47/3.90 &5.20/7.32/8.82/10.09 &\underline{4.09}/5.72/6.76/7.55   \\ 
ASTGCN          &2.71/3.72/4.28/4.60 &6.68/9.51/11.06/11.86 &4.64/6.53/7.60/8.13    \\
GWN                  &2.23/3.03/3.56/3.98       &5.26/7.63/9.25/10.56       &4.27/5.99/7.03/7.76       \\ 
GMAN                 &2.40/\underline{2.76}/\textbf{2.98}/\textbf{3.16}       &5.93/\underline{6.96}/\underline{7.66}/\textbf{8.16}       &4.74/5.57/\textbf{6.06}/\textbf{6.37}       \\ 
AGCRN                  &\underline{2.19}/2.81/3.15/3.42       &5.22/7.09/8.19/9.01       &4.12/\underline{5.49}/6.27/6.79       \\
LSGCN                  &2.23/2.99/3.50/3.95       &5.22/7.18/8.40/9.37       &\textbf{4.03}/5.59/6.54/7.30       \\ 
\hline
DeepGLO     &2.55/3.32/4.16/4.85
&6.10/8.31/11.16/13.19
&4.53/6.30/7.68/8.84       \\
N-BEATS                  &2.44/3.34/4.02/4.57    
&5.75/8.30/10.31/11.94      &4.55/6.51/7.84/8.80     \\ 
FC-GAGA                  &2.22/2.85/3.18/3.36     
&5.32/7.09/8.00/8.51    &4.29/5.77/6.46/6.82   \\ \hline
DeepAR                 &2.53/3.61/4.48/5.20       &6.15/9.30/12.17/14.49       &4.55/6.50/7.84/8.87       \\ 
DeepFactors                  &2.51/3.47/4.17/4.71       &6.14/9.04/11.21/12.93       &4.47/6.21/7.30/8.08       \\
MQRNN                  &2.22/3.03/3.58/4.00       &5.26/7.70/9.53/10.97       &4.23/5.91/6.98/7.73       \\ \hline
AGCGRU+flow &\textbf{2.15}/\textbf{2.70}/\underline{2.99}/\underline{3.19} &\textbf{5.13}/\textbf{6.75}/\textbf{7.61}/\underline{8.18} &4.11/\textbf{5.46}/\underline{6.12}/\underline{6.54}     \\ \hline 
\end{tabular}
\label{tab:pemsd7}
\end{table*}

\begin{table*}[htbp]
\caption{Average MAE, MAPE and RMSE for PeMSD8 dataset for 15/30/45/60 minutes horizons. The best and the second best results in each column are shown in bold and marked with underline respectively. Lower numbers are better.}
\centering
\begin{tabular}{c|ccc}
\hline
\multirow{2}{*}{Algorithm} & \multicolumn{3}{c}{PeMSD8 (15/ 30/ 45/ 60 min)} \\ \cline{2-4} 
                  &MAE       &MAPE(\%)       &RMSE       \\ \hline \hline
HA                  &2.47       &5.66       &5.19       \\ 
ARIMA                  &1.24/1.61/1.89/2.12 &2.33/3.15/3.77/4.31
&2.63/3.62/4.28/4.81       \\ 
VAR                  &1.37/1.79/2.04/2.23       &2.66/3.62/4.23/4.69       &2.67/3.53/4.01/4.36       \\ 
SVR                  &1.21/1.56/1.80/2.01       &2.32/3.12/3.72/4.24       &2.64/3.57/4.18/4.63       \\ 
FNN                  &1.19/1.54/1.79/2.01
&2.27/3.12/3.75/4.30
&2.59/3.55/4.17/4.63       \\ 
FC-LSTM                  &1.91/1.93/1.94/1.95       &4.63/4.66/4.69/4.72       
&4.71/4.75/4.78/4.81       \\ \hline 
DCRNN       &1.16/1.49/1.70/1.87 &2.25/3.16/3.85/4.37 &2.54/3.49/4.08/4.49        \\
STGCN           &1.22/1.56/1.79/1.98 &2.49/3.43/4.06/4.48 &2.67/3.65/4.22/4.59      \\ 
ASTGCN                 &1.36/1.64/1.81/1.92       &3.04/3.79/4.23/4.51      &2.98/3.77/4.20/4.47      \\ 
GWN                  &\textbf{1.11}/1.40/1.59/1.73       &\textbf{2.14}/\textbf{2.94}/3.49/3.90       &\textbf{2.52}/\underline{3.45}/4.00/4.38       \\
GMAN                 &1.23/\textbf{1.36}/\textbf{1.46}/\textbf{1.55}       &2.73/3.09/\textbf{3.38}/\textbf{3.63}       &3.05/3.50/\textbf{3.82}/\textbf{4.06}       \\ 
AGCRN                  &1.16/1.39/1.53/1.67       &2.49/3.10/3.50/3.84       &2.67/\textbf{3.44}/3.91/4.25       \\ 
LSGCN                  &1.21/1.54/1.75/1.89       &2.56/3.44/3.95/4.30       &2.71/3.64/4.14/4.46       \\ 
\hline
DeepGLO     &1.30/1.75/2.04/2.21
&2.48/3.42/4.06/4.50
&2.67/3.63/4.24/4.69       \\ 
N-BEATS                  &1.33/1.69/1.92/2.12  
&2.74/3.85/4.45/4.90   &2.81/3.94/4.52/4.92      \\ 
FC-GAGA                  &1.18/1.47/1.62/1.72    
&2.37/3.21/3.76/4.11     &2.65/3.61/4.10/4.39    \\ \hline
DeepAR                 &1.25/1.61/1.87/2.10       &2.53/3.40/4.08/4.67       &2.67/3.59/4.17/4.61       \\ 
DeepFactors                  &1.26/1.63/1.88/2.07       &2.51/3.42/4.08/4.61       &2.63/3.54/4.11/4.52       \\ 
MQRNN                  &\underline{1.13}/1.43/1.62/1.77       &\underline{2.19}/2.99/3.56/4.00       &\underline{2.54}/3.48/4.02/4.40       \\ \hline 
AGCGRU+flow &\underline{1.13}/\underline{1.37}/\underline{1.49}/\underline{1.57} &2.30/\underline{3.01}/\underline{3.40}/\underline{3.65} 
&2.59/\underline{3.45}/\underline{3.85}/\underline{4.09}  \\ \hline
\end{tabular}
\label{tab:pemsd8}
\end{table*}

\section{Detailed experimental results on the PeMS datasets}
\subsection{Baseline algorithms}
For the experiments on the PeMS road traffic datasets, we compare the proposed AGCGRU+flow algorithm with four different classes of forecasting techniques, listed as follows:

\textit{Graph agnostic statistical and  machine learning based point forecasting models}:
\begin{itemize}[leftmargin=*]
\item HA (Historical Average): uses the seasonality of the historical data. \item ARIMA~\cite{makridakis1997}: implemented using a Kalman filter.
\item Vector Auto-Regressive model (VAR)~\cite{hamilton1994}: generalization of AR model to multivariate setting. 
\item Support Vector Regression (SVR)~\cite{wu2004}
\item FNN (Feedforward Neural Network).
\item FC-LSTM~\cite{sutskever2014}: encoder-decoder architecture for sequence to sequence prediction using fully connected LSTM layers.
\end{itemize}
\textit{Spatio-temporal point forecast models}:
\begin{itemize}[leftmargin=*]
\item DCRNN~\cite{li2018diffusion}: Diffusion Convolutional Recurrent Neural Network, combines diffusion convolution with GRU to form an encoder-decoder architecture for sequence to sequence prediction.
\item STGCN~\cite{yu2018}: Spatio-Temporal Graph Convolutional Network, uses gated temporal convolution with graph convolution. 
\item ASTGCN~\cite{guo2019}: Attention Spatial-Temporal Graph Convolutional Network, spatial and temporal attentions to learn recent and seasonal patterns.
\item GWN~\cite{yu2019}: Graph WaveNet, built using graph convolution and dilated causal convolution, provision for learnable graph.
\item GMAN~\cite{zheng2020}: Graph Multi-Attention
Network, multiple spatio-temporal attention blocks to form an encoder-decoder architecture, transform attention between encoder and decoder. 
\item AGCRN~\cite{bai2020}: Adaptive Graph Convolutional Recurrent Network, node adaptive parameter learning for graph convolution using adaptive adjacency, combined with GRU. 
\item LSGCN~\cite{huang2020}: Long Shortterm Graph Convolutional Network, a novel attention mechanism and graph convolution, integrated into a spatial gated block. 
\end{itemize}

\textit{Deep-learning based point forecasting methods}:
\begin{itemize}[leftmargin=*]
\item DeepGLO~\cite{sen2019}: global matrix factorization, regularization using temporal convolution.
\item N-BEATS~\cite{oreshkin2020}: Neural Basis Expansion Analysis for Interpretable Time-Series, an univariate model, built using backward and forward residual connections and deep stack of fully-connected layers.  
\item FC-GAGA~\cite{oreshkin2021}: Fully Connected GAted Graph Architecture, fully connected hard graph gating combined with N-BEATS. 
\end{itemize}

\textit{Deep-learning based probabilistic forecasting methods}: 
\begin{itemize}[leftmargin=*]
\item DeepAR~\cite{salinas2020}: RNN based probabilistic method using parametric likelihood for forecasts.
\item DeepFactors~\cite{wang2019}: global deep learning component along with a local classical
model to account for uncertainty.
\item MQRNN~\cite{wen2017}: RNN based multiple quantile regression.   
\end{itemize}

\subsubsection{Detailed comparisons with baselines for  the PeMS datasets}
\begin{figure*}[htbp]
\begin{minipage}{0.48\textwidth}
  \includegraphics[width=\linewidth]{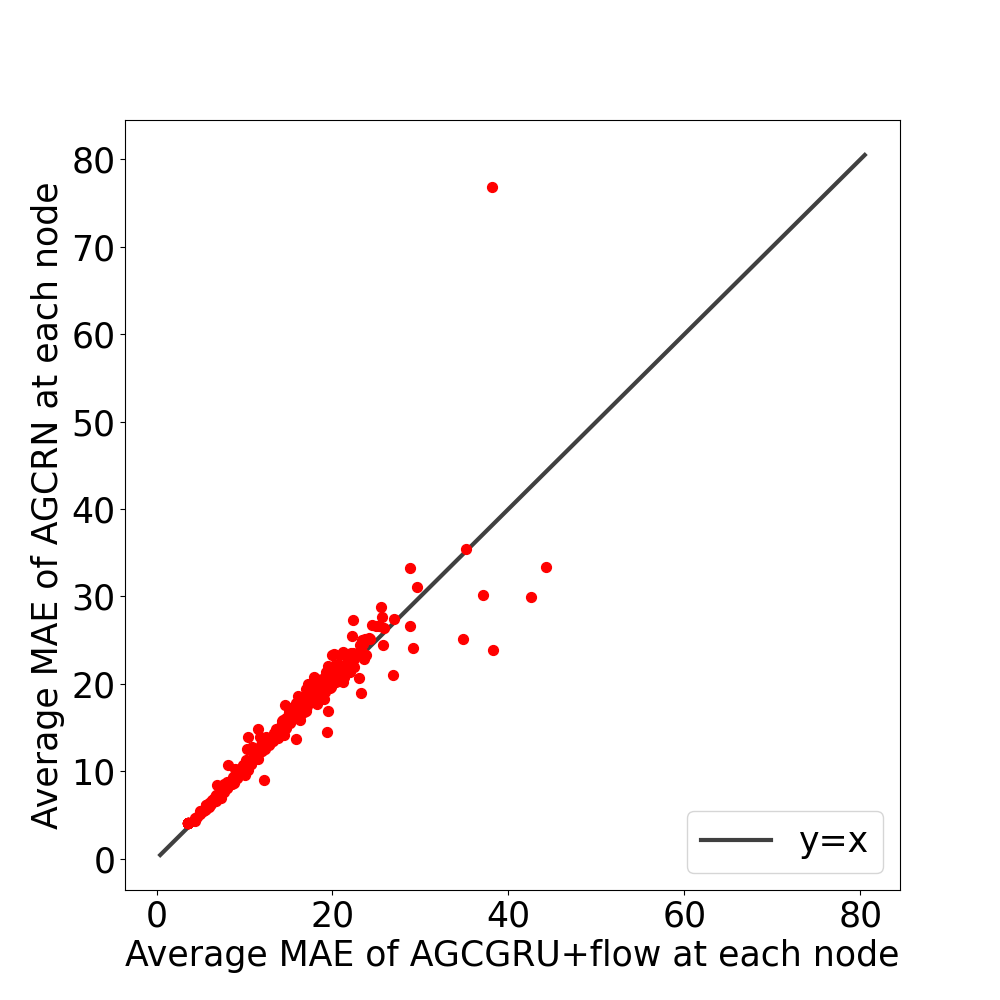}
  \subcaption{PeMSD3}
\end{minipage}%
\begin{minipage}{0.48\textwidth}
  \includegraphics[width=\linewidth]{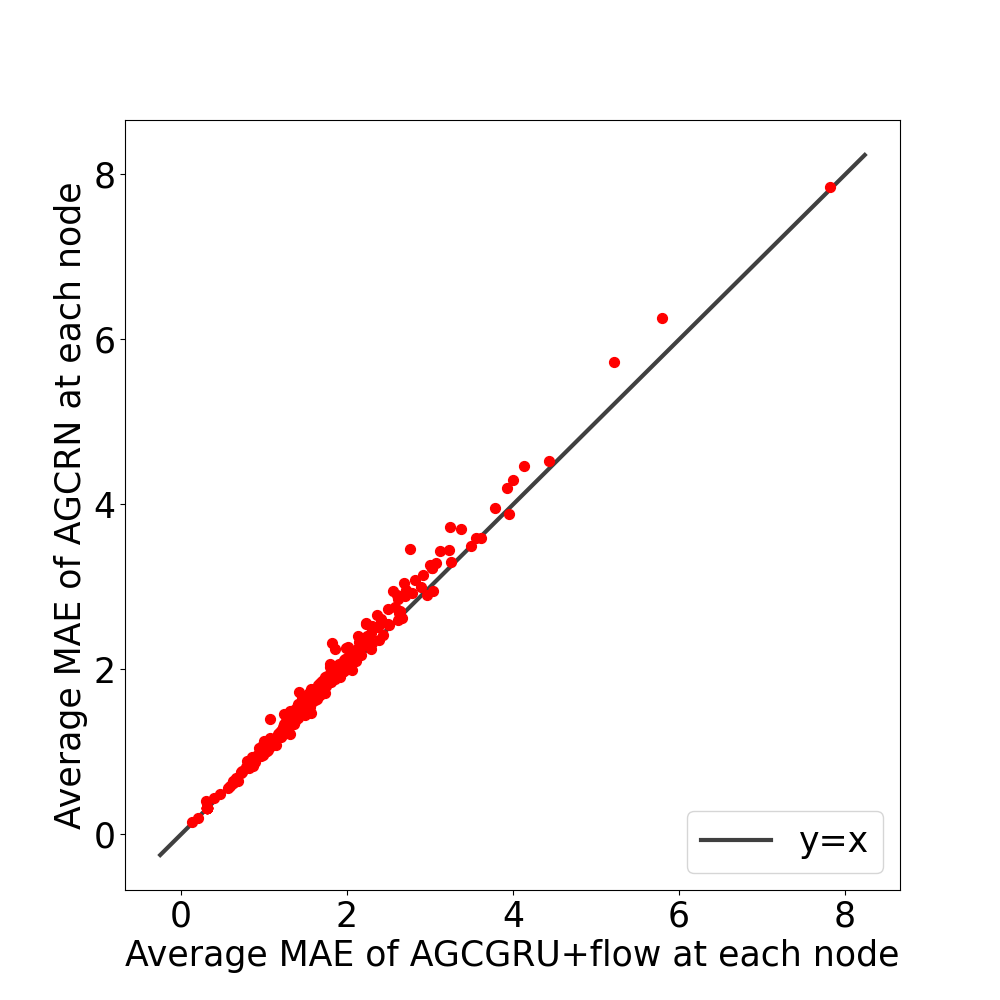}
  \subcaption{PeMSD4}
\end{minipage}  
\begin{minipage}{0.48\textwidth}
  \includegraphics[width=\linewidth]{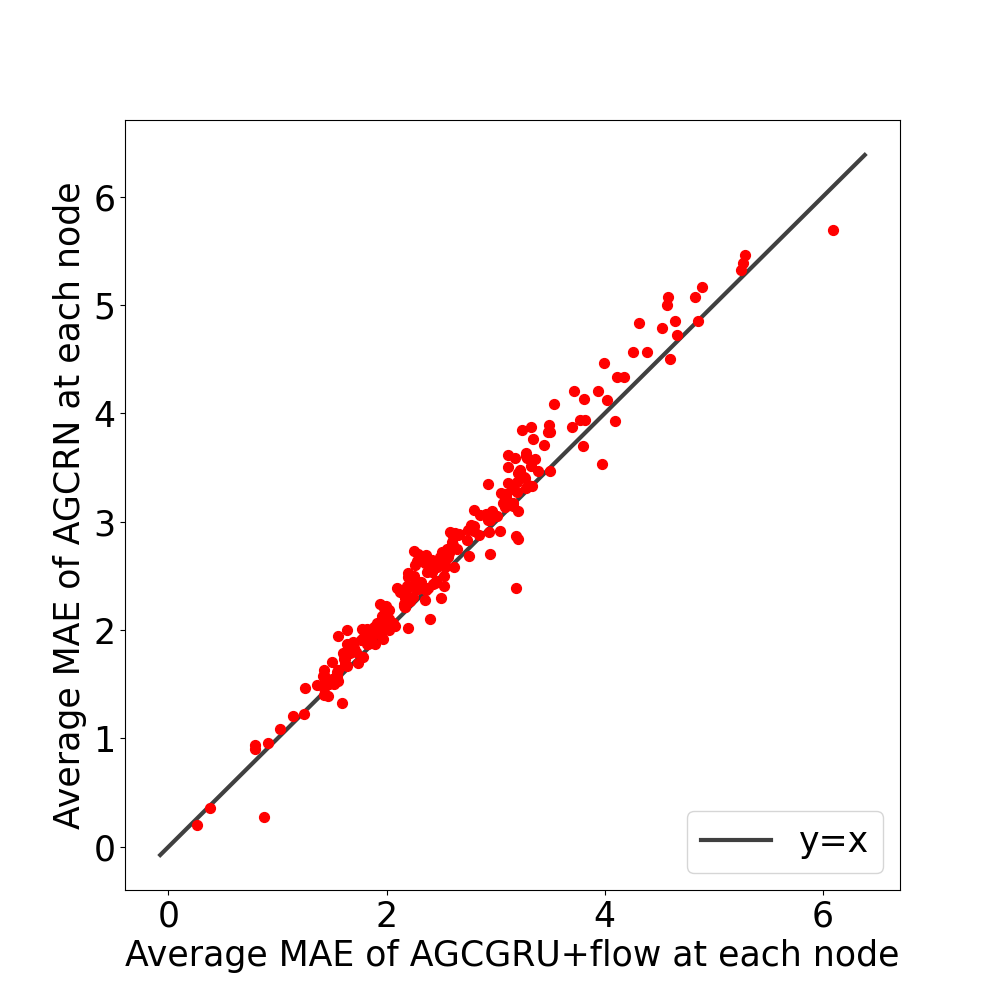}
  \subcaption{PeMSD7}
\end{minipage} %
\begin{minipage}{0.48\textwidth}
  \includegraphics[width=\linewidth]{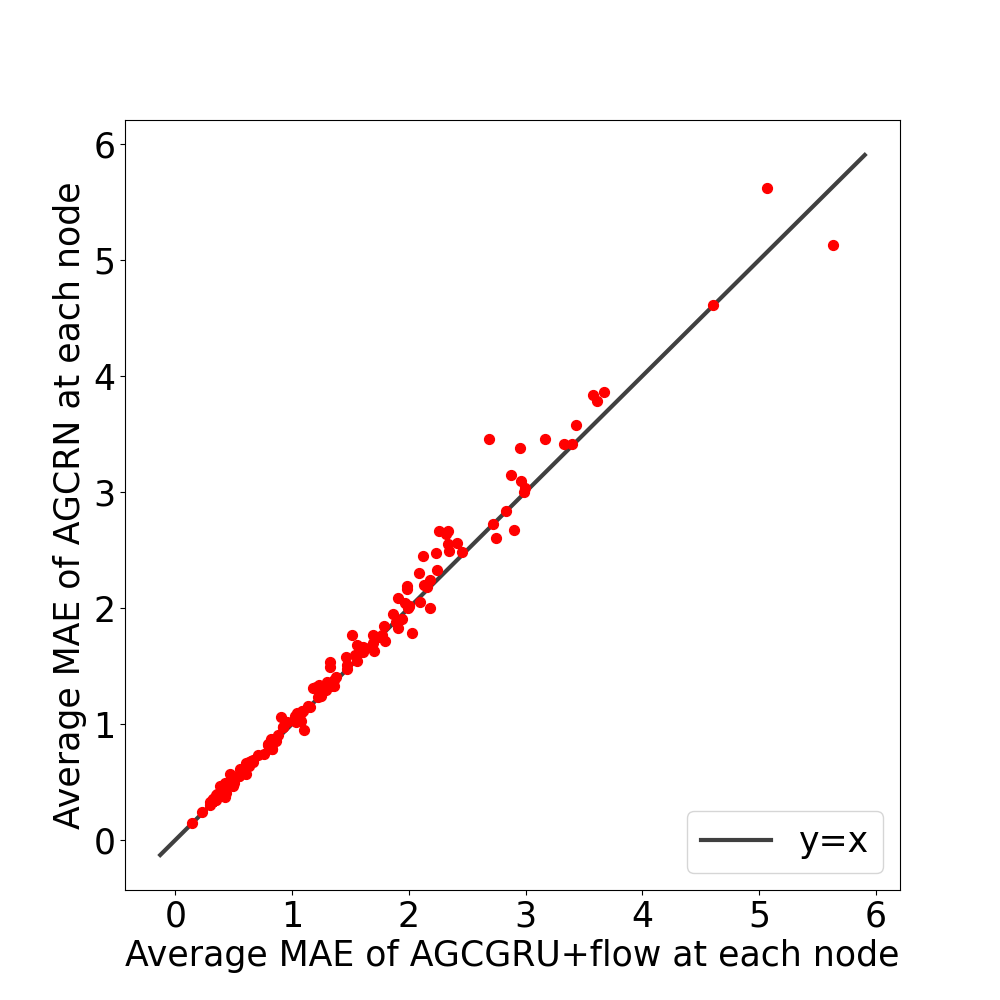}
  \subcaption{PeMSD8}
 \end{minipage}
 \caption{Scatter-plots of average MAE at each node for AGCGRU+flow v.s. that of AGCRN on PeMS datasets. The AGCGRU+flow has lower average MAE compared to AGCRN at most of the nodes for all four datasets.}
 \label{fig:mae_comp}
\end{figure*}

\begin{figure*}[htbp]
\centering
\includegraphics[trim=5cm 0.1cm 3cm 2.0cm, clip, height=5.5cm]{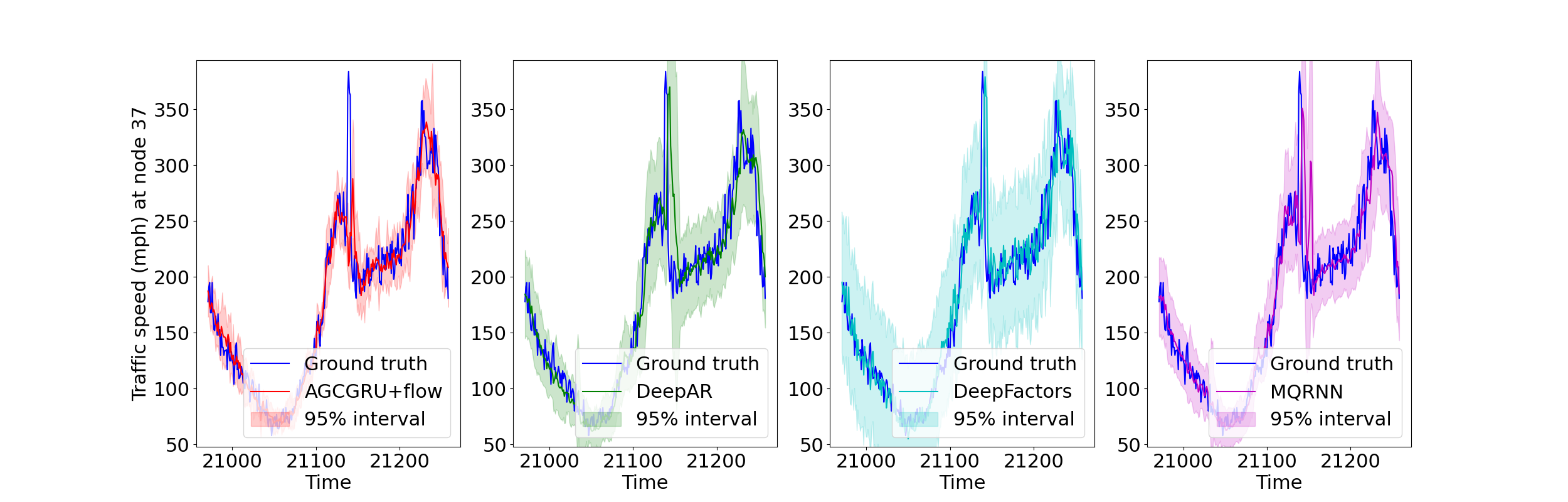}
\includegraphics[trim=5cm 0.1cm 3cm 2.0cm, clip, height=5.5cm]{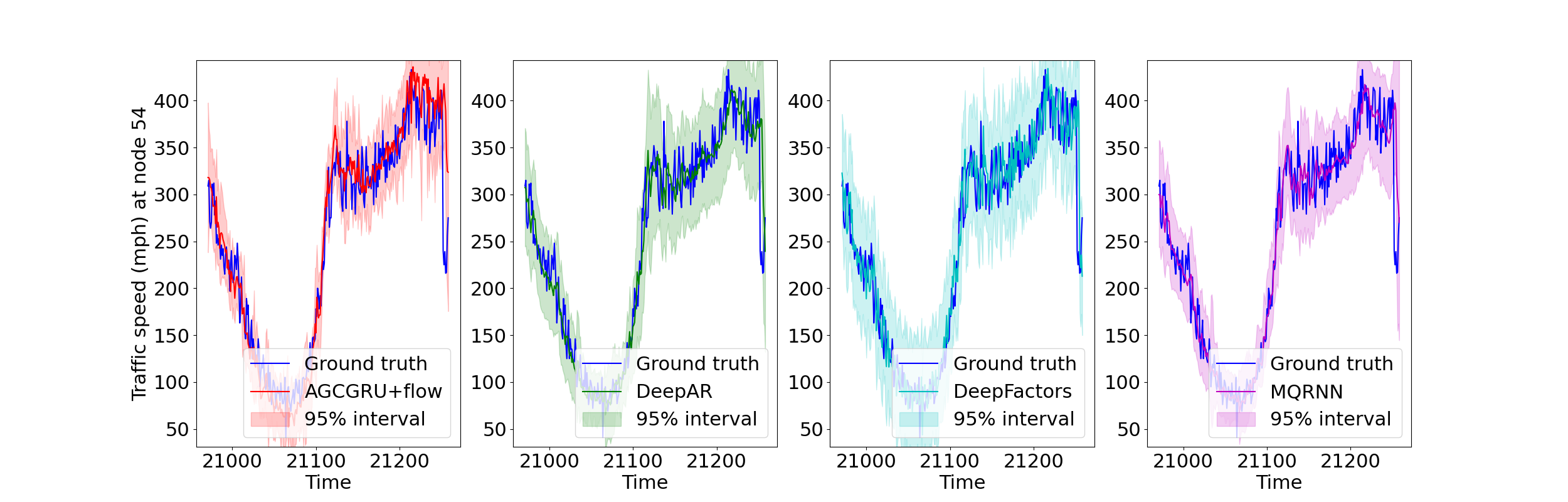}
\includegraphics[trim=5cm 0.1cm 3cm 2.0cm, clip, height=5.5cm]{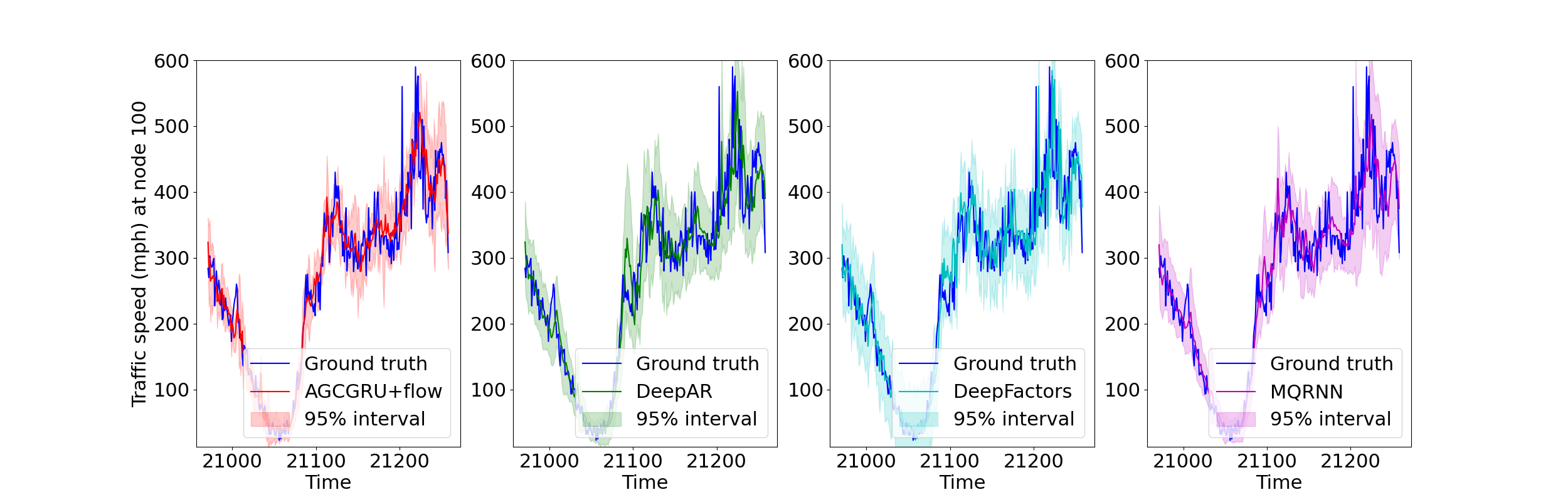}
\includegraphics[trim=5cm 0.1cm 3cm 2.0cm, clip, height=5.5cm]{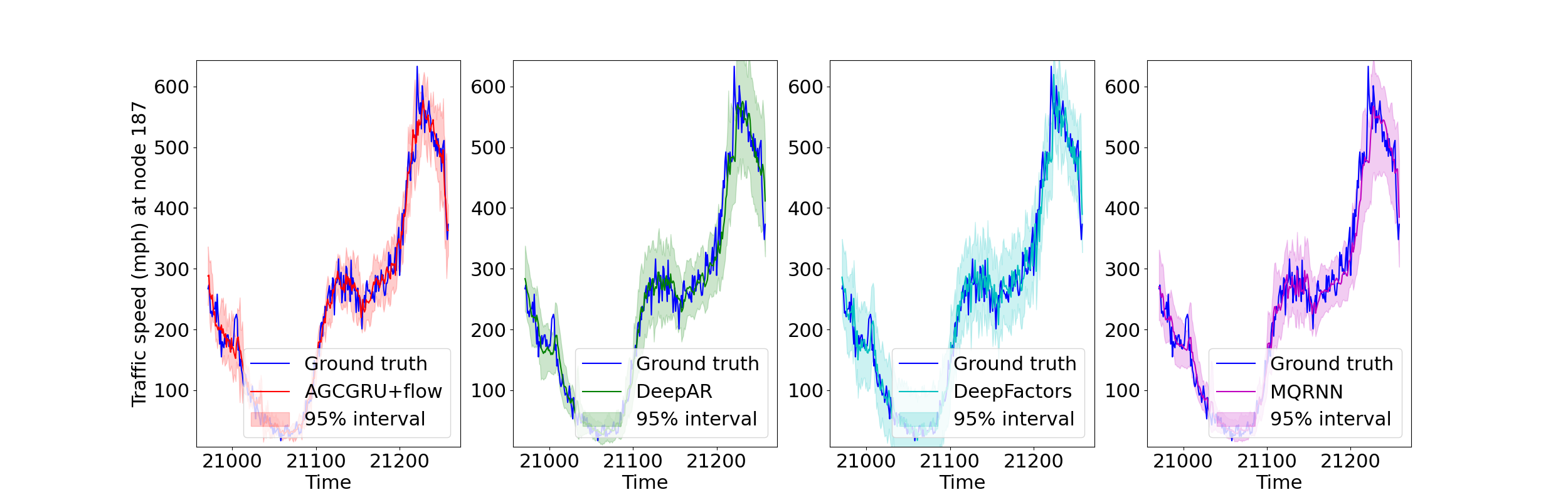}
\vspace{-0.35cm}
\caption{15 minutes ahead predictions from the probabilistic forecasting algorithms with confidence intervals at nodes 37, 54, 100, and 187 of PeMSD3 dataset for the first day in the test set. The proposed AGCGRU+flow algorithm provides tighter confidence interval than its competitors in most cases, which leads to lower quantile error.}
\label{fig:pems03}
\vspace{-0.45cm}
\end{figure*}

\begin{figure*}[htbp]
\centering
\includegraphics[trim=5cm 0.1cm 3cm 2.0cm, clip, height=5.5cm]{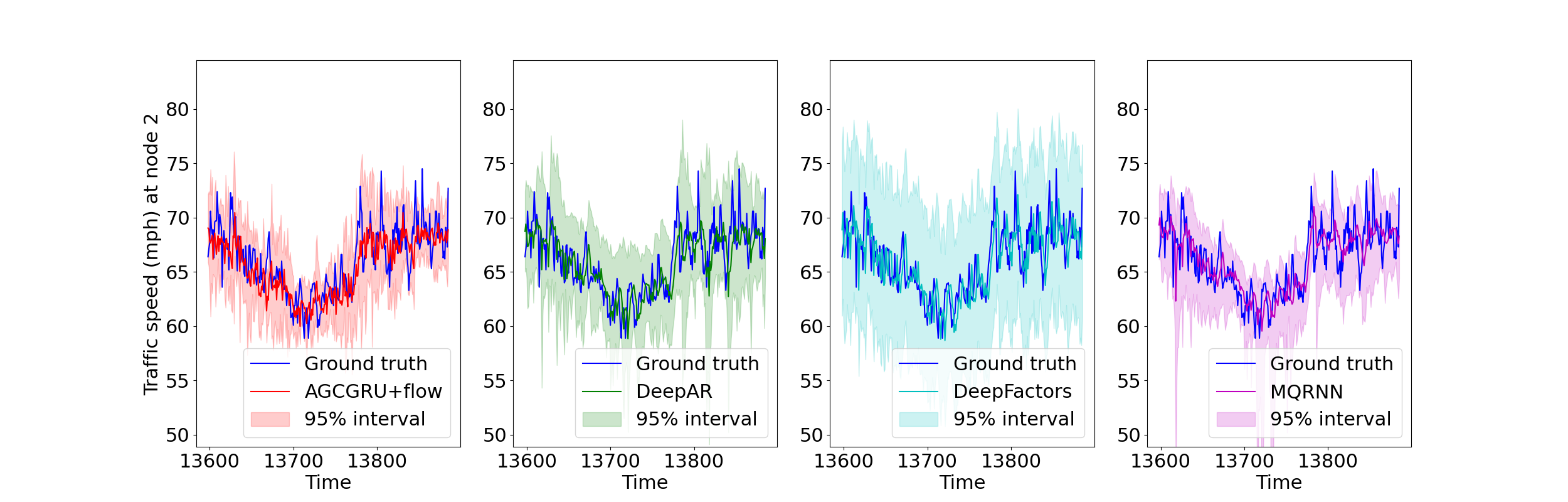}
\includegraphics[trim=5cm 0.1cm 3cm 2.0cm, clip, height=5.5cm]{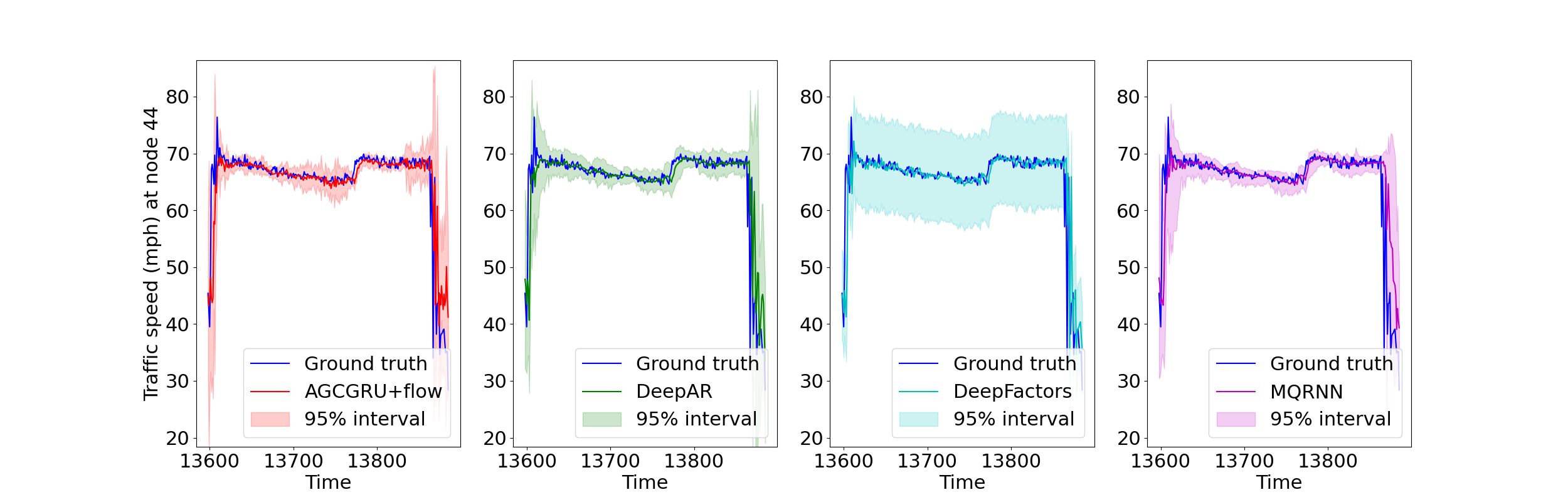}
\includegraphics[trim=5cm 0.1cm 3cm 2.0cm, clip, height=5.5cm]{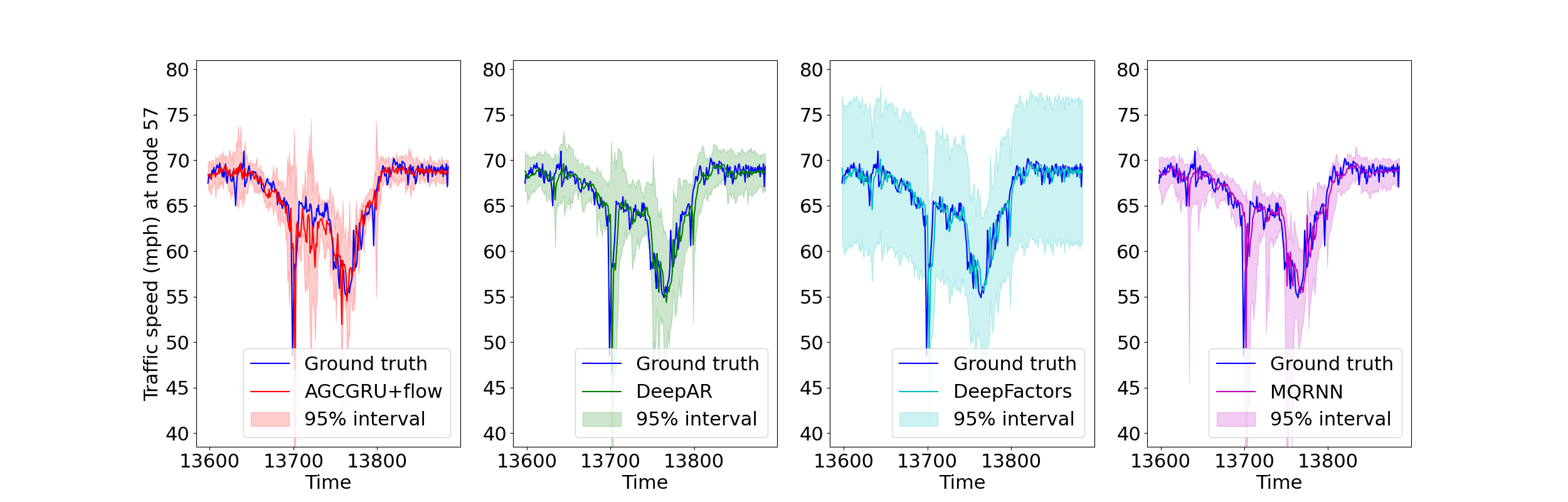}
\includegraphics[trim=5cm 0.1cm 3cm 2.0cm, clip, height=5.5cm]{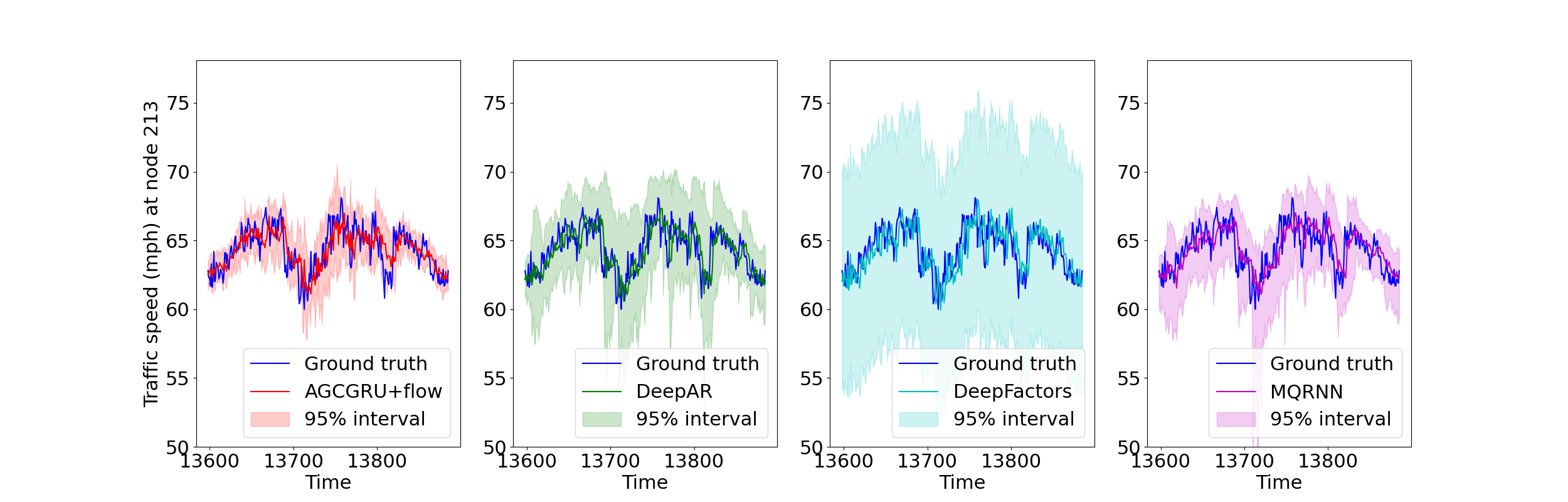}
\vspace{-0.35cm}
\caption{15 minutes ahead predictions from the probabilistic forecasting algorithms with confidence intervals at nodes 2, 44, 57, and 213 of PeMSD4 dataset for the first day in the test set. The proposed AGCGRU+flow algorithm provides tighter confidence interval than its competitors in most cases, which leads to lower quantile error.}
\label{fig:pems04}
\vspace{-0.45cm}
\end{figure*}

\begin{figure*}[htbp]
\centering
\includegraphics[trim=5cm 0.1cm 3cm 2.0cm, clip, height=5.5cm]{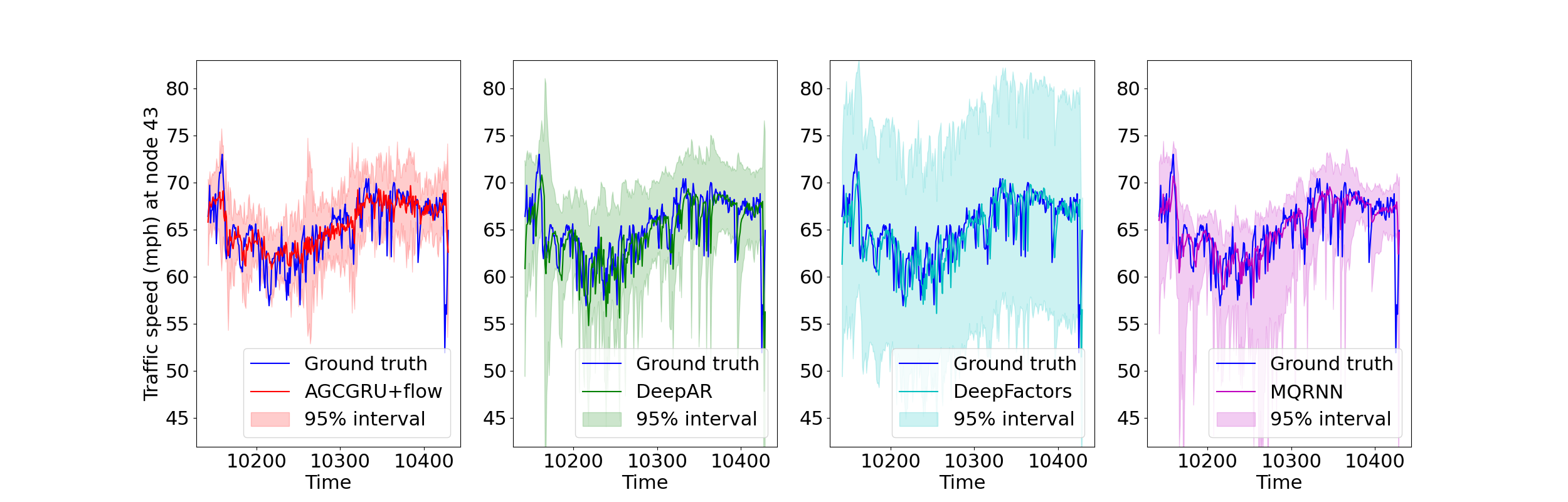}
\includegraphics[trim=5cm 0.1cm 3cm 2.0cm, clip, height=5.5cm]{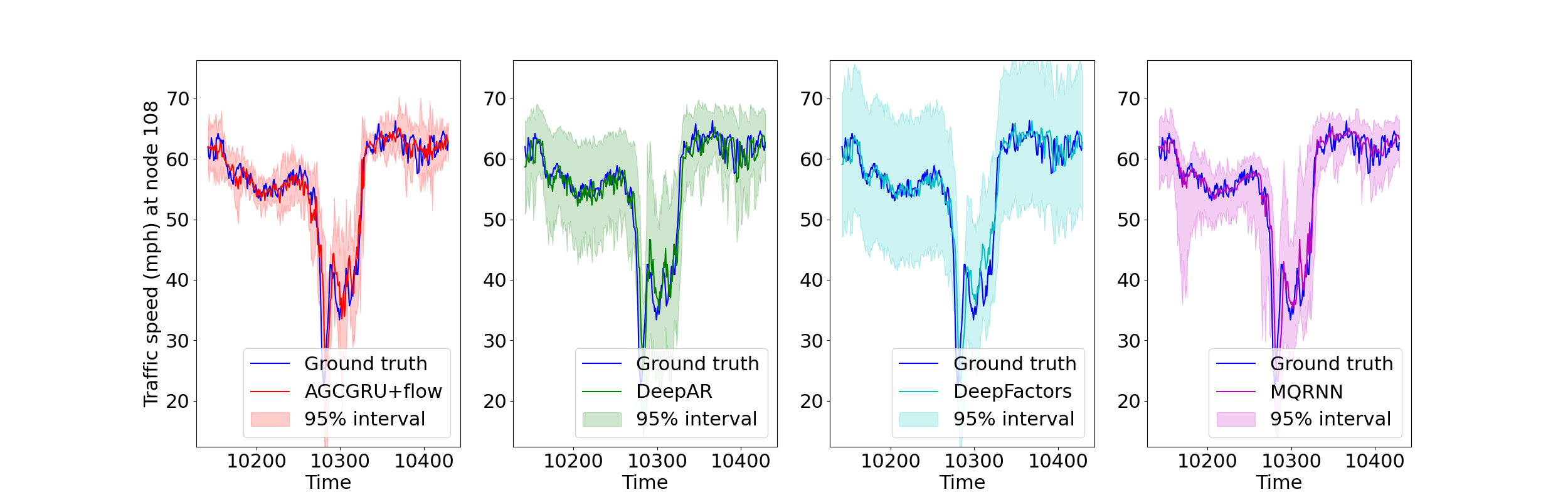}
\includegraphics[trim=5cm 0.1cm 3cm 2.0cm, clip, height=5.5cm]{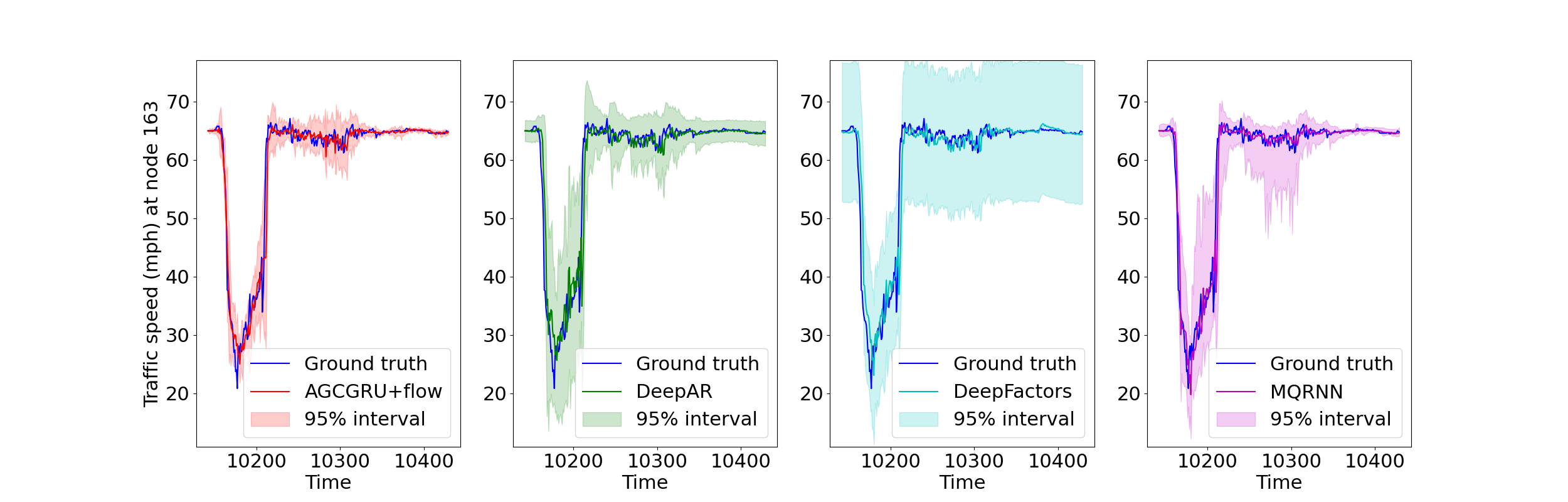}
\includegraphics[trim=5cm 0.1cm 3cm 2.0cm, clip, height=5.5cm]{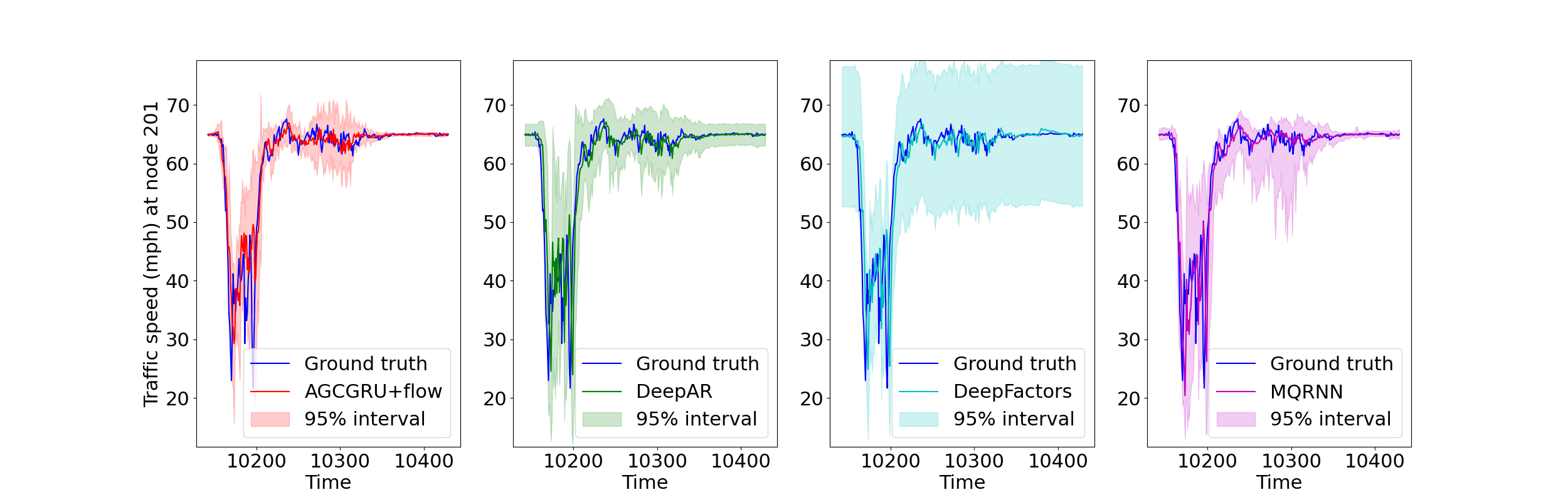}
\vspace{-0.35cm}
\caption{15 minutes ahead predictions from the probabilistic forecasting algorithms with confidence intervals at nodes 43, 108, 163, and 201 of PeMSD7 dataset for the first day in the test set. The proposed AGCGRU+flow algorithm provides tighter confidence interval than its competitors in most cases, which leads to lower quantile error.}
\label{fig:pems07}
\vspace{-0.45cm}
\end{figure*}

\begin{figure*}[htbp]
\centering
\includegraphics[trim=5cm 0.1cm 3cm 2.0cm, clip, height=5.5cm]{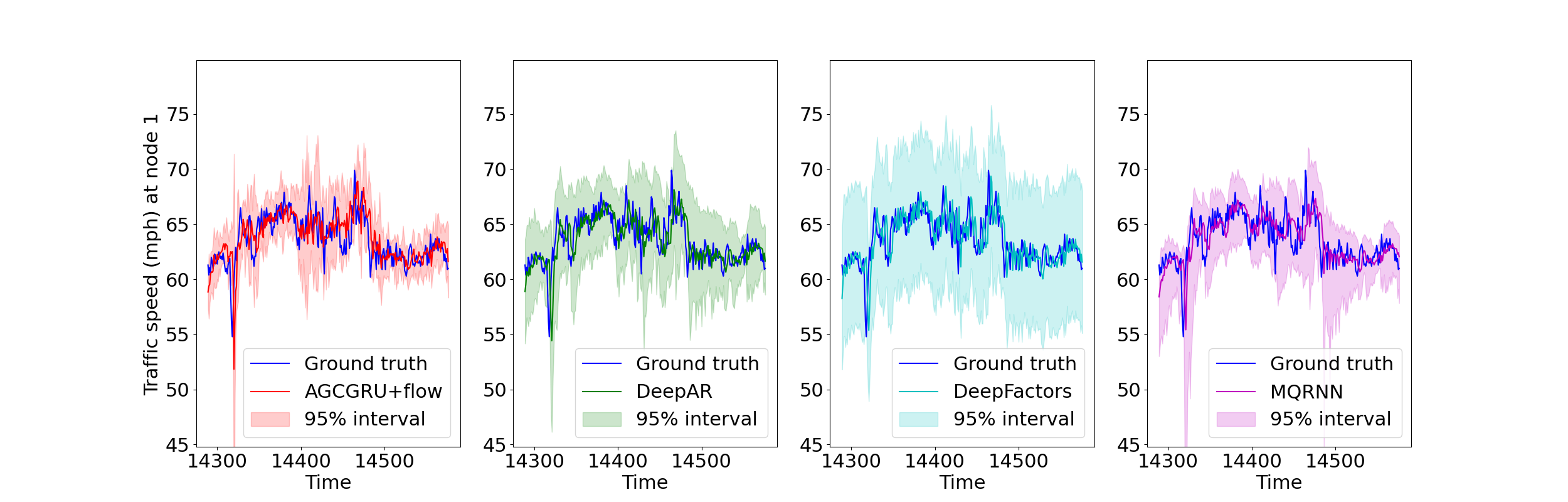}
\includegraphics[trim=5cm 0.1cm 3cm 2.0cm, clip, height=5.5cm]{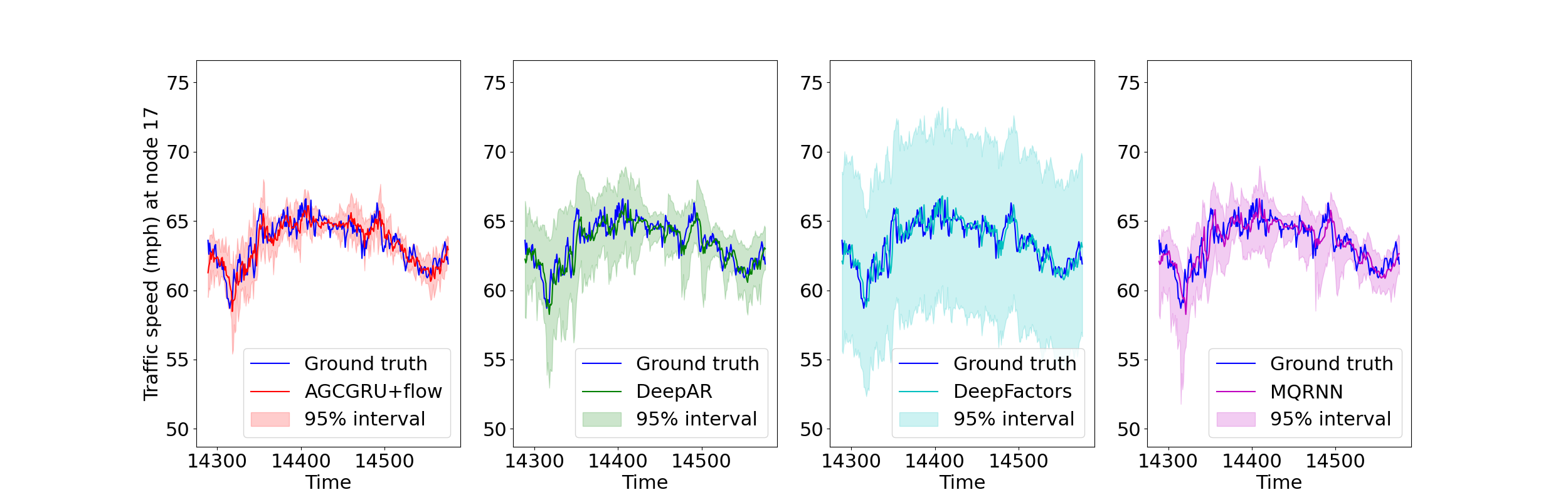}
\includegraphics[trim=5cm 0.1cm 3cm 2.0cm, clip, height=5.5cm]{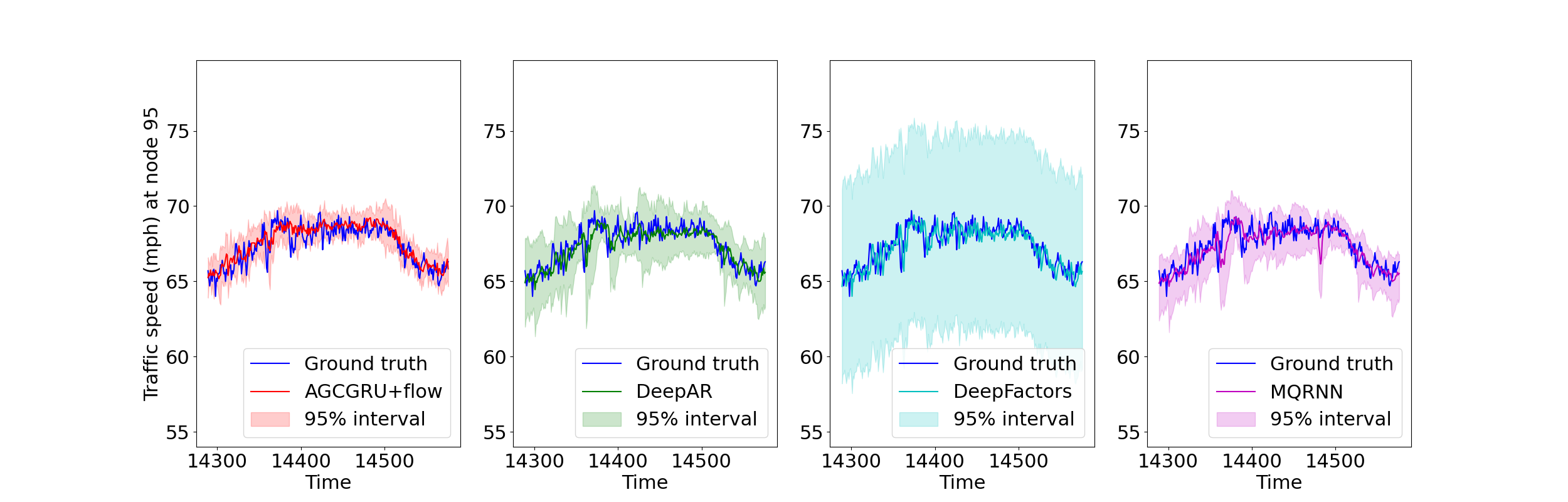}
\includegraphics[trim=5cm 0.1cm 3cm 2.0cm, clip, height=5.5cm]{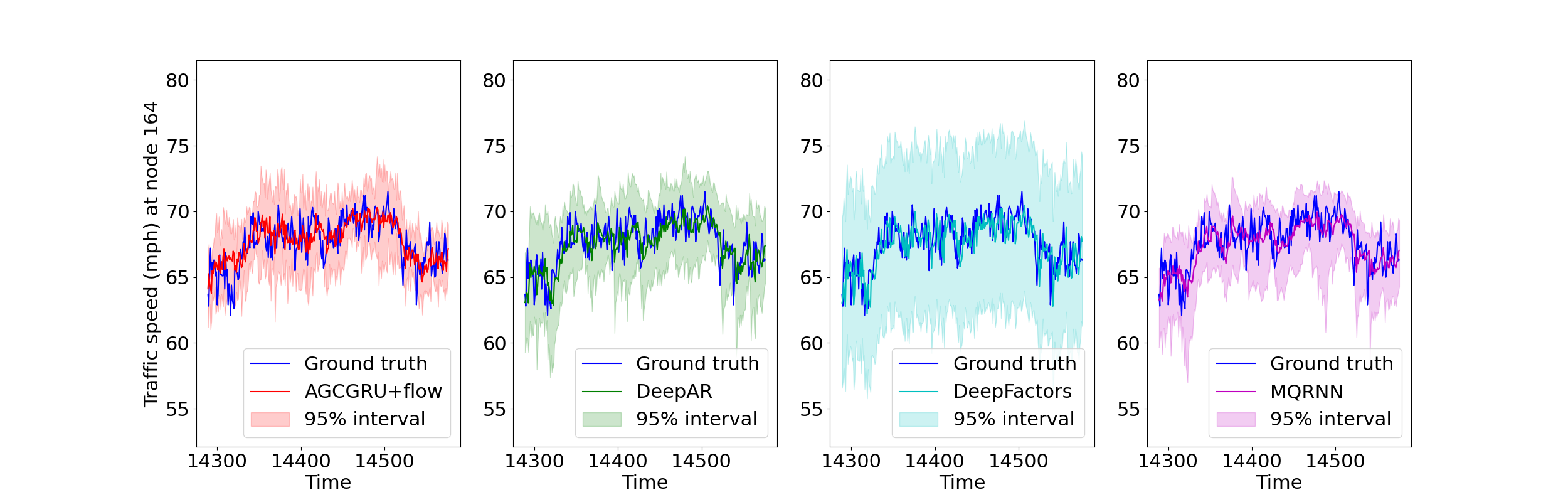}
\vspace{-0.35cm}
\caption{15 minutes ahead predictions from the probabilistic forecasting algorithms with confidence intervals at nodes 1, 17, 95, and 164 of PeMSD8 dataset for the first day in the test set. The proposed AGCGRU+flow algorithm provides tighter confidence interval than its competitors in most cases, which leads to lower quantile error.}
\label{fig:pems08}
\vspace{-0.45cm}
\end{figure*}

In Table~\ref{tab:mae} of the main paper, we report the average MAE of the top 10 algorithms. The detailed comparisons in terms of MAE, MAPE, and RMSE with all the baseline algorithms on the four PeMS datasets are provided in Tables~\ref{tab:pemsd3}, \ref{tab:pemsd4}, \ref{tab:pemsd7}, and \ref{tab:pemsd8}. We observe that statistical models such as HA, ARIMA, and VAR and basic machine learning models such as SVR, FNN, and FC-LSTM show poor predictive performance as they cannot model the complex spatio-temporal patterns present in the real world traffic data well. Graph agnostic deep learning models such as DeepGLO and N-BEATS perform better than the statistical models, but they cannot incorporate the graph structure when learning. FC-GAGA has lower forecasting errors as it is equipped with a graph learning module. The spatio-temporal graph-based models (especially AGCRN, GMAN, GWN, and LSGCN) display better performance. These models either use the observed graph or learn the graph structure from the data. In general, the deep learning based probabilistic forecasting algorithms such as DeepAR, DeepFactors, and MQRNN do not account for the spatial relationships in the data as well as the graph-based models, although MQRNN is among the best performing algorithms. DeepAR and DeepFactors aim to model the forecasting distributions and thus do not perform as well in the point forecasting task. The training loss function (negative log likelihood of the forecasts) does not match the evaluation metric. However, MQRNN shows better performance, possibly because it does target learning the median of the forecasting distribution along with other quantiles. The proposed AGCGRU+flow algorithm demonstrates comparable prediction accuracy to the best-performing spatio-temporal models and achieves the best average ranking across the four datasets.
Figure~\ref{fig:mae_comp} demonstrates that the proposed AGCGRU+flow has lower average MAE in most of the nodes compared to the second best performing AGCRN algorithm, for all four PeMS datasets. Some qualitative visualization of the
confidence intervals for 15-minute ahead predictions for the PeMSD3, PeMSD4, PeMSD7, and PeMSD8 datasets are shown in Figures~\ref{fig:pems03},~\ref{fig:pems04},~\ref{fig:pems07}, and~\ref{fig:pems08} respectively. We observe that the confidence intervals from the proposed algorithm are considerably tighter compared to its competitors in most cases, whereas the coverage of the ground truth is still ensured.

\begin{table*}[htbp]
\caption{Average MAE, MAPE, and RMSE for PeMSD3, PeMSD4, PeMSD7, and PeMSD8 for 15/30/45/60 minutes horizons for AGCGRU+flow
and AGCGRU+BPF. Lower numbers are better.}
\centering
\begin{tabular}{c|ccc}
\hline
\multirow{2}{*}{Algorithm} & \multicolumn{3}{c}{PeMSD3 (15/ 30/ 45/ 60 min)} \\ \cline{2-4} 
                  &MAE       &MAPE(\%)       &RMSE       \\ \hline \hline
AGCGRU+flow &\textbf{13.79}/\textbf{14.84}/\textbf{15.58}/\textbf{16.06} &\textbf{14.01}/\textbf{14.75}/\textbf{15.34}/\textbf{15.80} &\textbf{22.08}/\textbf{24.26}/\textbf{25.55}/\textbf{26.43}   \\
AGCGRU+BPF &14.19/15.13/15.85/16.35
&14.21/14.86/15.40/15.82
&25.69/27.38/28.51/29.26 \\
\hline \hline

\multirow{2}{*}{Algorithm} & \multicolumn{3}{c}{PeMSD4 (15/ 30/ 45/ 60 min)} \\ \cline{2-4} 
                  &MAE       &MAPE(\%)       &RMSE       \\ \hline \hline
AGCGRU+flow 
&\textbf{1.35}/\textbf{1.63}/\textbf{1.78}/\textbf{1.88}
&\textbf{2.67}/\textbf{3.44}/\textbf{3.87}/\textbf{4.16} 
&\textbf{2.88}/\textbf{3.77}/\textbf{4.20}/\textbf{4.46}   \\
AGCGRU+BPF 
&1.36/1.65/1.80/1.90
&2.71/3.46/3.90/4.18
&2.91/3.81/4.25/4.52 \\
\hline  \hline
\multirow{2}{*}{Algorithm} & \multicolumn{3}{c}{PeMSD7 (15/ 30/ 45/ 60 min)} \\ \cline{2-4} 
                  &MAE       &MAPE(\%)       &RMSE       \\ \hline \hline
AGCGRU+flow &\textbf{2.15}/\textbf{2.70}/2.99/3.19 &\textbf{5.13}/\textbf{6.75}/\textbf{7.61}/\textbf{8.18} 
&\textbf{4.11}/\textbf{5.46}/\textbf{6.12}/\textbf{6.54}     \\  
AGCGRU+BPF 
&2.19/2.73/2.99/\textbf{3.17}
&5.27/6.86/7.69/8.21
&4.18/5.52/6.16/6.53 \\
\hline \hline
\multirow{2}{*}{Algorithm} & \multicolumn{3}{c}{PeMSD8 (15/ 30/ 45/ 60 min)} \\ \cline{2-4} 
                  &MAE       &MAPE(\%)       &RMSE       \\ \hline \hline
AGCGRU+flow &\textbf{1.13}/\textbf{1.37}/\textbf{1.49}/\textbf{1.57} &\textbf{2.30}/\textbf{3.01}/\textbf{3.40}/\textbf{3.65} 
&\textbf{2.59}/\textbf{3.45}/\textbf{3.85}/\textbf{4.09}  \\  
AGCGRU+BPF 
&1.18/1.41/1.52/1.59
&2.47/3.13/3.50/3.74
&2.69/3.53/3.92/4.15 \\ \hline
\end{tabular}
\label{tab:bpf_point}
\end{table*}

\begin{table*}[htbp]
\caption{Average CRPS, P10QL, and P90QL for PeMSD3, PeMSD4, PeMSD7, and PeMSD8 for 15/30/45/60 minutes horizons for AGCGRU+flow
and AGCGRU+BPF. Lower numbers are better.}
\centering
\begin{tabular}{c|ccc}
\hline
\multirow{2}{*}{Algorithm} & \multicolumn{3}{c}{PeMSD3 (15/ 30/ 45/ 60 min)} \\ \cline{2-4} 
                  &CRPS       &P10QL(\%)       &P90QL(\%)       \\ \hline \hline
AGCGRU+flow  &\textbf{10.53}/\textbf{11.39}/\textbf{12.03}/\textbf{12.47}
&\textbf{4.01}/\textbf{4.44}/\textbf{4.76}/\textbf{4.97}
&\textbf{4.06}/\textbf{4.38}/\textbf{4.63}/\textbf{4.82}   \\
AGCGRU+BPF &11.32/11.94/12.55/12.92
&4.36/4.66/4.98/5.13
&4.39/4.65/4.88/5.07 \\
\hline
\hline
\multirow{2}{*}{Algorithm} & \multicolumn{3}{c}{PeMSD4 (15/ 30/ 45/ 60 min)} \\ \cline{2-4} 
                  &CRPS       &P10QL(\%)       &P90QL(\%)       \\ \hline \hline
AGCGRU+flow &\textbf{1.08}/1.32/1.46/1.56
&\textbf{1.28}/1.62/1.82/1.97
&\textbf{1.05}/1.26/1.37/1.45   \\
AGCGRU+BPF &1.10/1.32/\textbf{1.45}/\textbf{1.54}
&1.29/\textbf{1.60}/\textbf{1.79}/\textbf{1.92}
&1.06/1.26/1.37/1.45 \\
\hline  \hline
\multirow{2}{*}{Algorithm} & \multicolumn{3}{c}{PeMSD7 (15/ 30/ 45/ 60 min)} \\ \cline{2-4} 
                  &CRPS       &P10QL(\%)       &P90QL(\%)       \\ \hline \hline
AGCGRU+flow &\textbf{1.73}/\textbf{2.18}/\textbf{2.43}/\textbf{2.58}
&\textbf{2.27}/\textbf{2.97}/\textbf{3.36}/\textbf{3.60}
&\textbf{1.83}/\textbf{2.25}/\textbf{2.48}/\textbf{2.62}  \\
AGCGRU+BPF 
&1.79/2.24/2.49/2.66
&2.35/3.02/3.40/3.67
&1.86/2.29/2.53/2.69 \\
 \hline \hline
\multirow{2}{*}{Algorithm} & \multicolumn{3}{c}{PeMSD8 (15/ 30/ 45/ 60 min)} \\ \cline{2-4} 
                  &CRPS       &P10QL(\%)       &P90QL(\%)        \\ \hline \hline
AGCGRU+flow &\textbf{0.90}/\textbf{1.10}/\textbf{1.20}/1.28
&\textbf{1.10}/\textbf{1.43}/\textbf{1.61}/\textbf{1.73}
&\textbf{0.87}/\textbf{1.01}/1.09/1.14   \\
AGCGRU+BPF 
&0.96/1.13/1.22/1.28
&1.19/1.47/1.63/1.74
&0.91/1.03/1.09/\textbf{1.13} \\
\hline
\end{tabular}
\label{tab:bpf_probab}
\end{table*}

\subsection{Detailed results for comparison with particle filter}
In Table~\ref{tab:flow_bpf} of the main paper, we compare the average MAE and average CRPS of the proposed AGCGRU+flow with a Bootstrap Particle Filter (BPF)~\cite{gordon1993} based approach. Tables~\ref{tab:bpf_point} and~\ref{tab:bpf_probab} provide the detailed comparison both in terms of point forecasting and probabilistic forecasting metrics. We observe that the proposed AGCGRU+flow algorithm outperforms the particle filter based approach in most cases.

\subsection{Effect of number of particles}
For this experiment, we consider three different settings with varying number of particles $N_p = 1/10/50$ for testing. The model is trained using 1 particle in each case. From Table~\ref{tab:num_particle_point}, we observe that increasing the number of particles cannot improve the point forecasting accuracy significantly, whereas the results in Table~\ref{tab:num_particle_probab} show that characterization of the prediction uncertainty is improved as more particles are used to form the approximate posterior distribution of the forecasts. 
\begin{table*}[htbp] 
\caption{Average MAE, MAPE, and RMSE for PeMSD3, PeMSD4, PeMSD7, and PeMSD8 for 15/30/45/60 minutes horizons for AGCGRU+flow with different number of particles. Lower numbers are better.}
\centering
\begin{tabular}{c|ccc}
\hline
\multirow{2}{*}{Algorithm} & \multicolumn{3}{c}{PeMSD3 (15/ 30/ 45/ 60 min)} \\ \cline{2-4} 
                  &MAE       &MAPE(\%)       &RMSE       \\ \hline \hline
AGCGRU+flow ($N_p=1$) 
&13.82/14.87/15.60/16.08
&14.04/14.78/15.36/15.82
&22.33/24.41/25.70/26.54 \\
AGCGRU+flow ($N_p=10$) &\textbf{13.79}/\textbf{14.84}/\textbf{15.58}/\textbf{16.06}
&\textbf{14.01}/14.75/15.34/15.80 
&22.08/24.26/\textbf{25.55}/26.43   \\
AGCGRU+flow ($N_p=50$) &\textbf{13.79}/\textbf{14.84}/\textbf{15.58}/\textbf{16.06}
&\textbf{14.01}/\textbf{14.74}/\textbf{15.33}/\textbf{15.79}
&\textbf{22.02}/\textbf{24.20}/\textbf{25.55}/\textbf{26.42} \\
\hline \hline

\multirow{2}{*}{Algorithm} & \multicolumn{3}{c}{PeMSD4 (15/ 30/ 45/ 60 min)} \\ \cline{2-4} 
                  &MAE       &MAPE(\%)       &RMSE       \\ \hline \hline
AGCGRU+flow ($N_p=1$) 
&1.35/1.63/1.78/1.88
&2.68/3.45/3.89/4.18
&2.89/3.78/4.22/4.47  \\ 
AGCGRU+flow ($N_p=10$) 
&1.35/1.63/1.78/1.88
&\textbf{2.67}/\textbf{3.44}/\textbf{3.87}/\textbf{4.16} 
&\textbf{2.88}/\textbf{3.77}/\textbf{4.20}/4.46   \\
AGCGRU+flow ($N_p=50$) 
&1.35/1.63/1.78/1.88
&\textbf{2.67}/\textbf{3.44}/\textbf{3.87}/\textbf{4.16}
&\textbf{2.88}/\textbf{3.77}/\textbf{4.20}/\textbf{4.45} \\
 \hline  \hline
\multirow{2}{*}{Algorithm} & \multicolumn{3}{c}{PeMSD7 (15/ 30/ 45/ 60 min)} \\ \cline{2-4} 
                  &MAE       &MAPE(\%)       &RMSE       \\ \hline \hline
AGCGRU+flow ($N_p=1$) 
&2.16/2.71/3.00/3.20
&5.14/6.77/7.63/8.20
&4.12/5.47/6.14/6.56 \\
AGCGRU+flow ($N_p=10$) 
&\textbf{2.15}/\textbf{2.70}/\textbf{2.99}/\textbf{3.19} 
&5.13/\textbf{6.75}/\textbf{7.61}/\textbf{8.18}
&\textbf{4.11}/\textbf{5.46}/\textbf{6.12}/\textbf{6.54}     \\ 
AGCGRU+flow ($N_p=50$) 
&\textbf{2.15}/\textbf{2.70}/\textbf{2.99}/\textbf{3.19} 
&\textbf{5.12}/\textbf{6.75}/\textbf{7.61}/\textbf{8.18}
&\textbf{4.11}/\textbf{5.46}/\textbf{6.12}/\textbf{6.54} \\
\hline \hline
\multirow{2}{*}{Algorithm} & \multicolumn{3}{c}{PeMSD8 (15/ 30/ 45/ 60 min)} \\ \cline{2-4} 
                  &MAE       &MAPE(\%)       &RMSE       \\ \hline \hline
AGCGRU+flow ($N_p=1$) 
&1.14/1.38/1.50/1.57 
&2.31/3.02/3.41/3.67
&2.60/3.46/3.87/4.11 \\ 
AGCGRU+flow ($N_p=10$) &\textbf{1.13}/\textbf{1.37}/\textbf{1.49}/1.57 &\textbf{2.30}/\textbf{3.01}/\textbf{3.40}/\textbf{3.65} 
&\textbf{2.59}/3.45/\textbf{3.85}/\textbf{4.09}  \\
AGCGRU+flow ($N_p=50$) 
&\textbf{1.13}/\textbf{1.37}/\textbf{1.49}/1.57
&\textbf{2.30}/\textbf{3.01}/\textbf{3.40}/\textbf{3.65} 
&\textbf{2.59}/\textbf{3.44}/\textbf{3.85}/\textbf{4.09}
\\\hline
\end{tabular}
\label{tab:num_particle_point}
\end{table*}

\begin{table*}[htbp]
\caption{Average CRPS, P10QL, and P90QL for PeMSD3, PeMSD4, PeMSD7, and PeMSD8 for 15/30/45/60 minutes horizons for AGCGRU+flow with different number of particles. Lower numbers are better.}
\centering
\begin{tabular}{c|ccc}
\hline
\multirow{2}{*}{Algorithm} & \multicolumn{3}{c}{PeMSD3 (15/ 30/ 45/ 60 min)} \\ \cline{2-4} 
                  &CRPS       &P10QL(\%)       &P90QL(\%)       \\ \hline \hline
AGCGRU+flow ($N_p=1$) 
&19.34/20.44/21.24/21.80
&11.79/12.80/13.46/13.91
&10.46/10.72/10.98/11.18 \\ 
AGCGRU+flow ($N_p=10$)  &10.53/11.39/12.03/12.47
&4.01/4.44/4.76/4.97
&4.06/4.38/4.63/4.82   \\  
AGCGRU+flow ($N_p=50$) 
&\textbf{10.02}/\textbf{10.86}/\textbf{11.49}/\textbf{11.92}
&\textbf{3.67}/\textbf{4.05}/\textbf{4.33}/\textbf{4.53}
&\textbf{3.83}/\textbf{4.15}/\textbf{4.41}/\textbf{4.59} \\
\hline
\hline
\multirow{2}{*}{Algorithm} & \multicolumn{3}{c}{PeMSD4 (15/ 30/ 45/ 60 min)} \\ \cline{2-4} 
                  &CRPS       &P10QL(\%)       &P90QL(\%)       \\ \hline \hline
AGCGRU+flow ($N_p=1$) 
&1.95/2.34/2.58/2.73
&3.11/3.75/4.16/4.47
&3.00/3.59/3.92/4.10 \\ 
AGCGRU+flow ($N_p=10$) &1.08/1.32/1.46/1.56
&1.28/1.62/1.82/1.97
&1.05/1.26/1.37/1.45   \\
AGCGRU+flow ($N_p=50$) 
&\textbf{1.03}/\textbf{1.26}/\textbf{1.40}/\textbf{1.49}
&\textbf{1.21}/\textbf{1.54}/\textbf{1.73}/\textbf{1.87}
&\textbf{0.98}/\textbf{1.17}/\textbf{1.27}/\textbf{1.35} \\
 \hline  \hline
\multirow{2}{*}{Algorithm} & \multicolumn{3}{c}{PeMSD7 (15/ 30/ 45/ 60 min)} \\ \cline{2-4} 
                  &CRPS       &P10QL(\%)       &P90QL(\%)       \\ \hline \hline
AGCGRU+flow ($N_p=1$) 
&3.18/3.95/4.35/4.61
&5.57/6.96/7.67/8.15
&5.38/6.63/7.29/7.69 \\ 
AGCGRU+flow ($N_p=10$) &1.73/2.18/2.43/2.58
&2.27/2.97/3.36/3.60
&1.83/2.25/2.48/2.62  \\
AGCGRU+flow ($N_p=50$) 
&\textbf{1.64}/\textbf{2.09}/\textbf{2.32}/\textbf{2.47}
&\textbf{2.16}/\textbf{2.83}/\textbf{3.20}/\textbf{3.44}
&\textbf{1.71}/\textbf{2.10}/\textbf{2.31}/\textbf{2.45} \\
 \hline \hline
\multirow{2}{*}{Algorithm} & \multicolumn{3}{c}{PeMSD8 (15/ 30/ 45/ 60 min)} \\ \cline{2-4} 
                  &CRPS       &P10QL(\%)       &P90QL(\%)        \\ \hline \hline
AGCGRU+flow ($N_p=1$) 
&1.63/1.90/2.07/2.18
&2.73/3.28/3.63/3.87
&2.38/2.68/2.86/2.98 \\ 
AGCGRU+flow ($N_p=10$) &0.90/1.10/1.20/1.28
&1.10/1.43/1.61/1.73
&0.87/1.01/1.09/1.14   \\
AGCGRU+flow ($N_p=50$) 
&\textbf{0.86}/\textbf{1.05}/\textbf{1.16}/\textbf{1.22}
&\textbf{1.04}/\textbf{1.35}/\textbf{1.52}/\textbf{1.63}
&\textbf{0.83}/\textbf{0.95}/\textbf{1.03}/\textbf{1.08} \\
 \hline
\end{tabular}
\label{tab:num_particle_probab}
\end{table*}

\subsection{Effect of different learnable noise variance at each node}
In this experiment, we compare the proposed state-space model with different learnable noise variance at each node (parameterized by the softplus function in eq.~\eqref{eq:softplus} in the main paper with fixed and uniform noise standard deviation $\gamma = 0.01/0.05/0.10$ at all nodes. Other hyper-parameters and the training setup remain unchanged. The results in Table~\ref{tab:diff_var_point} demonstrate that the learnable noise variance approach is not particularly beneficial in comparison to a uniform, fixed variance approach in most cases. However, we note that the probabilistic metrics reported in Table~\ref{tab:diff_var_probab} are the lowest for the learnable noise variance model in all cases. This suggests that different time-series in these road traffic datasets have different degrees of uncertainty which cannot be effectively modelled by the uniform, fixed noise variance approach.   

\begin{table*}[htbp]
\caption{Average MAE, MAPE, and RMSE for PeMSD3, PeMSD4, PeMSD7, and PeMSD8 for 15/30/45/60 minutes horizons for AGCGRU+flow with learnable and fixed noise variance settings. The best result in each column is shown in bold. Lower numbers are better.}
\centering
\begin{tabular}{c|ccc}
\hline
\multirow{2}{*}{Algorithm} & \multicolumn{3}{c}{PeMSD3 (15/ 30/ 45/ 60 min)} \\ \cline{2-4} 
                  &MAE       &MAPE(\%)       &RMSE       \\ \hline \hline
AGCGRU+flow (learnable) &13.79/14.84/15.58/16.06 &\textbf{14.01}/\textbf{14.75}/\textbf{15.34}/\textbf{15.80} 
&22.08/24.26/25.55/26.43   \\
AGCGRU+flow ($\gamma=0.01$) 
&\textbf{13.68}/\textbf{14.75}/\textbf{15.49}/\textbf{16.02}
&14.57/15.37/16.02/16.57
&\textbf{21.74}/\textbf{23.95}/\textbf{25.27}/\textbf{26.21} \\
AGCGRU+flow ($\gamma=0.05$) 
&13.96/15.05/15.76/16.25
&15.87/16.66/17.23/17.62
&22.08/24.33/25.64/26.54 \\
AGCGRU+flow ($\gamma=0.10$) 
&13.86/14.91/15.68/16.17
&14.42/15.20/15.87/16.39
&22.04/24.25/25.60/26.41 \\
\hline \hline

\multirow{2}{*}{Algorithm} & \multicolumn{3}{c}{PeMSD4 (15/ 30/ 45/ 60 min)} \\ \cline{2-4} 
                  &MAE       &MAPE(\%)       &RMSE       \\ \hline \hline
AGCGRU+flow (learnable) 
&\textbf{1.35}/\textbf{1.63}/\textbf{1.78}/\textbf{1.88}
&\textbf{2.67}/\textbf{3.44}/\textbf{3.87}/\textbf{4.16} 
&\textbf{2.88}/3.77/\textbf{4.20}/\textbf{4.46}   \\
AGCGRU+flow ($\gamma=0.01$) 
&\textbf{1.35}/\textbf{1.63}/1.79/1.89
&2.68/3.45/3.89/4.20
&\textbf{2.88}/3.77/\textbf{4.20}/4.47 \\
AGCGRU+flow ($\gamma=0.05$) 
&1.36/1.65/1.80/1.91
&2.69/3.47/3.91/4.21
&\textbf{2.88}/\textbf{3.76}/\textbf{4.20}/\textbf{4.46} \\
AGCGRU+flow ($\gamma=0.10$) 
&1.36/1.65/1.80/1.90
&2.70/3.47/3.89/4.18
&2.92/3.81/4.24/4.49 \\
 \hline  \hline
\multirow{2}{*}{Algorithm} & \multicolumn{3}{c}{PeMSD7 (15/ 30/ 45/ 60 min)} \\ \cline{2-4} 
                  &MAE       &MAPE(\%)       &RMSE       \\ \hline \hline
AGCGRU+flow (learnable) 
&2.15/2.70/2.99/3.19 
&5.13/6.75/7.61/8.18 
&4.11/5.46/6.12/6.54     \\  
AGCGRU+flow ($\gamma=0.01$) 
&\textbf{2.14}/\textbf{2.69}/\textbf{2.98}/\textbf{3.16}
&\textbf{5.07}/\textbf{6.66}/\textbf{7.47}/\textbf{8.00}
&4.10/5.43/6.09/6.49 \\
AGCGRU+flow ($\gamma=0.05$) 
&2.16/2.71/3.00/3.20
&5.13/6.74/7.61/8.19
&\textbf{4.09}/\textbf{5.41}/\textbf{6.06}/\textbf{6.48} \\
AGCGRU+flow ($\gamma=0.10$) 
&2.16/2.73/3.01/3.20
&5.15/6.77/7.62/8.15
&4.12/5.48/6.15/6.54 \\
\hline \hline
\multirow{2}{*}{Algorithm} & \multicolumn{3}{c}{PeMSD8 (15/ 30/ 45/ 60 min)} \\ \cline{2-4} 
                  &MAE       &MAPE(\%)       &RMSE       \\ \hline \hline
AGCGRU+flow (learnable) 
&1.13/\textbf{1.37}/\textbf{1.49}/\textbf{1.57} &2.30/3.01/3.40/3.65 
&2.59/3.45/3.85/4.09  \\  
AGCGRU+flow ($\gamma=0.01$) 
&1.13/\textbf{1.37}/\textbf{1.49}/\textbf{1.57}
&2.31/3.03/3.44/3.71
&2.60/3.43/3.84/4.09 \\
AGCGRU+flow ($\gamma=0.05$) 
&1.13/\textbf{1.37}/\textbf{1.49}/\textbf{1.57}
&\textbf{2.26}/\textbf{2.95}/\textbf{3.35}/\textbf{3.62}
&\textbf{2.53}/\textbf{3.34}/\textbf{3.75}/\textbf{4.01} \\
AGCGRU+flow ($\gamma=0.10$) 
&1.13/1.38/1.51/1.60
&2.31/3.04/3.49/3.80
&2.57/3.41/3.86/4.14 \\ \hline
\end{tabular}
\label{tab:diff_var_point}
\end{table*}

\begin{table*}[htbp]
\caption{Average CRPS, P10QL, and P90QL for PeMSD3, PeMSD4, PeMSD7, and PeMSD8 for 15/30/45/60 minutes horizons for AGCGRU+flow with learnable and fixed noise variance settings. The best result in each column is shown in bold. Lower numbers are better.}
\centering
\begin{tabular}{c|ccc}
\hline
\multirow{2}{*}{Algorithm} & \multicolumn{3}{c}{PeMSD3 (15/ 30/ 45/ 60 min)} \\ \cline{2-4} 
                  &CRPS       &P10QL(\%)       &P90QL(\%)       \\ \hline \hline
AGCGRU+flow (learnable)   
&\textbf{10.53}/\textbf{11.39}/\textbf{12.03}/\textbf{12.47}
&\textbf{4.01}/\textbf{4.44}/\textbf{4.76}/\textbf{4.97}
&\textbf{4.06}/\textbf{4.38}/\textbf{4.63}/\textbf{4.82}   \\
AGCGRU+flow ($\gamma=0.01$) 
&12.83/13.90/14.63/15.17
&7.26/8.10/8.46/8.77
&6.68/7.08/7.55/7.86 \\
AGCGRU+flow ($\gamma=0.05$) 
&11.58/12.61/13.28/13.74
&5.78/6.52/6.99/7.25
&5.14/5.54/5.81/6.06 \\
AGCGRU+flow ($\gamma=0.10$) 
&13.14/14.18/14.95/15.43
&7.79/8.57/9.22/9.53
&6.64/7.05/7.28/7.53 \\
\hline
\hline
\multirow{2}{*}{Algorithm} & \multicolumn{3}{c}{PeMSD4 (15/ 30/ 45/ 60 min)} \\ \cline{2-4} 
                  &CRPS       &P10QL(\%)       &P90QL(\%)       \\ \hline \hline
AGCGRU+flow (learnable) 
&\textbf{1.08}/\textbf{1.32}/\textbf{1.46}/\textbf{1.56}
&\textbf{1.28}/\textbf{1.62}/\textbf{1.82}/\textbf{1.97}
&\textbf{1.05}/\textbf{1.26}/\textbf{1.37}/\textbf{1.45}   \\
AGCGRU+flow ($\gamma=0.01$) 
&1.28/1.55/1.70/1.81
&2.09/2.58/2.87/3.08
&1.74/2.08/2.26/2.38 \\
AGCGRU+flow ($\gamma=0.05$) 
&1.19/1.47/1.62/1.72
&1.82/2.30/2.57/2.77
&1.48/1.84/2.04/2.15 \\
AGCGRU+flow ($\gamma=0.10$) 
&1.32/1.60/1.75/1.85
&2.19/2.68/2.95/3.15
&1.84/2.23/2.43/2.54 \\
 \hline  \hline
\multirow{2}{*}{Algorithm} & \multicolumn{3}{c}{PeMSD7 (15/ 30/ 45/ 60 min)} \\ \cline{2-4} 
                  &CRPS       &P10QL(\%)       &P90QL(\%)       \\ \hline \hline
AGCGRU+flow (learnable) 
&\textbf{1.73}/\textbf{2.18}/\textbf{2.43}/\textbf{2.58}
&\textbf{2.27}/\textbf{2.97}/\textbf{3.36}/\textbf{3.60}
&\textbf{1.83}/\textbf{2.25}/\textbf{2.48}/\textbf{2.62}  \\
AGCGRU+flow ($\gamma=0.01$) 
&2.02/2.55/2.82/3.01
&3.59/4.57/5.05/5.35
&3.00/3.77/4.22/4.54 \\
AGCGRU+flow ($\gamma=0.05$) 
&1.90/2.42/2.70/2.90
&3.18/4.20/4.76/5.15
&2.56/3.27/3.65/3.91 \\
AGCGRU+flow ($\gamma=0.10$) 
&2.09/2.65/2.94/3.12
&3.80/4.87/5.41/5.77
&3.18/4.04/4.47/4.73 \\
 \hline \hline
\multirow{2}{*}{Algorithm} & \multicolumn{3}{c}{PeMSD8 (15/ 30/ 45/ 60 min)} \\ \cline{2-4} 
                  &CRPS       &P10QL(\%)       &P90QL(\%)        \\ \hline \hline
AGCGRU+flow (learnable)
&\textbf{0.90}/\textbf{1.10}/\textbf{1.20}/\textbf{1.28}
&\textbf{1.10}/\textbf{1.43}/\textbf{1.61}/\textbf{1.73}
&\textbf{0.87}/\textbf{1.01}/\textbf{1.09}/\textbf{1.14}   \\
AGCGRU+flow ($\gamma=0.01$) 
&1.07/1.29/1.41/1.49
&1.81/2.29/2.56/2.75
&1.35/1.57/1.67/1.73 \\
AGCGRU+flow ($\gamma=0.05$) 
&1.00/1.23/1.35/1.43
&1.58/2.04/2.31/2.50
&1.21/1.43/1.52/1.58 \\
AGCGRU+flow ($\gamma=0.10$) 
&1.10/1.34/1.47/1.56
&1.88/2.41/2.72/2.93
&1.47/1.71/1.81/1.87 \\
 \hline
\end{tabular}
\label{tab:diff_var_probab}
\end{table*}

\begin{table*}[htbp]
\caption{Average MAE, MAPE, and RMSE for PeMSD3, PeMSD4, PeMSD7, and PeMSD8 for 15/30/45/60 minutes horizons for the proposed flow based
approach and deterministic encoder-decoder models.  Lower numbers are better.}
\centering
\begin{tabular}{c|ccc}
\hline
\multirow{2}{*}{Algorithm} & \multicolumn{3}{c}{PeMSD3 (15/ 30/ 45/ 60 min)} \\ \cline{2-4} 
                  &MAE       &MAPE(\%)       &RMSE       \\ \hline \hline 
AGCGRU+flow &\textbf{13.79}/\textbf{14.84}/\textbf{15.58}/\textbf{16.06} &\textbf{14.01}/\textbf{14.75}/\textbf{15.34}/\textbf{15.80} &\textbf{22.08}/\textbf{24.26}/\textbf{25.55}/\textbf{26.43}   \\
FC-AGCGRU &13.96/15.37/16.52/17.45
&14.26/15.61/16.69/17.37
&25.28/27.43/29.09/30.43 \\ \hline
DCGRU+flow   &14.48/\textbf{15.67}/\textbf{16.52}/\textbf{17.36}
&15.06/16.06/16.91/\textbf{17.84}
&\textbf{23.86}/\textbf{26.12}/\textbf{27.54}/\textbf{28.76}     \\
FC-DCGRU  &\textbf{14.42}/15.87/17.10/18.29 &\textbf{14.57}/\textbf{15.78}/\textbf{16.87}/17.95
&24.33/27.05/28.99/30.76 \\ \hline
GRU+flow  &\textbf{14.40}/\textbf{16.10}/\textbf{17.63}/\textbf{19.18}
&\textbf{14.56}/\textbf{15.99}/\textbf{17.33}/\textbf{18.89}
&\textbf{23.06}/\textbf{26.15}/\textbf{28.64}/\textbf{30.97}    \\
FC-GRU 
&15.82/18.37/20.61/22.93
&15.87/18.82/21.32/23.75
&25.85/30.09/33.37/36.94 \\ \hline
\hline
\multirow{2}{*}{Algorithm} & \multicolumn{3}{c}{PeMSD4 (15/ 30/ 45/ 60 min)} \\ \cline{2-4} 
                  &MAE       &MAPE(\%)       &RMSE       \\ \hline \hline
AGCGRU+flow &\textbf{1.35}/\textbf{1.63}/\textbf{1.78}/\textbf{1.88} &\textbf{2.67}/\textbf{3.44}/\textbf{3.87}/\textbf{4.16}
&\textbf{2.88}/\textbf{3.77}/\textbf{4.20}/\textbf{4.46}   \\
FC-AGCGRU &1.37/1.74/2.00/2.20
&2.69/3.67/4.41/5.00
&2.92/3.96/4.62/5.09
\\ \hline
DCGRU+flow    &\textbf{1.38}/\textbf{1.71}/\textbf{1.92}/\textbf{2.08} &2.72/\textbf{3.63}/\textbf{4.23}/\textbf{4.67}
&\textbf{2.93}/\textbf{3.93}/\textbf{4.49}/\textbf{4.87}   \\
FC-DCGRU  &\textbf{1.38}/1.78/2.06/2.29       &\textbf{2.69}/3.72/4.51/5.16
&2.95/4.09/4.81/5.34 \\ \hline
GRU+flow  &\textbf{1.37}/\textbf{1.76}/\textbf{2.02}/\textbf{2.23}
&\textbf{2.70}/\textbf{3.74}/\textbf{4.52}/\textbf{5.15}
&\textbf{2.95}/\textbf{4.05}/\textbf{4.74}/\textbf{5.23} \\
FC-GRU 
&1.46/1.91/2.25/2.54
&2.84/3.97/4.88/5.66
&3.10/4.35/5.20/5.85
 \\ \hline  \hline
\multirow{2}{*}{Algorithm} & \multicolumn{3}{c}{PeMSD7 (15/ 30/ 45/ 60 min)} \\ \cline{2-4} 
                  &MAE       &MAPE(\%)       &RMSE       \\ \hline \hline
AGCGRU+flow &\textbf{2.15}/\textbf{2.70}/\textbf{2.99}/\textbf{3.19} &\textbf{5.13}/\textbf{6.75}/\textbf{7.61}/\textbf{8.18}
&\textbf{4.11}/\textbf{5.46}/\textbf{6.12}/\textbf{6.54}     \\ 
FC-AGCGRU &2.21/2.99/3.56/4.05
&5.18/7.39/9.12/10.64
&4.18/5.88/7.03/7.94
 \\ \hline
DCGRU+flow  &\textbf{2.19}/\textbf{2.87}/\textbf{3.29}/\textbf{3.61}
&\textbf{5.16}/\textbf{7.17}/\textbf{8.48}/\textbf{9.42}
&\textbf{4.16}/\textbf{5.66}/\textbf{6.54}/\textbf{7.14}     \\ 
FC-DCGRU &2.23/3.06/3.67/4.18 
&5.19/7.50/9.31/10.90 
&4.26/6.05/7.28/8.24 \\ \hline
GRU+flow  &\textbf{2.24}/\textbf{3.02}/\textbf{3.55}/\textbf{3.96}
&\textbf{5.27}/\textbf{7.58}/\textbf{9.30}/\textbf{10.60}
&\textbf{4.28}/\textbf{5.97}/\textbf{7.00}/\textbf{7.73}  \\ 
FC-GRU 
&2.41/3.40/4.17/4.84
&5.60/8.27/10.47/12.40
&4.56/6.68/8.17/9.34
 \\ \hline \hline
\multirow{2}{*}{Algorithm} & \multicolumn{3}{c}{PeMSD8 (15/ 30/ 45/ 60 min)} \\ \cline{2-4} 
                  &MAE       &MAPE(\%)       &RMSE       \\ \hline \hline
AGCGRU+flow &\textbf{1.13}/\textbf{1.37}/\textbf{1.49}/\textbf{1.57} &\textbf{2.30}/\textbf{3.01}/\textbf{3.40}/\textbf{3.65} 
&2.59/\textbf{3.45}/\textbf{3.85}/\textbf{4.09}  \\ 
FC-AGCGRU &1.16/1.48/1.70/1.87
&\textbf{2.30}/3.17/3.78/4.25
&\textbf{2.58}/3.53/4.12/4.54
\\ \hline
DCGRU+flow  &1.17/\textbf{1.44}/\textbf{1.58}/\textbf{1.70}
&2.35/\textbf{3.12}/\textbf{3.57}/\textbf{3.87}
&2.64/3.54/\textbf{4.00}/\textbf{4.28}     \\ 
FC-DCGRU &\textbf{1.16}/1.49/1.70/1.87 
&\textbf{2.25}/3.16/3.85/4.37 
&\textbf{2.54}/\textbf{3.49}/4.08/4.49 \\ \hline
GRU+flow  &\textbf{1.12}/\textbf{1.41}/\textbf{1.59}/\textbf{1.74}
&\textbf{2.17}/\textbf{2.94}/\textbf{3.50}/\textbf{3.92}
&\textbf{2.55}/\textbf{3.47}/\textbf{4.02}/\textbf{4.40}   \\ 
FC-GRU 
&1.20/1.56/1.81/2.02
&2.29/3.09/3.70/4.22
&2.63/3.61/4.24/4.73
 \\ \hline
\end{tabular}
\label{tab:enc_dec}
\end{table*}

\begin{table*}[htbp]
\caption{Average MAE, MAPE, and RMSE for PeMSD3, PeMSD4, PeMSD7, and PeMSD8 for 15/30/45/60 minutes horizons for AGCRN-ensemble,
GMAN-ensemble, and AGCGRU+flow. The best and the second best results in each column are shown in bold and marked with underline respectively. Lower numbers are better.}
\centering
\begin{tabular}{c|ccc}
\hline
\multirow{2}{*}{Algorithm} & \multicolumn{3}{c}{PeMSD3 (15/ 30/ 45/ 60 min)} \\ \cline{2-4} 
                  &MAE       &MAPE(\%)       &RMSE       \\ \hline \hline
AGCRN-ensemble 
&\underline{14.21}/\underline{15.12}/\underline{15.73}/\underline{16.22}
&\textbf{13.91}/\textbf{14.56}/\textbf{14.93}/\textbf{15.38}
&25.49/27.16/28.20/28.90
\\
GMAN-ensemble
&14.48/15.20/15.90/16.66
&15.01/15.64/16.41/17.36
&\underline{23.96}/\underline{25.20}/\underline{26.31}/\underline{27.44} \\
AGCGRU+flow  
&\textbf{13.79}/\textbf{14.84}/\textbf{15.58}/\textbf{16.06} 
&\underline{14.01}/\underline{14.75}/\underline{15.34}/\underline{15.80} 
&\textbf{22.08}/\textbf{24.26}/\textbf{25.55}/\textbf{26.43}   \\
\hline \hline
\multirow{2}{*}{Algorithm} & \multicolumn{3}{c}{PeMSD4 (15/ 30/ 45/ 60 min)} \\ \cline{2-4} 
                  &MAE       &MAPE(\%)       &RMSE       \\ \hline \hline
AGCRN-ensemble 
&\underline{1.35}/\underline{1.61}/\underline{1.76}/1.91
&2.75/\underline{3.40}/\underline{3.79}/4.17
&\underline{2.89}/\underline{3.65}/\underline{4.09}/4.47 \\
GMAN-ensemble 
&\textbf{1.33}/\textbf{1.57}/\textbf{1.72}/\textbf{1.84}
&\textbf{2.64}/\textbf{3.27}/\textbf{3.70}/\textbf{4.04}
&\underline{2.89}/\textbf{3.62}/\textbf{4.04}/\textbf{4.33} \\ 
AGCGRU+flow 
&\underline{1.35}/1.63/1.78/\underline{1.88} 
&\underline{2.67}/3.44/3.87/\underline{4.16} 
&\textbf{2.88}/3.77/4.20/\underline{4.46}   \\
\hline  \hline
\multirow{2}{*}{Algorithm} & \multicolumn{3}{c}{PeMSD7 (15/ 30/ 45/ 60 min)} \\ \cline{2-4} 
                  &MAE       &MAPE(\%)       &RMSE       \\ \hline \hline
AGCRN-ensemble 
&\underline{2.17}/\textbf{2.69}/\textbf{2.95}/\underline{3.20}
&\underline{5.25}/\textbf{6.75}/\textbf{7.55}/\underline{8.22}
&\textbf{4.09}/\textbf{5.29}/\textbf{5.94}/\textbf{6.45} \\ 
GMAN-ensemble
&2.42/2.80/3.08/3.35
&6.08/\underline{7.18}/8.00/8.74
&4.68/5.54/\underline{6.08}/\underline{6.51} \\
AGCGRU+flow 
&\textbf{2.15}/\underline{2.70}/\underline{2.99}/\textbf{3.19} 
&\textbf{5.13}/\textbf{6.75}/\underline{7.61}/\textbf{8.18} 
&\underline{4.11}/\underline{5.46}/6.12/6.54     \\  
\hline \hline
\multirow{2}{*}{Algorithm} & \multicolumn{3}{c}{PeMSD8 (15/ 30/ 45/ 60 min)} \\ \cline{2-4} 
                  &MAE       &MAPE(\%)       &RMSE       \\ \hline \hline
AGCRN-ensemble
&\underline{1.19}/\underline{1.36}/\underline{1.46}/1.58
&2.67/3.10/\underline{3.38}/3.68
&2.88/\underline{3.41}/\underline{3.76}/\underline{4.06} \\
GMAN-ensemble &\textbf{1.13}/\textbf{1.28}/\textbf{1.39}/\textbf{1.49}
&\underline{2.37}/\textbf{2.78}/\textbf{3.10}/\textbf{3.37}
&\underline{2.71}/\textbf{3.25}/\textbf{3.61}/\textbf{3.87} \\
AGCGRU+flow 
&\textbf{1.13}/1.37/1.49/\underline{1.57} 
&\textbf{2.30}/\underline{3.01}/3.40/\underline{3.65} 
&\textbf{2.59}/3.45/3.85/4.09  \\  
\hline
\end{tabular}
\label{tab:ensemble_point}
\end{table*}

\begin{table*}[htbp]
\caption{Average CRPS, P10QL, and P90QL for PeMSD3, PeMSD4, PeMSD7, and PeMSD8 for 15/30/45/60 minutes horizons for AGCRN-ensemble,
GMAN-ensemble, and AGCGRU+flow. The best and the second best results in each column are shown in bold and marked with underline respectively. Lower numbers are better.}
\centering
\begin{tabular}{c|ccc}
\hline
\multirow{2}{*}{Algorithm} & \multicolumn{3}{c}{PeMSD3 (15/ 30/ 45/ 60 min)} \\ \cline{2-4} 
                  &CRPS       &P10QL(\%)       &P90QL(\%)       \\ \hline \hline
AGCRN-ensemble 
&\underline{12.64}/\underline{13.44}/\underline{13.96}/\underline{14.27}
&\underline{6.90}/\underline{7.40}/\underline{7.54}/\underline{7.53}
&6.10/6.43/6.79/6.96\\  
GMAN-ensemble
&12.79/13.49/14.13/14.77
&7.17/7.67/8.08/8.45
&\underline{5.86}/\underline{6.16}/\underline{6.44}/\underline{6.68} \\
AGCGRU+flow
&\textbf{10.53}/\textbf{11.39}/\textbf{12.03}/\textbf{12.47}
&\textbf{4.01}/\textbf{4.44}/\textbf{4.76}/\textbf{4.97}
&\textbf{4.06}/\textbf{4.38}/\textbf{4.63}/\textbf{4.82}   \\
\hline
\hline
\multirow{2}{*}{Algorithm} & \multicolumn{3}{c}{PeMSD4 (15/ 30/ 45/ 60 min)} \\ \cline{2-4} 
                  &CRPS       &P10QL(\%)       &P90QL(\%)       \\ \hline \hline
AGCRN-ensemble 
&1.20/1.44/1.56/1.68
&1.82/2.21/2.39/2.57
&1.53/1.82/1.93/2.08 \\ 
GMAN-ensemble
&\underline{1.16}/\underline{1.38}/\underline{1.51}/\underline{1.62}
&\underline{1.73}/\underline{2.11}/\underline{2.35}/\underline{2.54}
&\underline{1.45}/\underline{1.70}/\underline{1.82}/\underline{1.92} \\
AGCGRU+flow 
&\textbf{1.08}/\textbf{1.32}/\textbf{1.46}/\textbf{1.56}
&\textbf{1.28}/\textbf{1.62}/\textbf{1.82}/\textbf{1.97}
&\textbf{1.05}/\textbf{1.26}/\textbf{1.37}/\textbf{1.45}   \\
 \hline  \hline
\multirow{2}{*}{Algorithm} & \multicolumn{3}{c}{PeMSD7 (15/ 30/ 45/ 60 min)} \\ \cline{2-4} 
                  &CRPS       &P10QL(\%)       &P90QL(\%)       \\ \hline \hline
AGCRN-ensemble
&\underline{1.90}/2.39/2.60/2.81
&3.22/4.15/4.55/4.89
&2.55/3.19/3.35/3.58 \\ 
GMAN-ensemble
&1.96/\underline{2.31}/\underline{2.53}/\underline{2.73}
&\underline{3.16}/\underline{3.83}/\underline{4.23}/\underline{4.53}
&\underline{2.20}/\underline{2.59}/\underline{2.81}/\underline{3.00} \\ 
AGCGRU+flow 
&\textbf{1.73}/\textbf{2.18}/\textbf{2.43}/\textbf{2.58}
&\textbf{2.27}/\textbf{2.97}/\textbf{3.36}/\textbf{3.60}
&\textbf{1.83}/\textbf{2.25}/\textbf{2.48}/\textbf{2.62}  \\
\hline \hline
\multirow{2}{*}{Algorithm} & \multicolumn{3}{c}{PeMSD8 (15/ 30/ 45/ 60 min)} \\ \cline{2-4} 
                  &CRPS       &P10QL(\%)       &P90QL(\%)        \\ \hline \hline
AGCRN-ensemble
&1.03/\underline{1.20}/1.28/\underline{1.38}
&1.63/1.97/2.14/2.34
&1.18/1.34/1.39/1.48 \\
GMAN-ensemble
&\underline{0.95}/\textbf{1.10}/\textbf{1.19}/\textbf{1.28}
&\underline{1.40}/\underline{1.68}/\underline{1.88}/\underline{2.04}
&\underline{1.12}/\underline{1.26}/\underline{1.34}/\underline{1.41} \\ 
AGCGRU+flow 
&\textbf{0.90}/\textbf{1.10}/\underline{1.20}/\textbf{1.28}
&\textbf{1.10}/\textbf{1.43}/\textbf{1.61}/\textbf{1.73}
&\textbf{0.87}/\textbf{1.01}/\textbf{1.09}/\textbf{1.14}   \\
\hline
\end{tabular}
\label{tab:ensemble_probab}
\end{table*}

\subsection{Detailed comparison with deterministic encoder-decoder models}
In Table~\ref{tab:flow_no_flow} of the main paper, we compare the average MAE of the proposed flow based approaches with those of the deterministic encoder-decoder based sequence to sequence prediction models for three different RNN architectures. In Table~\ref{tab:enc_dec}, we report the MAPE and RMSE, in addition to the MAE. We see that the particle flow based RNN models outperform the corresponding deterministic encoder-decoder models in most cases.

\subsection{Detailed results for comparison to ensembles}
In Table~\ref{tab:ensemble} of the main paper, we compare the average CRPS of the proposed AGCGRU+flow algorithm with ensembles of AGCRN and GAMN. From Table~\ref{tab:ensemble_point}, we observe that our approach is comparable or slightly worse compared to the ensembles in terms of the MAE, MAPE and RMSE of the point forecasts. However, the proposed AGCGRU+flow shows better characterization of the prediction uncertainty compared to the ensemble methods in almost all cases, as shown in Table~\ref{tab:ensemble_probab}. 

\subsection{Comparison with a Variational Inference (VI) based approach}
Although there is no directly applicable baseline forecasting method in the literature that incorporates VI, RNNs, and GNNs, we can derive a variational approach using equivalent GNN-RNN architectures and compare it to the particle flow approach. We wish to approximate $p_{\Theta}(\mathbf{y}_{P+1:P+Q}|\mathbf{y}_{1:P}, \mathbf{z}_{1:P+Q})$. So, the ELBO is defined as follows:
\begin{align}
\mathcal{L}(\Theta, \Omega) &= \mathbb{E}_{q_{\Omega}}\Bigg[\log p_{\Theta}(\mathbf{y}_{P+1:P+Q}, \mathbf{x}_{1:P}|\mathbf{y}_{1:P}, \mathbf{z}_{1:P+Q})  \,\nonumber\\
&\qquad - \log q_{\Omega}(\mathbf{x}_{1:P}|\mathbf{y}_{1:P+Q}, \mathbf{z}_{1:P+Q})\Bigg]\label{eq:elbo}\,.
\end{align}
Now, we approximate 
\begin{align}
&p_{\Theta}(\mathbf{y}_{P+1:P+Q}, \mathbf{x}_{1:P}|\mathbf{y}_{1:P}, \mathbf{z}_{1:P+Q}) \,\nonumber\\
&= \int \prod_{t=P+1}^{P+Q} \Big (p_{\phi, \gamma}(\mathbf{y}_{t} | \mathbf{x}_{t}, \mathbf{z}_{t}) \,\nonumber\\
&\qquad \qquad \qquad \quad p_{\psi, \sigma}(\mathbf{x}_{t} | \mathbf{x}_{t-1}, \mathbf{y}_{t-1}, \mathbf{z}_{t}) \Big)d\mathbf{x}_{P+1:P+Q}\,,\nonumber\\
&\approx \prod_{t=P+1}^{P+Q}\Bigg[\frac{1}{N_p}\sum_{j=1}^{N_p}p_{\phi, \gamma}(\mathbf{y}_t|\mathbf{x}_t^j, \mathbf{z}_{t}) \Bigg]\,,
\end{align}
where, in the decoder, we first sample $\mathbf{x}_{t}^j$ from  $p_{\psi, \sigma}(\mathbf{x}_t | \mathbf{x}_{t-1}^j, \mathbf{y}_{t-1}^j, \mathbf{z}_{t})$ (for $t>P+1$) or from $p_{\psi, \sigma}(\mathbf{x}_t | \mathbf{x}_{t-1}^j, \mathbf{y}_{t-1}, \mathbf{z}_{t})$ (for $t=P+1$) for $1 \leqslant j \leqslant N_p$ and then sample $\mathbf{y}_{t}^j$ from  $p_{\phi, \gamma}(\mathbf{y}_{t} | \mathbf{x}_{t}^j, \mathbf{z}_{t})$ for $1 \leqslant j \leqslant N_p$ to form the MC approximation. This decoder is initialized using the output of the encoder, i.e., we sample $\mathbf{x}_{1:P}^j$ from the inference distribution  $q_{\Omega}(\mathbf{x}_{1:P}|\mathbf{y}_{1:P+Q}, \mathbf{z}_{1:P+Q})$ for $1 \leqslant j \leqslant N_p$, which is assumed to be factorized as follows:
\begin{align}
&q_{\Omega}(\mathbf{x}_{1:P}|\mathbf{y}_{1:P+Q}, \mathbf{z}_{1:P+Q}) \,\nonumber\\
&\qquad= q_{\Omega}(\mathbf{x}_{1:P}|\mathbf{y}_{1:P}, \mathbf{z}_{1:P})\,,\nonumber\\
&\qquad=q_1(\mathbf{x}_{1}, \mathbf{z}_1, \rho) \prod_{t=2}^{P}  q_{\psi', \sigma'}(\mathbf{x}_t | \mathbf{x}_{t-1}, \mathbf{y}_{t-1}, \mathbf{z}_{t})\,.
\end{align}
Here, we set $q_1(\mathbf{x}_{1}, \mathbf{z}_1, \rho) = p_1(\mathbf{x}_{1}, \mathbf{z}_1, \rho)$ for simplicity and we use the same RNN architecture (i.e. AGCGRU) for $q_{\psi', \sigma'}$ and $p_{\psi, \sigma}$.

\textbf{Experimental details : } We treat $\rho$, $\sigma$ and $\sigma'$ as hyperparameters and set $\rho=1$ and $\sigma=\sigma' = 0$. This implies that $q_{\psi', \sigma'}$ is a Dirac-delta function and the maximization of ELBO (in eq.~\eqref{eq:elbo}) using SGD (SGVI) amounts to mimization of the same cost function as defined in eq.~\eqref{eq:prob_loss} in the main paper. The only difference is that now a) we have two separate AGCGRUs for encoder and decoder and b) there is no particle flow in the forward pass. We call this model AGCGRU+VI and compare it to AGCGRU+flow. The other hyperparameters are set to the same values as for the AGCGRU+flow algorithm. From Table~\ref{tab:vi_point}, we observe that for comparable RNN architectures, the flow based algorithm significantly outperforms the variational inference based approach in the point forecasting task. The results in  Table~\ref{tab:vi_probab} indicate that in the probabilistic forecasting task, both particle flow and VI approaches show 
comparable performance despite AGCGRU+flow having approximately half of the learnable parameters of the AGCGRU+VI model.

\begin{table*}[htbp]
\caption{Average MAE, MAPE, and RMSE for PeMSD3, PeMSD4, PeMSD7, and PeMSD8 for 15/30/45/60 minutes horizons for AGCGRU+flow
and AGCGRU+VI. Lower numbers are better.}
\centering
\begin{tabular}{c|ccc}
\hline
\multirow{2}{*}{Algorithm} & \multicolumn{3}{c}{PeMSD3 (15/ 30/ 45/ 60 min)} \\ \cline{2-4} 
                  &MAE       &MAPE(\%)       &RMSE       \\ \hline \hline
AGCGRU+flow &\textbf{13.79}/\textbf{14.84}/\textbf{15.58}/\textbf{16.06} &\textbf{14.01}/\textbf{14.75}/\textbf{15.34}/\textbf{15.80} &\textbf{22.08}/\textbf{24.26}/\textbf{25.55}/\textbf{26.43}   \\
AGCGRU+VI &15.08/16.10/16.83/17.53
&15.26/16.10/16.74/17.43
&26.17/28.02/29.13/30.17 \\
\hline \hline

\multirow{2}{*}{Algorithm} & \multicolumn{3}{c}{PeMSD4 (15/ 30/ 45/ 60 min)} \\ \cline{2-4} 
                  &MAE       &MAPE(\%)       &RMSE       \\ \hline \hline
AGCGRU+flow 
&\textbf{1.35}/\textbf{1.63}/\textbf{1.78}/\textbf{1.88}
&\textbf{2.67}/\textbf{3.44}/\textbf{3.87}/\textbf{4.16} 
&\textbf{2.88}/\textbf{3.77}/\textbf{4.20}/\textbf{4.46}   \\
AGCGRU+VI &1.46/1.76/1.94/2.06
&2.94/3.73/4.20/4.52
&2.97/3.78/4.22/4.48 \\
\hline  \hline
\multirow{2}{*}{Algorithm} & \multicolumn{3}{c}{PeMSD7 (15/ 30/ 45/ 60 min)} \\ \cline{2-4} 
                  &MAE       &MAPE(\%)       &RMSE       \\ \hline \hline
AGCGRU+flow &\textbf{2.15}/\textbf{2.70}/\textbf{2.99}/\textbf{3.19} &\textbf{5.13}/\textbf{6.75}/\textbf{7.61}/\textbf{8.18} 
&\textbf{4.11}/\textbf{5.46}/6.12/6.54    \\  
AGCGRU+VI &2.33/2.92/3.23/3.45
&5.59/7.26/8.16/8.78
&4.22/5.48/\textbf{6.10}/\textbf{6.50} \\
\hline \hline
\multirow{2}{*}{Algorithm} & \multicolumn{3}{c}{PeMSD8 (15/ 30/ 45/ 60 min)} \\ \cline{2-4} 
                  &MAE       &MAPE(\%)       &RMSE       \\ \hline \hline
AGCGRU+flow &\textbf{1.13}/\textbf{1.37}/\textbf{1.49}/\textbf{1.57} &\textbf{2.30}/\textbf{3.01}/\textbf{3.40}/\textbf{3.65} 
&\textbf{2.59}/\textbf{3.45}/\textbf{3.85}/\textbf{4.09}  \\  
AGCGRU+VI &1.29/1.52/1.65/1.74
&2.94/3.51/3.86/4.10 
&2.96/3.59/3.94/4.17 \\ \hline
\end{tabular}
\label{tab:vi_point}
\end{table*}

\begin{table*}[htbp]
\caption{Average CRPS, P10QL, and P90QL for PeMSD3, PeMSD4, PeMSD7, and PeMSD8 for 15/30/45/60 minutes horizons for AGCGRU+flow
and AGCGRU+VI. Lower numbers are better.}
\centering
\begin{tabular}{c|ccc}
\hline
\multirow{2}{*}{Algorithm} & \multicolumn{3}{c}{PeMSD3 (15/ 30/ 45/ 60 min)} \\ \cline{2-4} 
                  &CRPS       &P10QL(\%)       &P90QL(\%)       \\ \hline \hline
AGCGRU+flow  &\textbf{10.53}/\textbf{11.39}/\textbf{12.03}/\textbf{12.47}
&\textbf{4.01}/\textbf{4.44}/\textbf{4.76}/\textbf{4.97}
&\textbf{4.06}/\textbf{4.38}/\textbf{4.63}/\textbf{4.82}   \\
AGCGRU+VI &11.00/11.80/12.38/12.94
&4.14/4.53/4.82/5.10
&4.27/4.58/4.81/5.02 \\
\hline
\hline
\multirow{2}{*}{Algorithm} & \multicolumn{3}{c}{PeMSD4 (15/ 30/ 45/ 60 min)} \\ \cline{2-4} 
                  &CRPS       &P10QL(\%)       &P90QL(\%)       \\ \hline \hline
AGCGRU+flow &1.08/1.32/1.46/1.56
&1.28/1.62/1.82/1.97
&1.05/1.26/1.37/1.45   \\
AGCGRU+VI &1.08/\textbf{1.31}/\textbf{1.45}/\textbf{1.54}
&\textbf{1.26}/\textbf{1.59}/\textbf{1.79}/\textbf{1.93}
&\textbf{1.04}/\textbf{1.25}/\textbf{1.36}/1.45  \\
\hline  \hline
\multirow{2}{*}{Algorithm} & \multicolumn{3}{c}{PeMSD7 (15/ 30/ 45/ 60 min)} \\ \cline{2-4} 
                  &CRPS       &P10QL(\%)       &P90QL(\%)       \\ \hline \hline
AGCGRU+flow &1.73/2.18/2.43/\textbf{2.58}
&2.27/2.97/\textbf{3.36}/\textbf{3.60}
&1.83/2.25/2.48/\textbf{2.62}  \\
AGCGRU+VI &\textbf{1.72}/2.18/\textbf{2.42}/2.60
&\textbf{2.25}/2.97/3.39/3.66
&\textbf{1.80}/\textbf{2.24}/\textbf{2.47}/2.63 \\
 \hline \hline
\multirow{2}{*}{Algorithm} & \multicolumn{3}{c}{PeMSD8 (15/ 30/ 45/ 60 min)} \\ \cline{2-4} 
                  &CRPS       &P10QL(\%)       &P90QL(\%)        \\ \hline \hline
AGCGRU+flow &\textbf{0.90}/\textbf{1.10}/\textbf{1.20}/\textbf{1.28}
&\textbf{1.10}/\textbf{1.43}/\textbf{1.61}/\textbf{1.73}
&\textbf{0.87}/\textbf{1.01}/\textbf{1.09}/\textbf{1.14}   \\
AGCGRU+VI &0.95/1.13/1.24/1.31
&1.15/1.44/1.62/1.76
&0.90/1.03/1.10/1.15 \\
\hline
\end{tabular}
\label{tab:vi_probab}
\end{table*}

\subsection{Comparison of execution time, GPU memory usage and model size}
Table~\ref{tab:run_time} summarizes the run time, GPU usage during training, and the size of the learned model for AGCRN-ensemble, GMAN-ensemble, and the proposed AGCGRU+flow for the four PeMS datasets. We observe that if we choose the ensemble size so that the algorithms have an approximately equal execution time, then the model-size of the ensemble algorithms are comparable to our approach as well. However, our method requires more GPU memory compared to the ensembles during training because of the particle flow in the forward pass.  

\begin{table}[htbp]
\caption{Execution time, memory consumption (during training) and model size for AGCRN-ensemble, GMAN-ensemble and AGCGRU+flow for the four PeMS datasets. Lower numbers are better.}
\footnotesize
\setlength{\tabcolsep}{2pt}
\centering
\begin{tabular}{c|cccc}
\hline
\multirow{2}{*}{Algorithm} & \multicolumn{3}{c}{Execution time (minutes)} \\ \cline{2-5} 
                  &PEMS03       &PEMS04       &PEMS07  &PEMS08       \\ \hline \hline
AGCRN-ensemble &369 &243 &183 &224 \\
GMAN-ensemble &444 &224 &195 &185 \\
AGCGRU+flow &\textbf{325} &\textbf{205} &\textbf{154} &\textbf{177}   \\
\hline \hline
\multirow{2}{*}{Algorithm} & \multicolumn{3}{c}{GPU memory (GB)} \\ \cline{2-5} 
                 &PEMS03       &PEMS04       &PEMS07  &PEMS08       \\ \hline \hline
AGCRN-ensemble &\textbf{6.55} &\textbf{5.19} &\textbf{4.09} &\textbf{3.47} \\
GMAN-ensemble &15.45 &9.45 &8.46 &4.45 \\
AGCGRU+flow &25.27 &18.76 &12.45 &8.45   \\
\hline \hline
\multirow{2}{*}{Algorithm} & \multicolumn{3}{c}{Model Size (MB)} \\ \cline{2-5} 
                 &PEMS03       &PEMS04       &PEMS07  &PEMS08       \\ \hline \hline
AGCRN-ensemble &11.52 &11.52 &11.45 &11.45 \\
GMAN-ensemble &\textbf{9.54} &\textbf{9.51} &\textbf{9.45} &\textbf{9.35} \\
AGCGRU+flow &12.88 &12.86 &12.86 &12.85   \\
\hline 
\end{tabular}
\label{tab:run_time}
\end{table}
\newpage
\bibliographystyle{icml2021}
\bibliography{reference}
\end{document}